\documentclass[11pt]{article}
\usepackage{natbib}
\usepackage{latexsym}
\usepackage{amsmath}
\usepackage{amssymb}
\usepackage{amsthm}
\usepackage{epsfig}
\usepackage{xcolor}
\usepackage{graphicx}
\usepackage{bm}
\usepackage{enumitem}
\usepackage{mathtools}
\usepackage{graphicx}
\usepackage{tikz}
\usepackage{ mathrsfs }
\usepackage[toc,page]{appendix}
\usepackage{graphicx}
\usepackage{bbm}

\usepackage[top=1in, bottom=1in, left=1in, right=1in]{geometry}

\usepackage{hyperref}
\hypersetup{
    colorlinks,
    linkcolor={blue!80!black},
    citecolor={green!50!black},
}
\colorlet{linkequation}{blue}

\usepackage{subfig}
\usepackage{mathtools}
\newcommand{\PP}{\mathbb{P}}
\newcommand{\E}{\mathbb{E}}

\newcommand{\R}{\mathbb{R}}
\newcommand{\C}{\mathbb{C}}

\newcommand{\eps}{\varepsilon} 

\def\id{{\mathbf I}}

\newcommand{\<}{\langle}
\renewcommand{\>}{\rangle}
\newcommand{\sign}{\text{sign}}
\newcommand{\diag}{\text{diag}}

\newcommand{\infnorm}[1]{\lVert #1 \rVert_{\infty}}
\def\sT{{\mathsf T}}
\def\bzero{{\boldsymbol 0}}

\newcommand{\norm}[1]{\lVert #1 \rVert}
\newcommand{\perm}[2]{\mathtt{P}(#1,#2)}

\newtheorem{theorem}{Theorem}[section]

\newtheorem{assumption}[theorem]{Assumption}
\newtheorem{assump}[theorem]{A}
\newtheorem{lemma}[theorem]{Lemma}
\newtheorem{proposition}[theorem]{Proposition}
\newtheorem{corollary}[theorem]{Corollary}

\newtheorem{definition}{Definition}

\newtheorem{claim}{Claim}

\newtheorem{remark}[theorem]{Remark}

\DeclareSymbolFont{rsfs}{U}{rsfs}{m}{n}
\DeclareSymbolFontAlphabet{\mathscrsfs}{rsfs}

\def\P{{P}}

\def\bH{{\boldsymbol H}}

\def\bK{{\boldsymbol K}}

\def\bM{{\boldsymbol M}}

\def\bI{{\mathbf{I}}}
\def\ba{{\boldsymbol a}}

\def\bi{{\boldsymbol i}}
\def\bj{{\boldsymbol j}}

\def\bu{{\boldsymbol u}}
\def\bv{{\boldsymbol v}}
\def\bw{{\boldsymbol w}}
\def\bx{{\boldsymbol x}}

\def\bnu{{\boldsymbol \nu}}
\def\bz{{\boldsymbol z}}

\def\bbeta{{\boldsymbol \beta}}

\def\blambda{{\boldsymbol \lambda}}

\def\bphi{{\boldsymbol \phi}}
\def\btheta{{\boldsymbol \theta}}

\def\bxi{{\boldsymbol \xi}}

\def\bLambda{{\boldsymbol \Lambda}}

\def\bTheta{{\boldsymbol \Theta}}

\def\cR{\mathcal{R}}

\def\defeq{{:=}}

\def\supp{{\rm supp}}

\def\de{{\rm d}}

\def\de{{\rm d}}
\def\Unif{{\rm Unif}}

\def\s{{s}}

\def\cV{{\mathcal V}}

\def\cP{{\mathcal P}}

\def\cC{{\mathcal C}}
\def\cQ{{\mathcal Q}}
\def\cL{{\mathcal L}}
\def\cF{{\mathcal F}}

\def\cS{{\mathcal S}}
\def\cI{{\mathcal I}}
\def\cV{{\mathcal V}}

\def\cP{{\mathcal P}}

\def\cH{{\mathcal H}}
\def\cA{{\mathcal A}}
\def\cK{{\mathcal K}}

\def\K{{\mathbb K}}

\def\Unif{{\sf Unif}}
\def\normal{{\sf N}}

\def\NN{{\sf NN}}

\def\naturals{{\mathbb N}}

\def\normal{{\sf N}}

\def\K{{\mathbb K}}

\def\Unif{{\sf Unif}}
\def\normal{{\sf N}}

\def\NN{{\sf NN}}

\def\naturals{{\mathbb N}}

\def\cD{{\mathcal D}}
\def\cX{{\mathcal X}}
\def\cF{{\mathcal F}}
\def\cS{{\mathcal S}}
\def\cI{{\mathcal I}}

\def\de{{\rm d}}
\def\Unif{{\rm Unif}}

\def\normal{{\sf N}}

\def\cX{{\mathcal X}}

\def\btheta{{\boldsymbol \theta}}
\def\barbtheta{\Bar{{\boldsymbol \theta}}}
\def\barrho{\Bar{\rho}}
\def\bTheta{{\boldsymbol \Theta}}
\def\bLambda{{\boldsymbol \Lambda}}
\def\blambda{{\boldsymbol \lambda}}

\def\cM{{\mathcal M}}

\def\cV{{\mathcal V}}

\def\diag{{\rm diag}}

\def\osigma{\overline{\sigma}}

\def\bzeta{{\boldsymbol \zeta}}

\def\ind{\mathbbm{1}}
\def\brho{{\boldsymbol{\rho}}}

\def\balpha{\boldsymbol{\alpha}}
\def\bgamma{\boldsymbol{\gamma}}

\def\cY{\mathcal{Y}}

\def\balpha{\boldsymbol{\alpha}}
\def\bgamma{\boldsymbol{\gamma}}

\def\sA{{\sf A}}

\def\Leap{{\sf Leap}}
\def\relLeap{{\sf relLeap}}

\def\Tc{T_{{\sf C}}}
\def\tTheta{\widetilde{\Theta}}

\def\iid{\overset{\scriptscriptstyle\text{i.i.d.}}{\sim}}

\def\osigma{\overline{\sigma}}

\def\Id{{{\rm Id}}}
\newcommand{\fNN}{\hat{f}_{\NN}}

\def\SQ{{\sf SQ}}
\def\naSQ{{\sf naSQ}}
\def\CSQ{{\sf CSQ}}
\def\naCSQ{{\sf naCSQ}}
\def\DLQ{{\sf DLQ}}
\def\naDLQ{{\sf naDLQ}}

\def\Cover{{\sf Cover}}
\def\relCover{{\sf relCover}}

\def\naA{{\sf naA}}
\def\tbx{\Tilde{\bx}}

\def\bm{{\boldsymbol m}}

\def\bkappa{{\boldsymbol \kappa}}

\def\ocR{\overline{\cR}}

\def\hatbu{\hat{\bu}}

\def\bfeta{{\boldsymbol \eta}}
\def\Lip{{\rm Lip}}

\def\obH{\overline{\bH}}
\def\obLambda{\overline{\bLambda}}

\title{On the Complexity of Learning Sparse Functions with\\Statistical and Gradient Queries}
\author{%
  Nirmit Joshi \\
  \texttt{nirmit@ttic.edu} 
   \and
  Theodor Misiakiewicz \\
\texttt{theodor.misiakiewicz@ttic.edu} 
\and 
Nathan Srebro \\
\texttt{nati@ttic.edu} 
}
\date{Toyota Technological Institute at Chicago}

\begin{document}
\maketitle
\begin{abstract}
The goal of this paper is to investigate the complexity of gradient algorithms when learning sparse functions (juntas). We introduce a type of Statistical Queries ($\SQ$), which we call Differentiable Learning Queries ($\DLQ$), to model gradient queries on a specified loss with respect to an arbitrary model. We provide a tight characterization of the query complexity of $\DLQ$ for learning the support of a sparse function over generic product distributions. This complexity crucially depends on the loss function. For the squared loss, $\DLQ$ matches the complexity of Correlation Statistical Queries $(\CSQ)$—potentially much worse than $\SQ$. But for other simple loss functions, including the $\ell_1$ loss, $\DLQ$ always achieves the same complexity as $\SQ$. We also provide evidence that $\DLQ$ can indeed capture learning with (stochastic) gradient descent by showing it correctly describes the complexity of learning with a two-layer neural network in the mean field regime and linear scaling.
\end{abstract}

\section{Introduction}\label{sec:intro}
In recent years, major efforts have been devoted to understanding which distributions can be learned efficiently using gradient-type algorithms on generic models 
\citep{abbe2020universality,allen2020backward,malach2021quantifying,damian2022neural,abbe2021power,abbe2021staircase,abbe2022merged,abbe2023sgd,bietti2023learning,glasgow2023sgd,dandi2023learning,dandi2024benefits,edelman2024pareto,kou2024matching}. In this paper, we focus on learning sparse functions (i.e.~``juntas'' \citep{blum1997selection}), that is functions that depend only on a small number $\P$ out of a much larger set $d \gg \P$ of input coordinates. The challenge in this setting is to identify the few relevant coordinates. 
For some sparse functions, such as noisy parities, learning is believed to require $O(d^\P)$ runtime \citep{kearns1998efficient}, while others, such as linear functions, are easy to learn in $\tilde{O}(d)$ time. 
Which functions are easy to learn and which are hard?  What is the complexity of learning a specific sparse function?  Recent works \citep{abbe2022merged,abbe2023sgd} unveiled  a rich ``leap'' hierarchical structure and saddle-to-saddle dynamics that drives learning in this setting. The goal of the present paper is to provide a general characterization for the complexity of learning sparse functions that go beyond \textit{(i)} hypercube data and Fourier analysis, and \textit{(ii)} $\CSQ$ (see below) and focusing only on the squared loss.

The notion of complexity we consider is the \textit{Statistical Query} ($\SQ$) complexity, which studies learning by measuring expectations up to some worst-case tolerance (see \citep{kearns1998efficient,bshouty2002using,reyzin2020statistical} and Section \ref{sec:sq-and-DLQ}). Although based on worst-case error (or almost equivalently, additive independent noise) rather than the sampling error encountered in practice, statistical query complexity has been proven to be a useful guideline for studying the complexity of learning.  In particular, gradient computations are a special case of statistical queries.  In specific cases which include binary functions or gradients on the squared or cross-entropy loss, gradient queries are equivalent\footnote{This equivalence holds if the input distribution is known, which is the case we consider here.} to the restricted class of {\em Correlation Statistical Queries} ($\CSQ$) which are strictly less powerful than general statistical queries.  Lower bounds on the $\CSQ$ complexity have thus been seen as corresponding to the complexity of gradient-based learning\footnote{In Section \ref{sec:discussion}, we discuss recent work \cite{dandi2024benefits} that showed that even with the squared loss, multiple gradient evaluations on the same batch can be strictly more powerful than $\CSQ$.} in these restricted cases. Part of the motivation for this paper is to emphasize that this relationship is limited to very specific loss functions and does not hold more generally.  In order to study the complexity of gradient algorithms for general output and loss functions, we introduce a type of statistical query which we call \textit{Differentiable Learning Query} ($\DLQ$). These queries are defined with respect to a specific loss $\ell$---denoted by $\DLQ_\ell$---and are given by gradients on a loss $\ell$ with respect to an arbitrary model. Specifically, $\DLQ_\ell$ algorithms correspond to $\SQ$ algorithms with queries of the type
\begin{equation}\label{eq:DLQ_intro}
\phi (y,\bx) = \frac{\partial}{\partial \omega} \ell (f(\bx,\omega), y) \Big\vert_{\omega = 0},  \;\;\;\text{where} \quad f : \cX^d \times \R \to \R.
\end{equation}
Depending on the loss $\ell$ and target distributions, learning with $\DLQ_\ell$ can be less, equal, or more powerful than $\CSQ$.

For inputs on the hypercube $\bx \sim \Unif ( \{ \pm 1\}^d)$ and $\CSQ$, \cite{abbe2022merged,abbe2023sgd} showed that the complexity of learning sparse functions is sharply captured by a \textit{leap exponent} defined in terms of the non-zero Fourier coefficients of the sparse function. Informally, it states that $\Theta (d^{k_*})$ queries are necessary and sufficient to learn with $\CSQ$, where $k_*$ is the minimum number of coordinates one need to add at once to ``leap'' between non-zero Fourier coefficients. E.g., consider the following two sparse functions:
\begin{equation}\label{eq:examples_intro}
\begin{aligned}
y_1 =&~ x_{s_*(1)} + x_{s_*(1)} x_{s_*(2)} + x_{s_*(1)}x_{s_*(2)}x_{s_*(3)} + x_{s_*(1)}x_{s_*(2)}x_{s_*(3)}x_{s_*(4)}, \\
y_2 =&~ x_{s_*(1)}x_{s_*(2)}x_{s_*(3)} + x_{s_*(1)}x_{s_*(2)}x_{s_*(4)} +x_{s_*(1)}x_{s_*(3)}x_{s_*(4)} + x_{s_*(2)}x_{s_*(3)}x_{s_*(4)}.
\end{aligned}
\end{equation}
Both functions depend on an (unknown) subset of $\P = 4$ coordinates $\{s_*(1),s_*(2),s_*(3),s_*(4)\}$. For $y_1$, the monomials are ordered such that we add only one new coordinate to the support at a time: the leap exponent is $1$ and $\Theta(d)$ queries are sufficient to learn $y_1$. For $y_2$, we need to add three new coordinates at once: the leap exponent is $3$ and $\Theta (d^3)$ queries are required to learn $y_2$.


We extend and generalize this charactarization in several significant ways:
\begin{itemize}
    \item[\textit{(a)}] We go beyond binary input space and allow for arbitrary measurable space with product distribution. In particular, the leap complexity arises from the structure of the permutation group rather than a specific functional basis. 

    \item[\textit{(b)}] We go beyond $\CSQ$ and squared loss, and tightly characterize the query complexity for learning sparse functions with $\SQ$ and $\DLQ_\ell$ algorithms with any loss $\ell$. This complexity is in terms of a leap exponent defined analogously as above, but now over a set system $\cC_{\sA} \subseteq 2^{[P]}$ of \textit{``detectable''} subsets $S \subseteq [P]$, where ``detectability'' depends on the type of queries $\sA \in \{\SQ,\CSQ,\DLQ_\ell\}$. Learning with different loss functions are not necessarily comparable to $\CSQ$ or $\SQ$ and depend on the sparse function.
    However, we show that some losses, such as $\ell_1$ or exponential loss, are ``generic'', i.e. $\cC_{\DLQ_\ell} = \cC_{\SQ}$ for any sparse function, and always match the $\SQ$-leap exponent. This shows that differentiable learning with these losses is as powerful as learning with $\SQ$ for juntas. 
    \item[\textit{(c)}] Finally, we introduce a \textit{cover exponent}---defined on the same system of detectable sets $\cC_\sA$---which captures learning with $\sA\in \{\SQ,\CSQ,\DLQ_\ell\}$ when the queries are chosen \textit{non-adaptively}, i.e., without adapting to the responses of previous queries. This can be roughly thought of as learning with a single gradient step, versus many consecutive steps in the adaptive case.  The contrast between the two helps understand to what extent learning relies on adaptivity and can help frame other related results---see discussion in Section \ref{sec:discussion}.
\end{itemize}

\begin{table}[!t]
    \centering
    \begin{tabular}{|c|c|c|}
    \hline
        $q/\tau^2=\Theta(d^{k_*})$ & Adaptive & Non-adaptive \\
         \hline
        $\SQ$ & $k_*=\Leap\left(\cC_{\SQ }\right)$ & $k_*=\Cover(\cC_{\SQ})$\\
        \hline
        $\CSQ$ & $k_*=\Leap(\cC_{\CSQ})$ & $k_*=\Cover(\cC_{\CSQ})$\\
        \hline
        $\DLQ_{\ell}$ & $k_*=\Leap( \cC_{\DLQ_{\ell}})$ & $k_*=\Cover(\cC_{\DLQ_{\ell}})$\\
        \hline
    \end{tabular}
    \vspace{2mm}
    \small \caption{\label{tab:summary} \small Complexity of learning a sparse function over $d$ input coordinates with different query types based on Theorem \ref{thm:meta_complexity_leap}. $\DLQ_{\ell}$ is defined in Section \ref{sec:setting}, $\Leap$ and $\Cover$ of a set system in Definition \ref{def:leap-and-cover}, and systems $\cC_\sA$ of detectable sets in Definition \ref{def:leap_cover_exponent} based on test functions depending on query type $\sA$, as specified in Eq.~\eqref{eq:Psi}.}
\end{table}

We summarize the complexity of learning a sparse function with $\SQ,\CSQ,\DLQ_\ell$ and adaptive/non-adaptive queries in Table \ref{tab:summary}. To see some of these notions play out, consider again the two examples in Eq.~\eqref{eq:examples_intro}:
\[
\begin{aligned}
    y_1:&~ \qquad \Leap (\cC_{\CSQ}) = 1, \quad \Cover(\cC_{\CSQ}) = 4, \qquad &~\Leap (\cC_{\SQ}) = \Cover (\cC_{\SQ}) = 1, \\
    y_2:&~ \qquad \Leap (\cC_{\CSQ}) = \Cover (\cC_{\CSQ}) = 3, \qquad &~ \Leap (\cC_{\SQ}) = \Cover (\cC_{\SQ}) = 1.
\end{aligned}
\]
For $y_1$, $\DLQ_\ell$ will require $\Theta (d)$ adaptive or $\Theta (d^4)$ non-adaptive queries to learn with squared loss---equivalent to $\CSQ$, while $\Theta (d)$ adaptive/non-adaptive queries suffice with $\ell_1$-loss---equivalent to $\SQ$. For $y_2$, $\DLQ_\ell$ will only require $\Theta(d)$ queries, either adaptive or non-adaptive, with $\ell_1$-loss, compared to $\Theta (d^3)$ queries with the squared loss. Proposition \ref{prop:DLQvsSQ_CSQ}.(c) provides an example where $\Leap (\cC_{\SQ}) < \Leap (\cC_{\DLQ_\ell}) < \Leap (\cC_{\CSQ})$. 

$\DLQ_\ell$ algorithms are quite different than (stochastic) gradient descent algorithms. On one hand, $\DLQ_\ell$ allows evaluating arbitrary gradients rather than following a trajectory, and is thus potentially more powerful than (S)GD on generic models\footnote{Following \cite{abbe2021power}, we can construct an ``emulation'' differentiable model, i.e., a model $f(\bx;\bw)$ on which GD emulates $\DLQ_\ell$. However, this offers limited insights as the model $f(\bx;\bw)$ is highly non-standard.}. On the other hand, the lower bound on $\DLQ_\ell$ algorithms in Table \ref{tab:summary} is against \textit{worst-case} noise---much more pessimistic than sampling noise encountered in SGD or GD-on-the-training-set \citep{abbe2021power}. Nevertheless, $\DLQ_\ell$  does provide useful guidance and we expect that it {\em does} capture important aspects of the computational complexity of GD in a number of settings.  To demonstrate this, Section \ref{sec:gd-on-NN} considers learning sparse functions on the hypercube using online SGD on a two-layer neural network. We show that in this setting, the $\DLQ_\ell$-leap exponent correctly captures learnability in the mean-field regime and linear scaling. Namely, for $\DLQ_\ell$-leap $1$ functions, SGD on loss $\ell$ learns in $O(d)$-steps, while for leap greater than $1$, the dynamics remains stuck in a suboptimal saddle subspace and require $\omega(d)$-steps to escape.

\section{Setting} \label{sec:setting}
\paragraph{Junta Recovery Problem.} A sparse recovery (junta learning) problem with $\P$ relevant coordinates is defined in terms of a coordinate input space $\cX$, an output space $\cY$\footnote{Formally, $\cX$ and $\cY$ are measurable Polish spaces endowed with sigma algebras $\Sigma_{\cX}$ and $\Sigma_{\cY}$ respectively.}, a marginal distribution $\mu_{x}$ over $\cX$ and a link distribution $\mu_{y\mid \bz}$ over $\cY$ given elements of $\cX^\P$. This specifies a joint distribution $\mu_{y,\bz}$ over a measurable space $\cY \times \cX^\P$, where the marginal distribution $\bz \sim \mu_{x}^\P$ is the product distribution, and $y \mid \bz \sim \mu_{y \mid \bz}$. We further denote $\mu_{y}$ to be the marginal distribution over $\cY$. We further denote the junta problem by a tuple $\mu:=(\mu_x,\mu_{y\mid \bz})$, where the support size $P$ and the spaces $\cX$ and $\cY$ are implicit. Consider some examples,   
\begin{equation}
    \bz \sim \Unif(\{\pm 1\}^\P), \;\;\;y=h_*(\bz)+\eps\; \;\text{ for some $\;h_*:\{\pm 1\}^\P \to \R\;$ and noise $\eps$}; 
\end{equation}
\begin{equation}
    \bz \sim \normal(\bzero, \bI_{\P}), \;\;\;y \sim \text{Bernoulli}(\text{sigmoid}(h_*(\bz))) \;\;\text{ for some target $\;h_*: \R^\P \to \R$}.
\end{equation}
For $d \geq P$, we define the junta recovery problem as learning the family of distributions 
\begin{equation}
    \cH_{\mu}^d := \left\{ \cD^d_{\mu, \s}: \s \in \perm{[d]}{\P}\right\}
\end{equation}
where $\perm{[d]}{\P}$ is the set of non-repeating sequences from $[d]$ of length $\P$, and the distribution $\cD_{\mu, \s}^d$ is a distribution over $\cY \times \cX^d$ such that
$$ \bx \sim \mu_{x}^d\;\; \text{ and } \;\;y \mid (x_{\s(1)},\dots, x_{\s(\P)}) \sim \mu_{y \mid \bz}.$$
In words, $\cH^d_{\mu}$ is the set of distributions on $(y,x_1, \ldots,x_d)$ where $\bx$ follows the product distribution $\mu_x^d$ and the output $y$ only depends on an (unknown) sequence of $\P$ coordinates. For ease of notation, we will denote $\cD^d_\s = \cD^d_{\mu,\s}$ when the junta problem $\mu=(\mu_{x},\mu_{y \mid \bz})$ is clear from context.

\paragraph*{Success as Support Recovery.} For any sequence $s\in \perm{[d]}{\P}$, we additionally denote the associated unordered set of its elements by $S:=\{s(i): i \in [\P]\}$.
    We consider learning as ``succeeding'' if it outputs the correct index set.  That is, we say a junta learning algorithm ``succeeds'' in learning $\cH^d_{\mu}$ if for every $\s_* \in \perm{[d]}{\P}$, it outputs $\hat{S}\subseteq [d]$ such that  $\hat{S}=S_*=\{s_*(i): i \in [\P]\}.$

For our purpose, just the recovery of relevant coordinates is the objective as we only care about the complexity as a scaling of $d$. Once we recover the relevant coordinates $S_*$, learning the ordered sequence $\s_*$ corresponds to a problem of size $\P$, and thus its complexity is independent of $d$. Note that the precise ordering $\s_*$ may not be identifiable if for some other $\s_*'$ with $S_*=S_*'$ we have $\cD^d_{\s_*}=\cD^d_{\s_*'}$; this is possible when the measure $\mu_{y \mid \bz}$ has symmetries with respect to coordinates of $\bz$. We further emphasize that even the support $S_*$ may not be identifiable, when $y$ is independent of some coordinates of $\bz$ or when only using a restricted query access to the distribution---we will return to this issue in Section \ref{sec:main-results}. 

\paragraph*{Does the learner know the link $\mu_{y \mid \bz}$?} We always assume that the learner knows $\mu_{x}$. In our formulation, we also assume that the learner additionally knows the link $\mu_{y\mid \bz}$, and the only unknown is $\s_*$. Indeed, our lower bounds hold even for this easier setting. But under mild assumptions, our upper bounds can be obtained with an algorithm that does not require knowing $\mu_{y \mid \bz}$ (the complexity still depends on the problem $\mu$)---see discussion in Section \ref{sec:main-results}. 

\paragraph{Well Behaved Distributions.} 
For our lower bounds, we will consider junta problems $ \mu=(\mu_{x}, \mu_{y \mid \bz})$, which are ``well-behaved'' in a way that is standard in the hypothesis testing literature. Let $ \mu_{y,\bz}$ be the induced joint distribution on $(y,\bz)$ where $\bz\sim\mu_x^P$ and $y\mid\bz\sim\mu_{y\mid\bz}$. For any subset $U \subseteq [\P]$, let $\mu_{y,\bz,U}$ be the marginal distribution of $(y,(z_i)_{i\in U})$ and 
\begin{equation}\label{eq:def_cD_S}
\mu_{y,\bz,U}^{0} = \mu_{y,\bz,U} \otimes \mu_{\cX}^{\P - |U|} ,
\end{equation}
meaning that $(y,z_1, \ldots, z_\P) \sim \mu_{y,\bz,U}^{0}$ has $(y,(z_i)_{i\in U}) \sim \mu_{y,\bz,U}$ and $(z_i)_{i \in [\P] \setminus U} \iid \mu_{x}$ independently of $(y,(z_i)_{i\in U})$ (we replace $z_i$ for $i\notin U$ with independent draws from the marginal). The marginal distribution of $\bz$ under $\mu_{y,\bz,U}^{0}$ is still $\mu_{x}^P$.
\begin{assumption}\label{ass:well-behaved}
 For any $U \subseteq [\P]$, we have $\mu_{y,\bz} \ll \mu_{y,\bz,U}^{0}$, and the Radon-Nikodym derivative $\de \mu_{y,\bz} / \de \mu_{y,\bz,U}^{0}$ is square integrable w.r.t.~$\mu_{y,\bz,U}^{0}$, i.e.
\begin{equation}\label{eq:def_ocP_P}
   \hspace{-5pt} \frac{\de \mu_{y,\bz}}{\de \mu_{y,\bz,U}^{0}} \in L^2 ( \mu_{y,\bz,U}^{0})\;\; \text{ for all $\;\;U \subseteq [\P]$}.
\end{equation}
\end{assumption}
This is a standard and implicit assumption in the hypothesis testing literature whenever a corresponding null distribution is considered (here all $\mu_{y,\bz,U}^{0}$, i.e., with label $y$ depending only on a strict subset $U \subsetneq [P]$); more specifically for statistical query lower bounds \citep{feldman2017statistical,damian2024computational}, low-degree likelihood ratio \citep{hopkins2018statistical,kunisky2019notes}, or contiguity lower bounds \citep{perry2018optimality}.  It always holds when $\cX$ is discrete. We further comment on the necessity of this assumption in Remark \ref{rmk:abs_continuity} in the next section.

\section{Statistical and Differentiable Learning Queries}\label{sec:sq-and-DLQ}
We will consider three classes of learning algorithms, all based on the statistical query paradigm, but differing in the type of queries allowed, as captured by a set $\cQ$ of allowed queries.  

For a number of queries $q$, tolerance $\tau>0$ and a set $\cQ \subseteq \R^{\cY \times \cX^d}$ of measurable functions $\cY \times \cX^d \rightarrow \R$, a {\bf $\cQ$-restricted statistical query algorithm} $\cA \in \cQ\text{-}\SQ(q,\tau)$ for junta learning takes an input distribution $\cD$ over $\cY \times \cX^d$ and operates in $q$ rounds where at each round $t \in \{1, \ldots , q\}$, it issues a query $\phi_t\in\cQ$, and receives a response $v_t$ such that
\begin{equation}\label{eq:SQ_def}
\left\lvert v_t - \E_{\cD} \left[\phi_t(y,\bx)\right] \right\rvert \leq \tau \sqrt{\E_{\cD_0}[\phi_t(y,\bx)^2]},
\end{equation}
where $\cD_0=\cD_y \otimes \cD_{\bx}$ is the associated decoupled ``null'' distribution where $\bx$ and $y$ are independent, but follow their marginals\footnote{Equivalently, we could restrict the $L_2(\cD_0)$ norm of the query. See also Remark \ref{rmk:Linfty_queries}.}. The query $\phi_t$ can depend on the past responses $v_1, \ldots , v_{t-1}$. After issuing $q$ queries, the learner $\cA$ outputs $\hat{S}\subseteq[d]$.  We say that \textbf{$\cA$ succeeds in learning $\cH^d_\mu$} if for any $\cD_{s_*} \in \cH^d_\mu$ and any responses $v_t$ satisfying \eqref{eq:SQ_def} for $\cD=\cD_{s_*}$, $\cA$ outputs $\hat{S}=S_*$.

Above we allow the queries to be chosen adaptively, i.e., depending on past responses. We also consider {\bf{$\cQ$-restricted non-adaptive statistical query algorithms}}, which we denote by $\cQ\text{-}\naSQ(q,\tau)$, where the query functions $\{ \phi_t \}_{t \in [q]}$ are fixed in advance and do not depend on the past responses.  I.e.~a non-adaptive algorithm is specified by a list of queries and a mapping from the responses to an output $\hat{S}\subseteq[d]$.
\paragraph{Statistical Queries (SQ):} In regular, unrestricted \textit{Statistical Query} learning, the allowed query set, denoted by $\cQ_{\SQ}$, is the set of all measurable functions.  With slight overloading of notation, we refer to the class of these algorithms simply as $\SQ(q,\tau)$ and $\naSQ(q,\tau)$.
\paragraph{Correlation Statistical Queries (CSQ):} It is a special subclass of statistical queries, which require $\cY \subseteq \R$ (usually we allow the input and output spaces to be abstract), and are restricted to:
\begin{equation}
    \cQ_\CSQ = \left\{ \phi(y,\bx)=y\cdot \tilde{\phi}(\bx) \mid \tilde{\phi}:\bx \rightarrow \R \textrm{ measurable} \right\}.
\end{equation}
We denote the class of adaptive and non-adaptive algorithms making such queries as $\CSQ(q,\tau)$ and $\naCSQ(q,\tau)$ respectively.

\paragraph{Differentiable Learning Queries (DLQ) with Loss $\ell$:} Let $\cF \subseteq V$, which is an open subset of some normed vector space $V$, be the output space of our models, e.g.~usually $V=\cF=\R$ for models with a single output unit, but we may have $V=\cF=\R^r$ with multiple output units. We consider a loss function $\ell: \cF \times \cY \to \R$ that is locally Lipschitz continuous in its first argument for every $y \in \cY$. The loss is additionally equipped with a derivative operator $\nabla \ell: \cF \times \cY \to V$ as a part of the loss definition such that for any $\bu\in \cF$ and $y \in \cY$, we have $\nabla \ell(\bu,y) \in \partial_{1} \ell(\bu,y)$, the set of generalized Clarke subderivatives of $\ell(\cdot,y)$ at $\bu \in \cF$. This is a standard generalization of derivatives to non-differentiable and non-convex losses; in particular, note that $(i)$ for differentiable losses, $\partial_{1}\ell(\bu,y)$ is a singleton with the true gradient of $\ell(\cdot,y)$ at $\bu \in \cF$, and $(ii)$ for convex losses (in the first argument), $\partial_{1}\ell(\bu,y)$ is the set of subderivatives of $\ell(\cdot,y)$ at $\bu \in \cF$. Finally, let $ \cM:=\{f: \cX^d \times \R \to \cF \mid f(\bx, \omega) \text{ is differentiable at } \omega=0 \text{ for all } \bx \in \cX^d\}$ be the set of allowed models\footnote{We restrict the model class $\cM$ to only contain models that are always differentiable at $\omega=0$ without loss of generality, as for any non-differentiable model $f_1: \cX^d\times \R \to \cF$ under any ``reasonable'' generalized notion of derivative with $\frac{\de}{\de \omega} f_1(\bx,\omega) \mid_{\omega=0}=g(\bx) \in V$, one can define a differentiable model $f_2(\bx,\omega)=f_1(\bx,0)+\omega \cdot g(\bx) $ such that $f_2(\bx,0)=f_1(\bx,0)$ and $\frac{\de}{\de \omega} f_2(\bx,\omega) \mid_{\omega=0}=g(\bx)$.}. Then the allowed differentiable learning query set is:
 \begin{equation}\label{eq:query-set-DLQ}
     \cQ_{\DLQ_\ell} = \left\{ \phi(y,\bx) = \left[ \frac{\de}{\de \omega} f (\bx, \omega) \right]^\sT_{\omega = 0} \nabla \ell(f (\bx, 0),y)  \;\middle|\; f \in \cM \right\}.\;
 \end{equation} 
 That is, at each round the algorithm chooses a parametric model $f(\bx,\omega)$, parameterized by a single scalar $\omega$, and the query corresponds to the derivative (with respect to the single parameter $\omega$) of the loss applied to the model, at $\omega=0$. This captures the first gradient calculation for a single-parameter model initialized at zero.  But the derivative at any other point can also be obtained by querying at a shifted model $f_{(\nu)}(\bx,\omega)=f(\bx,\nu+\omega)$, and the gradient with respect to a model $f(\bx,\bw)$ with $r$ parameters $\bw\in\R^r$ can be obtained by issuing $r$ queries, one for each coordinate, of the form 
 $f_{(\bw,i)}(\bx,\omega)=f(\bx,\bw+\omega e_i)$, where $e_i$ is the standard basis vector.  Queries of the form $\cQ_{\DLQ_\ell}$ can thus be used to implement gradient calculations for any differentiable model, noting that the number of queries $q$ is the number of gradient calculations {\em times the number of parameters}. Finally, observe that, for differentiable losses, the queries of the form \eqref{eq:DLQ_intro} are equivalent to the form mentioned in \eqref{eq:query-set-DLQ} due to the chain rule.
 
We denote the class of adaptive and non-adaptive algorithms making such queries as \\$\DLQ_\ell(q,\tau)\defeq \cQ_{\DLQ_\ell}\text{-}\SQ(q,\tau)$ and $\naDLQ_\ell(q,\tau)\defeq \cQ_{\DLQ_\ell}\text{-}\naSQ(q,\tau)$.

 \begin{remark}\label{rmk:Linfty_queries}
    More common in the SQ literature is to restrict the query in $L^\infty$ (or equivalently, require precision relative to $L^\infty$).  Precision relative to $L^2(\cD_0)$ is more similar to VSTAT \citep{feldman2017statistical}, and is more powerful than relative to $L^\infty$, and our lower bounds hold against this stronger notion. In our algorithms and upper bounds we only need this additional power when $y \mapsto \nabla \ell (\bu, y)$ is unbounded.  If we further assume that these functions are bounded, e.g., the labels $y$ are bounded and $\nabla \ell$ continuous, our queries have bounded $L^\infty$ and thus operate in the more familiar SQ setting.
 \end{remark}

\begin{remark} \label{rmk:abs_continuity}
    Let us further elaborate on the necessity of Assumption \ref{ass:well-behaved} in our setting. Informally, we require the label to have ``enough noise''. This is necessary to obtain meaningful lower-bounds for general SQ algorithms, as noted in \cite{valiant2012finding,song2017complexity,vempala2019gradient}.  When learning real-valued target function classes, allowing for general measurable queries is too weak (note that arbitrary measurable functions are not practical anyway). Indeed, \cite{vempala2019gradient} showed that any finite set $\cH$ of (noiseless) functions can be learned with $\log |\cH|$ (bounded) queries and constant tolerance\footnote{The result holds under a ``non-degeneracy'' condition, namely, for any $f,g \in \cH$, $f \neq g$ almost surely. Consider for example $y = x_{s_*(1)}$ with $\bx \sim \normal (\bzero, \id_d)$: using the construction in \cite{vempala2019gradient}, we can learn $\cH_\mu^d$ with $q/\tau^2 = \Theta (\log (d))$ queries instead of $q/\tau^2 = \Theta (d)$ as in Theorem \ref{thm:meta_complexity_leap}.}. They identify three possible approaches to address this challenge: (i) Require $y \mapsto \phi (y,\bx)$ to be Lipschitz for every fixed $\bx$; (ii) Insist on noisy concepts, e.g., $y = f(\bx) + \zeta$, with $\zeta \sim \normal (0,\sigma^2)$, which is equivalent to Lipschitz queries; (iii) Restrict the form of the queries, such as in $\CSQ$. Our Assumption \ref{ass:well-behaved} can be viewed as a quantitative version of approach (ii). For approach (iii), if we restrict ourselves to $\CSQ$ or $\DLQ_\ell$ with $\ell$ piecewise analytic, we can indeed remove Assumption \ref{ass:well-behaved}  by  modifying the proof of Theorem \ref{thm:meta_complexity_leap}.
\end{remark}

\section{Leap and Cover Complexities}\label{sec:Leap-and-Cover}
\begin{definition}\label{def:leap-and-cover}
We define the leap and cover complexities for any system of subsets $\cC \subseteq 2^{[\P]}$.
\begin{enumerate} 
    \item[(i)] For any system of subsets $ \cC \subseteq 2^{[\P]}$, its \emph{leap complexity} is defined as
     \begin{equation}\label{eq:leap_set_definition}
    \Leap (\cC) := \min_{\substack{U_1, \dots, U_r \in \cC\\ \bigcup\limits_{i=1}^{r} U_i=[\P]}} \;\max_{i \in [r]}\; | U_{i} \setminus \cup_{j=0}^{i-1} U_{j} | .
    \end{equation}
    \item[(ii)] We define the \emph{cover complexity} of $\cC$ to be 
        \begin{equation}
    \Cover ( \cC) := \max_{i \in [\P]} \; \min_{ U \in \cC, i \in U} \;|U|.
    \end{equation}
\end{enumerate}
\end{definition}
\begin{remark}
We always have $\Leap(\cC) \leq \Cover(\cC)$. Both $\Leap$ and $\Cover$ complexities are closed under taking the union of subsets in $\cC$. Also, when $\supp(\cC)=\cup_{U \in \cC} U \neq [\P]$, then we use the convention $\Leap(\cC)=\Cover(\cC)=\infty$.  See a discussion about this convention in Section \ref{app:relative} and in particular, the definition of the relative leap and cover complexities in Definition \ref{def:relleap relcover}. 
\end{remark}
Here $\cC$ is the system of subsets which are ``detectable'', and will depend on the the query access model. Intuitively, the leap and cover complexities of $\cC$ capture the exponent of $d$ in the query complexity when recovering the support of an unknown $\s_* \in \perm{[d]}{\P}$, for adaptive and non-adaptive algorithms respectively. To discover the relevant coordinates of $s_*$, that correspond to $U \in \cC$, one needs to enumerate over $\Theta (d^{|U|})$. Hence, a non-adaptive algorithm, which fixes the queries in advance, requires $\Theta (d^{k_i})$ queries to discover $i^\mathrm{th}$ relevant coordinate i.e. $s_*(i)$, where $k_i = \min_{i \in U \in \cC} |U|$. Therefore, non-adaptive algorithms need a total number of queries that scales as $\Theta (d^{\Cover (\cC)})$ to learn $\supp(s_*)$. On the other hand, adapting queries using previous answers can greatly reduce this complexity as seen in the example in \eqref{eq:examples_intro} in Section \ref{sec:intro}. This is captured by the leap complexity, which measures the maximum number of coordinates we need to discover at once. 
Finally, the set system of detectable subsets will depend on the type of allowed queries. 
\begin{definition}[Detectable Subsets]\label{def:leap_cover_exponent}
    Let $\mu$ be a junta problem. Denote 
    \[
    L^2_0 (\mu_{x}) =\left\{ T \in L^2(\mu_{x}): \E_{z\sim\mu_{x}}[ T(z) ] =0 \right\}
    \]
    the set of zero-mean functions.  For a set of {\em test function} $\Psi \subseteq L_2(\mu_y)$ we say that $U \subseteq [P]$ is \textbf{$\Psi$-detectable} iff
\begin{equation}\label{eq:detectable}
\exists {T\!\in\!\Psi}, \; \exists {T_i\!\in\!L^2_0(\mu_x) \,\,\textrm{\rm for each}\, i\in U} \quad \textrm{such that}\quad \E_{\mu_{y,\bz}}\Big[ T (y) \prod_{i \in U} T_i(z_i) \Big] \neq 0.
\end{equation}
We denote $\cC_\Psi(\mu)$ the set of $\Psi$-detectable sets, i.e.~those sets satisfying \eqref{eq:detectable}.
\end{definition}

The set of relevant test functions depend on the query types allowed and we define:
\begin{equation}\label{eq:Psi}
\begin{aligned}
&\textrm{For } \SQ, 
&&\Psi_{\SQ} = L^2 (\mu_y) \textrm{ (i.e.~all $L^2$ functions)}. \\
&\textrm{For } \CSQ \textrm{ (recall $\cY\subseteq\R$)}, 
&&\Psi_{\CSQ} = \{ y \mapsto y \} \textrm{ (just the identity)}.\\
&\textrm{For } \DLQ_\ell, 
&&\Psi_{\DLQ_\ell} = \{ y \mapsto  \bv^\sT \nabla \ell (\bu,y) : \bu \in \cF,\bv \in V \}.
\end{aligned}
\end{equation}
While the queries of the form \eqref{eq:detectable}, where $\Psi_{\sA}$ for $ \sA \in \{\SQ, \CSQ, \DLQ_{\ell}\}$ is given by \eqref{eq:Psi}, are less general than $\phi(y,\bx)$ with $ \phi \in \cQ_{\sA}$, they can be implemented by the corresponding query types $\cQ_{\sA}$ and are sufficient for deciding between ``$S\subseteq [d]$ maps to the corresponding $U \subseteq [P]$'' or ``$S \not\subseteq S_*$''. The sets $\Psi_{\SQ}, \Psi_{\CSQ}, \Psi_{\DLQ_{\ell}}$ from \eqref{eq:Psi} used for detectibility arise naturally in the proof of the lower bounds of Theorem \ref{thm:meta_complexity_leap}.

To ease notation, for query type $\sA$, we use the shorthand $\cC_\sA:=\cC_{\Psi_\sA}$, and  
\begin{equation*}
\Leap_{\sA} (\mu) := \Leap (\cC_{\sA} (\mu)) = \Leap(\cC_{\Psi_\sA}(\mu)), \qquad \Cover_{\sA} (\mu) :=  \Cover (\cC_{\sA} (\mu)) = \Cover(\cC_{\Psi_\sA}(\mu)).
\end{equation*}
We refer to these as the \emph{$\sA$-leap exponent} and \emph{$\sA$-cover exponent} of the problem $\mu$.
\section{Main Result: Characterizing the Complexity of Learning Juntas}\label{sec:main-results}
\begin{theorem}\label{thm:meta_complexity_leap} For any junta problem $\mu$ and any loss $\ell$, there exists $C>c>0$ (that depend on $\P$,$\mu$ and the loss, but not on $d$), such that for query types $\sA \in \{\SQ,\CSQ,\DLQ_\ell\}$ with corresponding test function sets $\Psi_\sA$ as defined in \eqref{eq:Psi}:

\begin{description}
\item[Adaptive.] Let $k_* = \Leap_{\sA} (\mu)$. There exists an algorithm $\cA \in \sA(q,\tau)$ that succeeds in learning $\cH^d_{\mu}$ with $\tau = c$ and $q = C d^{k_*}$.  And if $\mu$ satisfies Assumption \ref{ass:well-behaved}, then for any $(q,\tau)$ such that $q/\tau^2 \leq c d^{k_*}$, no algorithm $\cA \in \sA(q,\tau)$ succeeds at learning $\cH^d_{\mu}$. 

\item[Non-adaptive.] Let $k_* = \Cover_{\sA} (\mu)$. There exists an algorithm $\cA \in \naA(q,\tau)$ that succeeds in learning $\cH^d_{\mu}$ with $\tau = c$ and $q = C d^{k_*}$.  And if $\mu$ satisfies Assumption \ref{ass:well-behaved}, then for any $(q,\tau)$ such that $q/\tau^2 \leq c d^{k_*}$, no algorithm $\cA \in \naA(q,\tau)$ succeeds at learning $\cH^d_{\mu}$. 
\end{description}
\end{theorem}
\begin{remark}\label{rem:(q,tau)tradeoff}
    In the positive results in Theorem \ref{thm:meta_complexity_leap}, we used all the allowed complexity to have many ($q=\Theta(d^{k_*})$) queries, and kept the tolerance constant.  More generally, it is possibly to trade off between the number of queries $q$ and tolerance $\tau$, at a cost of a log-factor: For $k_*=\Leap_{\sA}(\mu)$ and $k_*=\Cover_{\sA}(\mu)$, respectively, there exists algorithms $\cA \in \sA(q,\tau) $ and $\cA \in \naA(q,\tau)$ that learn $\cH_{\mu}^d$ for any $q \geq C \log (d)$ and $\tau \leq c$ with $q/\tau^2 \geq C d^{k_*} \log (d)$. 
\end{remark}
The proof of this theorem is deferred to Appendix \ref{app:proof_meta_complexity}. Theorem \ref{thm:meta_complexity_leap} shows that the leap and cover complexities sharply capture the scaling in $d$ of statistical query algorithms when learning $\cH^d_\mu$.
\begin{remark}
    The above upper bound uses that $\mu$ and therefore the $T,T_i$ in Definition \ref{def:leap_cover_exponent} are known. In the case when $\mu$ is unknown, one can follow a similar strategy as in \cite{damian2024computational} and randomize the transformations $T$ and $T_i$ over a sufficiently large (but finite independent of $d$) linear combination of functions in $\Psi_{\SQ}$ and $L^2_0(\mu_x)$. Under some regularity assumption on $\mu$ and $\ell$, one can show by anti-concentration that with constant probability, the expectation in condition \eqref{eq:detectable} is bounded away from $0$ by a constant independent of the dimension.
\end{remark}


\subsection{Identifiability and Relative Leap and Cover Complexities}\label{app:relative}
The $\Leap$ and $\Cover$ complexity may be infinite when $\supp(\cC_{\sA}) = \cup_{U \in \cC_{\sA}} U \subsetneq[\P]$, in which case Theorem \ref{thm:meta_complexity_leap} (correctly) implies that we cannot recover the relevant coordinates that correspond to $[\P]\setminus \supp(\cC_{\sA})$ using queries of type $\sA$ (even with perfect precision). In this case, we can characterize the complexity of recovering $\supp(\cC_{\sA})$ instead, and this is captured by the following ``relative'' complexities:
\begin{definition}\label{def:relleap relcover}
For any system of subsets $\cC \subseteq 2^{[\P]}$ the relative leap and cover complexities are:
\begin{align}
    \relLeap (\cC) &:= \min_{\substack{U_1, \dots, U_r \in \cC\\ \bigcup\limits_{i=1}^{r} U_i=\supp(\cC)}} \;\max_{i \in [r]}\; | U_{i} \setminus \cup_{j=0}^{i-1} U_{j} |,\\
    \relCover ( \cC) &:= \max_{i \in \supp(\cC)} \; \min_{ U \in \cC, i \in U} \;|U|.
\end{align}
\end{definition}
A slight variant of Theorem \ref{thm:meta_complexity_leap} can then be shown, where $\relLeap_{\sA}(\mu)$ and $\relCover_{\sA}(\mu)$ for $\sA \in \{\SQ, \CSQ, \DLQ_{\ell}\}$ characterizes the complexity of recovering $\supp(\cC_{\sA})$. 

For $\SQ$, we may have $\supp(\cC_{\SQ}) \subsetneq [P]$ only when $y$ doesn't actually depend on some of the coordinates in $[P]$ (i.e.~$y\mid \bz = y | (\bz_i)_{i\in\supp(\cC)}$).  In this case, once we recover $\supp(\cC)$ (and possibly enumerating over permutations of its coordinates) we can also recover the conditional $y|\bx$.  That is,  $\relLeap_{\SQ}$ and $\relCover_{\SQ}$ characterize the complexity of learning the distribution $\cD_{s_*}$, even if not the (unidentifiable) set $S_*$.

For $\DLQ_{\ell}$, including $\CSQ$, we may not be able to identify some coordinates even if $y$ does depend on them.  Consider for example the junta problem of size $P=2$ where $y|\bz = z_1 + \mathcal{N}(0,z_2^2)$.  Although $y$ does depend on the second coordinate $z_2$, it is not identifiable using $\CSQ$ queries,  $\supp(\cC_{\CSQ})=\{1\}$, and using $\CSQ$ queries we cannot recover the conditional distribution $\cD_{s_*}$ (even if we know the link $\mu_{y\mid\bz}$).  What we {\em can} say for $\DLQ_\ell$ after recovering $\hat{S}=\supp(\cC_{\DLQ_\ell})$, is that although we might not know $\cD_{s_*}$, we can find the $\ell$-risk minimizer
$f_* = \arg\min_{f:\cX\rightarrow\cF} \E_{(y,\bx)\sim \cD_{s_*}}\left[ \ell(f(x),y) \right]$.  That is, $\relLeap_{\DLQ_\ell}$ and $\relCover_{\DLQ_\ell}$ characterize the complexity of $\ell$-risk minimization.

\section{Relationship Between \texorpdfstring{$\SQ, \CSQ$ and $\DLQ_{\ell}$}{TEXT}}

 Obviously, $\CSQ \subseteq \SQ$, and indeed we see that $\cC_{\CSQ}\subseteq \cC_{\SQ}$ in Definition \ref{def:leap_cover_exponent} because of which $\Leap_{\CSQ} \geq \Leap_{\SQ}$ and $\Cover_{\CSQ} \geq \Cover_{\SQ}$. For binary $\cY$, these query models collapse, but otherwise there can be an arbitrary gap.
\begin{proposition}[{$\SQ$ versus $\CSQ$}]\label{prop:SQvsCSQ}
   For any $\mu$, let $\cC_{\SQ} := \cC_{\SQ} (\mu)$, and $\cC_{\CSQ}:= \cC_{\CSQ} (\mu)$. If $|\cY|= 2$ (binary output), then we always have $\cC_{\CSQ} = \cC_{\SQ}$ and $\Leap_{\SQ}=\Leap_{\CSQ}$ and $\Cover_{\SQ}=\Cover_{\CSQ}$. On the other hand if $|\cY | >2$, the $\SQ$-exponents can be much smaller than the $\CSQ$-exponents: e.g., there exist a setting with $ \Leap_{\SQ} = \Cover_{\SQ}= 1$ and $\Leap_{\CSQ} = \Cover_{\CSQ} = P$.
\end{proposition}
 
Similarly, $\DLQ_{\ell} \subseteq \SQ$ by definition, and thus, $\Leap_{\DLQ_{\ell}} \geq \Leap_{\SQ}$ and $\Cover_{\DLQ_{\ell}} \geq \Cover_{\SQ}$.
\begin{proposition}[{$\DLQ_\ell$ versus $\SQ$ and $\CSQ$}]\label{prop:DLQvsSQ_CSQ}
   Consider any $\mu=(\mu_y,\mu_{y\mid \bz})$.  
    \begin{itemize}
         \item[(a)] Let $\cY, \cF \subseteq \R$. For the squared loss $\ell:(u,y) \mapsto (u-y)^2 $, we always have $ \cC_{\DLQ_{\ell}}=\cC_{\CSQ}$, and thus, $\Leap_{\CSQ}=\Leap_{\DLQ_{\ell}}$ and $\Cover_{\CSQ}=\Cover_{\DLQ_{\ell}}$.
        \item[(b)] A sufficient condition for $\cC_{\SQ} = \cC_{\DLQ_\ell}$ is to have ${\rm span}(\Psi_{\DLQ_\ell})$ dense in $L^2_0(\mu_y)$. Conversely, if there exists nonzero $T \in L^2_0 (\mu_y)$ bounded with $T(y)$ orthogonal in $L^2 (\mu_y)$ to any functions in $\Psi_{\DLQ_\ell}$, then there exists a problem $\mu$  such that $\Cover_{\DLQ_{\ell}}=\Leap_{\DLQ_{\ell}}>\Cover_{\SQ}=\Leap_{\SQ}$.
        \item[(c)] There exists a loss $\ell$ and a junta problem $\mu$ such that $
        \Leap_{\SQ} (\mu) < \Leap_{\DLQ_\ell} (\mu) < \Leap_{\CSQ} (\mu).$
        Similarly, we can have $\Leap_{\DLQ_{\ell}} > \Leap_{\CSQ}$.     


 
    \end{itemize}
\end{proposition}

The condition in Proposition \ref{prop:DLQvsSQ_CSQ}.(b) can be seen as a universal approximation property of neural networks with activation $y \mapsto \nabla \ell (\bu,y)$ \citep{cybenko1989approximation,hornik1991approximation,sonoda2017neural}. The next lemma gives a few examples of losses with $\DLQ_\ell = \SQ$. For concreteness, we consider $\cF = \R$ and $\cY \subseteq \R$.

\begin{theorem}\label{thm:universal_loss}
    For $\ell \in \{\ell_1: (u,y) \mapsto |u-y|, \ell_{\text{hinge}}: (u,y) \mapsto \max (1 - uy,0)\} $, then $\cC_{\DLQ_\ell} (\mu) = \cC_{\SQ} (\mu)$ for all\footnote{For Hinge loss, we further restrict $\mu_y$ to measures with finite second moment so that $\Psi_\sA \subseteq L^2(\mu_y)$.} problems $\mu$. If we further assume $\cY \subseteq [-M,M]$ for  $M\geq 0$, then $\ell (u,y) = e^{-uy}$ (exponential loss) has $\cC_{\DLQ_\ell} (\mu) = \cC_{\SQ} (\mu)$ for all problems $\mu$.
\end{theorem}
The cases of $\ell_1$ and Hinge loss follow directly from universal approximation of neural networks with linear threshold activation. The proofs of the above propositions and lemma can be found in Appendix \ref{app:proof_technical}. Propositions \ref{prop:SQvsCSQ} and \ref{prop:DLQvsSQ_CSQ} combined with Theorems \ref{thm:universal_loss} and \ref{thm:meta_complexity_leap} directly imply a number of separation results between adaptive and non-adaptive algorithms and between different loss functions. See examples \eqref{eq:examples_intro} in the introduction, and further examples in Appendix \ref{app:proof_technical}.

\section{Gradient Descent on Neural Networks}\label{sec:gd-on-NN}
The goal of this section is to connect the complexity of $\DLQ$ to gradient descent on standard neural networks. We focus on the simple case of $\bx\sim \Unif (\{+1,-1\}^d)$ uniformly distributed on the hypercube, and $\cY \subseteq \R$ and $\cF = \R$. In this setting, condition \eqref{eq:detectable} in Definition \ref{def:leap_cover_exponent} simplifies to: there exists $T \in \Psi_{\sA}$ such that $\E_{\mu_{y,\bz}} \left[T(y) \prod_{i\in U} z_i\right] = \E_{\mu_{y,\bz}} \left[T(y) \chi_{U}(\bz)\right] \neq 0$, where $\chi_{U} (\bz) := \prod_{i\in U} z_i$ denote the standard Fourier-Walsh basis. In particular, the set $\cC_{\CSQ}$ contains exactly all non-zero Fourier coefficients of $h_* (\bz) := \E[y|\bz]$, and we recover the leap exponent of \cite{abbe2023sgd} as discussed in the introduction.

Fix a link distribution $y| \bz \sim \mu_{y|\bz}$ and consider data distribution $(y,\bx) \sim \cD^d_{s_*} \in \cH^d_{\mu}$ with $\mu =(\mu_x,\mu_{y|\bz})$ and (unknown) $s_* \in \perm{[d]}{\P}$. We consider learning using a two-layer neural network with parameters $\bTheta \in \R^{M(d+2)+1}$ and an activation $\sigma : \R \to \R$:
\begin{equation}\label{eq:NN_original}
f_{\NN} ( \bx ; \bTheta) = c+\frac{1}{M} \sum_{j \in [M]} a_j \sigma (\< \bw_j,\bx\> + b_j), \qquad c, a_j, b_j, \in \R , \quad \bw_j  \in \R^d, \text{ for } j \in [M] \tag{NN1}
\end{equation}
For convenience, we will reparametrize the neural network for $\bTheta = (\btheta_j)_{j \in [M]}\in \R^{M(d+3)}$ with $\btheta_j = ( a_j,\bw_j, b_j , c_j) \in \R^{d +3}$ and
\begin{equation}\label{eq:NN_param}
f_{\NN} ( \bx ; \bTheta) = \frac{1}{M} \sum_{j \in [M]}  \sigma_* (\bx;\btheta_j), \qquad \sigma_* (\bx;\btheta_j) = a_j \sigma ( \<\bw_j,\bx\> + b_j ) + c_j. \tag{NN2}
\end{equation}
Indeed, the two neural networks \eqref{eq:NN_original} and \eqref{eq:NN_param} remain equal throughout training under the initialization $c_1 = \ldots = c_M = c$.

\paragraph{} We train using online SGD with a loss function $\ell : \R \times \cY \to \R_{\geq 0}$, i.e. SGD on $ \E_{(\bx,y) \sim \cD^d_{s_*}}[\ell ( f (\bx),y )]$. More specifically, we train the parameters $\bTheta$ using batch-SGD with loss $\ell$ and batch size $b$ from initialization $(\btheta_j)_{j \in [M]} \iid \rho_0$ given by 
\begin{equation}\label{eq:init-main}
    (a^0,b^0,\sqrt{d} \cdot \bw^0, c^0) \sim \mu_a \otimes \mu_b \otimes \mu_w^{\otimes d} \otimes \delta_{c = \overline{c}}.
\end{equation}
At each step, given samples $(\{(\bx_{ki},y_{ki}): i \in [b]\})_{k \geq 0}$, the weights are updated using
\begin{equation}\label{eq:batch_SGD}
\btheta_j^{k+1} = \btheta_j^k - \frac{\eta}{b} \left( \sum_{i \in [b]}  \ell' (f_\NN (\bx_{ki};\bTheta^t),y_{ki}) \nabla_{\btheta} \sigma_* ( \bx_{ki};\btheta_j^k ) +\lambda \,\btheta_j^k \right), \tag{$\ell$-bSGD}
\end{equation}
where  $\eta$ is the step-size and we allow for a $\ell_2$ regularization with parameter $\lambda \in \R_{+}$. Recall that $\ell'(u,y) \in \partial_{1} \ell(u,y)$ is the defined derivative of $\ell(\cdot,y)$ at $u\in \R$. We define the test error
\begin{equation}\label{eq:risk_functional}
\cR (f) = \E_{\cD_{s_*}^d} \left[  \ell (f(\bx), y)\right],
\end{equation}
and further introduce the excess test error $\ocR (f) = \cR(f) - \inf_{\scriptscriptstyle \bar{f}:\{\pm 1\}^d \to \R} \cR(\bar{f})$.

\paragraph{Limiting dimension-free dynamics:} In the junta learning setting, when $y$ only depends on $P \lll d$ coordinates, following \citet[Secion 3]{abbe2022merged}, the SGD dynamics \eqref{eq:batch_SGD} concentrates on an effective \textit{dimension-free} (DF) dynamics as $M,d \to \infty$ and $\eta \to 0$. This equivalence holds under a certain assumption on the loss function, and other assumptions on the initialization and activation that are similar to the setup of \cite{abbe2022merged} (see Appendix \ref{app:DF_PDE} for details). This limiting dynamics corresponds to the gradient flow on $\cR(f)+(\lambda/2) \cdot \int \| \btheta \|_2^2 \rho (\de \btheta)$ of the following effective \textit{infinite-width} neural network (recall $\bz \in \R^P$ is the support)
\begin{equation}
    f_{\NN} (\bz ; \barrho_t) = \int \osigma_* (\bz;\barbtheta^t) \barrho_t (\de \barbtheta^t), \qquad  \osigma_* (\bz;\barbtheta^t) = c^t + a^t \E_G [\sigma ( \< \bz , \bu^t \> +s^tG + b^t)],
\end{equation}
where $G \sim \normal (0,1)$ and $\barrho_t \in \cP(\R^{P+4})$ is the distribution over $\barbtheta^t = (a^t,b^t,\bu^t,c^t,s^t)$ with $\bu^t \in \R^P$. This distribution evolves according to the following PDE
\begin{align}\label{eq:DF-dynamics}
       \partial_t \barrho_t =&~ \nabla_{\barbtheta} \cdot\left(\barrho_t \cdot \nabla_{\barbtheta} \psi (\barbtheta , \barrho_t) \right), \nonumber\\
    \psi (\barbtheta , \barrho_t) =&~  \E_{\mu_{y,\bz}} \left[ \ell'(f_{\NN}(\bz ; \barrho_t), y) \osigma_* ( \bz ; \barbtheta )\right] + \frac{\lambda}{2} \norm{\barbtheta}_2^2, \tag{DF-PDE}
\end{align}
from initialization 
\[
\barrho_0 :=  \mu_a \otimes \mu_b \otimes \delta_{\bu^0 = \bzero} \otimes \delta_{c^0 = \overline{c}} \otimes \delta_{s^0 = \sqrt{m_2^w}},
\]
where $m_2^w=\E_{W\sim \mu_{w}}[W^2]^{1/2}$ is the second moment of $\mu_w$. Following a similar argument as in \citet{abbe2022merged}, we have the following non-asymptotic bound between \eqref{eq:batch_SGD} and \eqref{eq:DF-dynamics} dynamics: with probability at least $1 - 1/M$, we have for all integers $k \leq T/\eta$
    \begin{equation}\label{eq:DF-SGD-approx}
      \hspace{-7pt}  \left| \cR (f_{\NN} (\cdot ; \bTheta^k )) - \cR (f_{\NN} (\cdot ; \rho_{\eta k} ))\right| \leq C \left\{ \sqrt{\frac{P}{d}} + \sqrt{\frac{\log M}{M}} + \sqrt{\eta}\sqrt{\frac{d + \log M}{b} \vee 1}\right\}.
    \end{equation}
  See Theorem \ref{thm:SGD_to_DF} for a formal statement. 
   Hence, the batch-SGD dynamics \eqref{eq:batch_SGD} is well approximated by the DF dynamics \eqref{eq:DF-dynamics} for number of SGD-steps $\Upsilon= O(d/b)$, step size $\eta = O(b/d)$ and $M = \Omega(1)$. That is \eqref{eq:DF-dynamics} tracks the \eqref{eq:batch_SGD} dynamics in the linear scaling of total number of samples $b\Upsilon = O(d)$. 

\paragraph{Dimension-free dynamics' alignment with the support. }
In the above limiting regime, $\Leap_{\DLQ_\ell}$ crisply characterizes \eqref{eq:DF-dynamics} dynamics' alignment with the support. 
\begin{theorem}[Informal version of Theorem \ref{thm:DF_PDE_Leap}]\label{thm:DF_PDE_Leap_informal}
If $\Leap_{\DLQ_\ell}=1$ then for some time $t$, the output of the model at time $t$ of \eqref{eq:DF-dynamics} dynamics depends on all  coordinates $z_i$. On the other hand, if $\Leap_{\DLQ_\ell} > 1$, then there exists a coordinate $i$ such that for any time $t \geq 0$, the output of the model at time $t$ of \eqref{eq:DF-dynamics} dynamics does no depend on $z_i$.
\end{theorem}
This establishes that the ability of \eqref{eq:DF-dynamics} dynamics (comparable to \eqref{eq:batch_SGD} in the linear scaling) to learn all relevant coordinates depends on $\Leap_{\DLQ_\ell}(\mu)=1$. That is if $\Leap_{\DLQ_\ell} > 1$, then \eqref{eq:DF-dynamics} remains stuck in a suboptimal saddle subspace. On the other hand, if $\Leap_{\DLQ_\ell} = 1$, then \eqref{eq:DF-dynamics} dynamics escapes this subspace and the weights align with the entire support.


\paragraph*{Learning of $\Leap_{\DLQ_\ell} = 1$ with finite SGD.} Showing directly that \eqref{eq:DF-dynamics} dynamics indeed reach a near global minimizers of the test error remains challenging. Alternatively, we show that a specific layer-wise training dynamics similar to \cite{abbe2022merged} achieves a vanishing excess error for $\Leap_{\DLQ_\ell}(\mu) = 1$ settings in the linear scaling of samples. 

Roughly speaking, we train the first layer for $k_1 = P$ steps and then the second layer weights for $k_2 = O_d(1)$ steps using batch-SGD with batch size $b=O_d(d)$, both for a loss $\ell$. We consider a polynomial activation $\sigma(x)=(1+x)^L$ of degree $L \geq 2^{8P}$. The most notable difference from \cite{abbe2022merged} is that we further slightly perturb step-sizes for each coordinate $\eta^{w_i} = \eta \kappa_i$ with $\kappa_i \in [1/2,3/2]$, and denote $\bkappa = (\kappa_1,\ldots,\kappa_d) \in \R^d$ for the first layer training. This perturbation is necessary to break possible coordinate symmetries; see Remark \ref{rem:coordinate-symmetries}. 

\begin{theorem}[Informal version of Theorem \ref{thm:leap-one-ub}]\label{thm:learning_leap_1_informal}
For a convex and analytic loss $\ell$, almost surely over the perturbation $\bkappa$ and the initial bias $\bar{c}\in \R$ the following holds. If $\Leap_{\DLQ_\ell} (\mu) = 1$,  then the above layer-wise SGD training dynamics with total sample size $n = \Theta_d ( d)$ and $M = \Theta_d (1)$ achieves excess test error $\ocR  (f_{\NN} (\cdot;\bTheta^{k_1+k_2}))=o(1)$ with high probability.  
\end{theorem}
The formal statement and the precise training specifications can be found in Appendix \ref{app:DF_PDE}.  This result generalizes \citet[Theorem 9]{abbe2022merged} beyond squared loss.
\begin{remark}\label{rem:coordinate-symmetries}
A slight coordinate-wise perturbation in the step-sizes for the first layer training is necessary to break the potential coordinate symmetries in the output $y$---see discussion in \citet[Appendix A]{abbe2022merged}. This can be removed by either stating the theorem for all but some measure zero set of Leap-1 functions as in \cite{abbe2022merged}, or by studying the dynamics for $O(\log d)$ steps.
\end{remark}
Finally, the query complexity of $b$-batch SGD on $M$ neurons for $\Upsilon$ SGD-steps is $\Tc = \Theta (b M \Upsilon d)$. The above theorems show that for $\Leap_{\DLQ_\ell} (\mu) = 1$, $\Upsilon = \Theta (d/b)$ steps with $M = \Theta (1)$ neurons---and therefore $\Tc = \Theta (d^2)$---suffices to learn the support and minimize the excess test error. Furthermore, for $\Leap_{\DLQ_\ell} (\mu) > 1$ and neural networks trained in the mean-field regime, $\Upsilon =\Theta(d)$ (and therefore $\Tc = \Theta(d^2)$) is not enough. The conjectural picture is for $k_* := \Leap_{\DLQ_\ell} (\mu)$, learning the junta problem with online SGD on loss $\ell$ requires
\[
\begin{aligned}
    &~ k_* = 1: \qquad &~ n =\Theta (d), \qquad &~ \Tc = \Theta (d^2), \\
    &~ k_* = 2: \qquad &~ n = \Theta (d \log d ), \qquad &~ \Tc = \tTheta (d^2), \\
    &~ k_* >2 : \qquad &~ n = \tTheta (d^{k_* - 1}), \qquad &~ \Tc = \tTheta ( d^{k_*}).
\end{aligned}
\]
Partial evidence was provided towards this conjecture for the square loss in \cite{abbe2023sgd,bietti2023learning}: they showed that learning in this setting happens through a saddle-to-saddle dynamics, where each time the neural network needs to align to $k>1$ new coordinates, a saddle appears and SGD requires $\tTheta(d^{k-1})$ steps to escape. 

\subsection{Numerical illustration}
\begin{figure}[!h]
\small    \centering
    \subfloat[\centering \small $\ell(u,y)=(u-y)^2$]{{\includegraphics[scale=0.36]{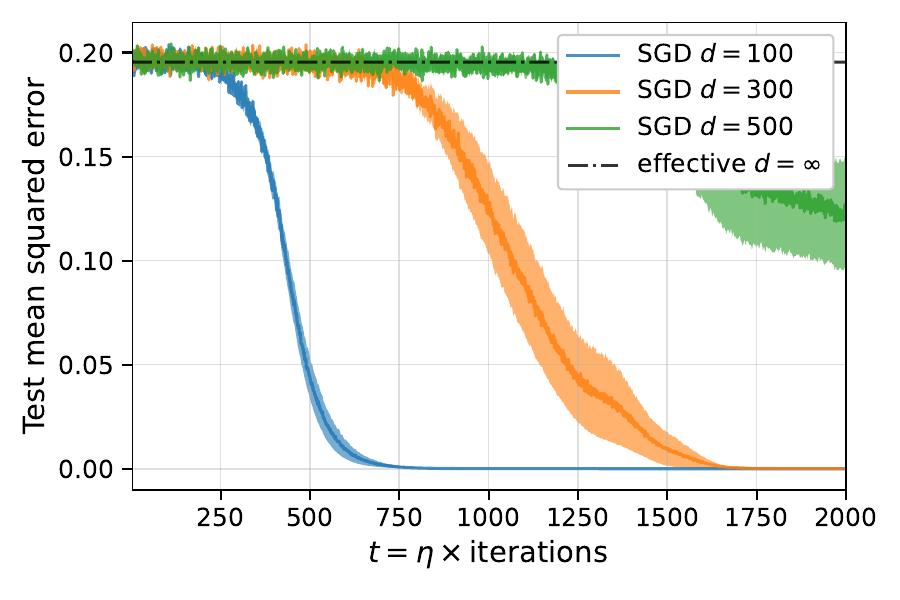} }}%
    \subfloat[\centering \small $\ell(u,y)=|u-y|$]{{\includegraphics[scale=0.36]{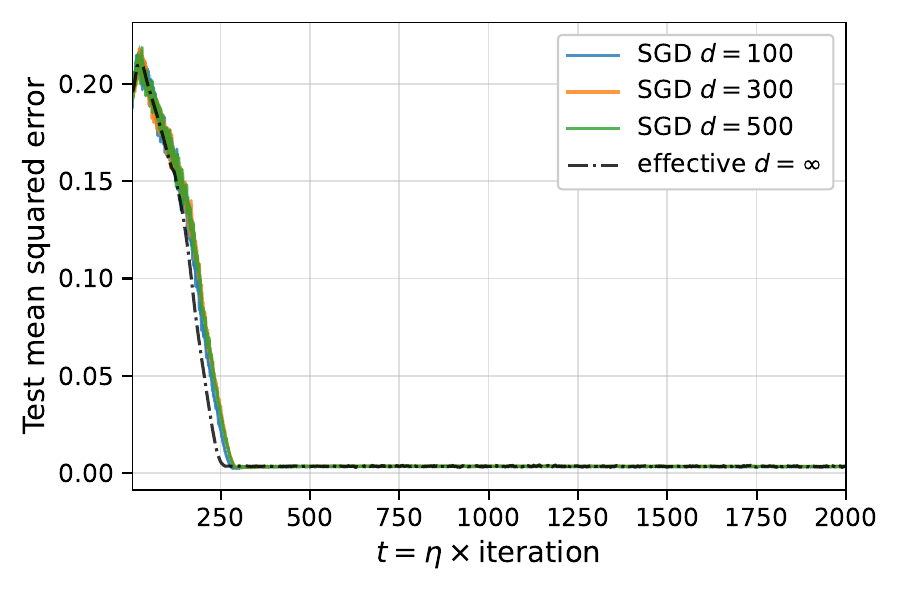} }}%
        \subfloat[\centering \small $\ell(u,y)=(u-y)^2+|u-y|^3$]{{\includegraphics[scale=0.36]{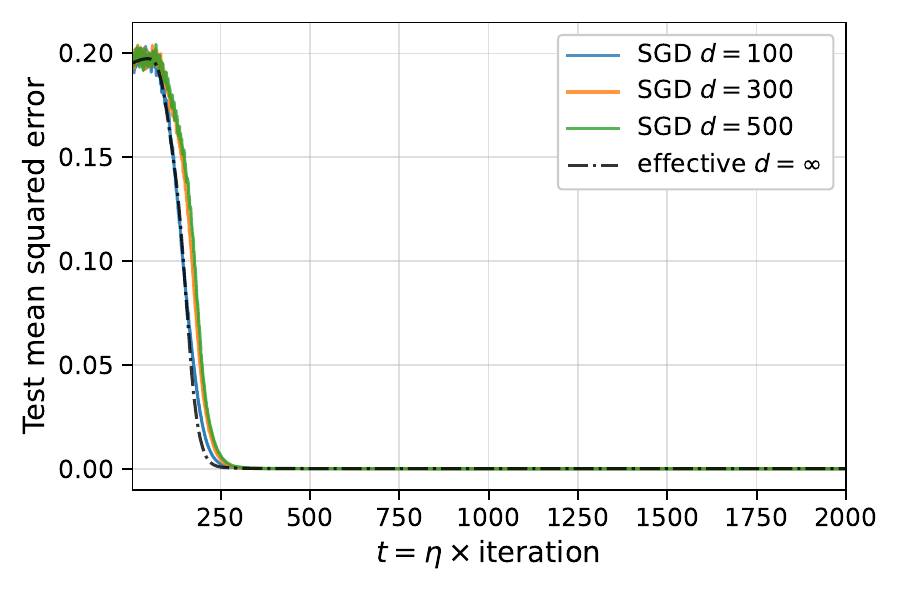} }}%
    \caption{\small The function $h_*(\bz)$ in \eqref{eq:example-simulation} has $\Leap_{\CSQ}=3$ but $\Leap_{\SQ}=1$. For the squared loss (left plot), \eqref{eq:DF-dynamics} remains stuck at initialization (no learning), and to escape the saddle, SGD requires a number of iterations that increases faster than $O(d)$. For the absolute loss (center plot) or the other loss (right plot), we have $\Leap_{\DLQ_{\ell}}=\Leap_{\SQ}=1$, and the SGD dynamics learns in $\Theta(d)$ steps and \eqref{eq:DF-dynamics} learns in $O(1)$ continuous time.\vspace{-3mm} }%
    \label{fig:sqvsother-comparison}%
\end{figure}

We consider a function similar to $y_2$ in \eqref{eq:examples_intro} with $\Leap_{\CSQ}=3$ but $\Leap_{\SQ}=1$. Specifically, we set $P=4$ and $\cC=\{ \{1,2,3\}, \{1,2,4\},\{1,3,4\},\{2,3,4\} \}$, and define $y = h_*(\bz)$ where
\begin{equation}\label{eq:example-simulation}
  h_*(\bz)=\sum_{U\in \cC} \hat{h}_*(U) \chi_{U}(\bz), \; \text{where }\, \hat{h}_*(U) \sim \Unif([-2,2]) \text{ for all } U\in \cC.  
\end{equation}
We train with online 1-batch SGD \eqref{eq:batch_SGD} on a two-layer net with different loss functions (without any regularization) and stepsize $\eta \propto 1/d$, where we consider ambiant dimensions $d \in \{100,300,500\}$. In Figure \ref{fig:sqvsother-comparison}, we plot the test mean squared error versus $\eta \times \text{SGD-iterations}$ (thus also scaled with $1/d$), over 10 trials. Additionally, we also plot the continuous time \eqref{eq:DF-dynamics} in (dashed black line) that corresponds to the limit $d \rightarrow \infty$. 

\section{Conclusion and Outlook}\label{sec:discussion}

In this paper, we considered learning juntas over general product distributions with statistical query algorithms. To capture learning with gradient evaluations over a general loss and arbitrary model, we introduced \textit{Differentiable Learning Queries} ($\DLQ_\ell$), which can be seen as a generalization of correlation statistical queries beyond squared loss. We then showed that the complexity of learning juntas with either $\SQ$, $\CSQ$, or $\DLQ_\ell$ algorithms is sharply characterized in terms of a \textit{leap exponent} (adaptive queries) or a \textit{cover exponent} (non-adaptive queries). These exponents are defined in terms of a minimal combination of detectable sets to cover the support, where the system of detectable sets depends on the allowed queries. In general, the leap and cover exponents for different losses are not comparable. However, we identify ``generic'' losses, including $\ell_1$, where $\DLQ_\ell$ algorithms are as powerful as $\SQ$ algorithms for learning juntas. We further showed that $\DLQ_\ell$ can indeed capture the complexity of learning with SGD in the case of data on the hypercube.

\paragraph*{Worst-case v.s.~One-pass v.s.~Multi-pass SGD.} $\DLQ_\ell$ (like $\SQ$) is defined in terms of worst-case noise. It is well understood that worst-case noise is theoretically very different from estimating population gradients based on samples (either independent samples as in one-pass (S)GD, or with repeated use of samples as in full-batch or multi-pass) when highly specialized models are allowed  \citep{abbe2021power}.  However, we expect $\DLQ_\ell$ to still capture the complexity of one-pass SGD in many settings of interest---with ``regular'' models---such as in Section \ref{sec:gd-on-NN}. With sample reuse (e.g.~multi-pass) the situation is more complex: \cite{dandi2024benefits} showed that two steps of full-batch gradient descent with square loss goes beyond $\CSQ$. Heuristically, and in light of our work, two steps on the same batch can be seen as a single gradient evaluation on a modified loss, thus going beyond $\CSQ=\DLQ_{\ell_{sq}}$, but remaining inside $\DLQ_{\tilde{\ell}}\subseteq\SQ$ for some perturbed non-quadratic loss $\tilde{\ell}$.  Indeed, we expect generally that multi-pass SGD on a ``regular'' model will remain in $\SQ$.

\paragraph{Multi-index and beyond.} In this paper, we focused on learning sparse functions.  We hope the modular nature of our analysis framework (defining detectable sets in terms of test functions, and leap and cover complexities of set systems), our definition of $\DLQ$, and the distinctions we emphasize between $\CSQ$, $\SQ$ and $\DLQ$ and between adaptive and non-adaptive complexities, will be helpful in guiding and contextualizing analysis in other settings such as learning single-index or multi-index functions \cite[e.g.][]{refinetti2021classifying,mousavi2022neural,abbe2022non,damian2022neural,bietti2022learning,bietti2023learning,damian2024smoothing}. For example, the information exponent for single-index \citep{arous2021online} can be seen as analogous to our $\CSQ$-cover exponent, the generative exponent for single-index \citep{damian2024computational} as analogous to our $\SQ$-cover exponent, and the isoLeap exponent for multi-index \citep{abbe2023sgd,dandi2023learning,bietti2023learning} as analogous to our $\CSQ$-leap exponent.  It would be interesting to obtain a unified understanding of these specialized treatments and extend our general framework to multi-index models, and also to learning under other invariances beyond permutations and rotations.

In our setup, we emphasize generic input and output spaces, without a field structure.  This emphasizes that when learning juntas,  polynomials or degree of input or output coordinates is irrelevant.  Defining multi-index models and introducing rotational invariance necessitates a field structure and gives rise to the relevance of polynomial degrees and decomposition.

An important point is that when considering permutation (as in juntas) vs.~rotational (as in multi-index models) invariance, one must consider not only the invariance structure of the target function class, but also the input distribution (i.i.d.~coordinates as in our case, or more generally exchangeable vs.~spherical) and learning rule.  E.g.,~learning a parity over input coordinates requires only $\Theta (\log d)$ samples, but a rotationally equivariant algorithm effectively learns parities also over rotated axis, which requires $\Omega (d)$ samples \citep{glasgow2023sgd} (and thus $\Omega(d^2)$ runtime). This also explains the need for $\Theta(d)$ steps and thus $\Theta (d^2)$ runtime to learn leap-1 functions using SGD on a rotationally invariant neural net in Section \ref{sec:gd-on-NN}.  In order to break the $\Omega (d)$-sample lower bound, we need to break the rotation-equivariance, e.g.~using sparse initialization and $\ell_1$ regularization, which indeed can achieve $\Theta(\log (d))$ sample complexity.

\bibliographystyle{plainnat}
\bibliography{bibliography}
\clearpage
\appendix
\section{Additional details}

\subsection{Property of SQ Leap and Cover Complexities}\label{app:discussion_Leap}

We show a useful property of the $\SQ$-exponent. This characterization is similar to the definition of the generative exponent for single-index models on Gaussian data in \cite[Definition 2.4]{damian2024computational}, while the detectable set Definition \ref{def:leap_cover_exponent} is similar to their variational representation \citep[Proposition 2.6]{damian2024computational}.

\begin{proposition}[Property of $\SQ$-detectable sets] \label{prop:properties-SQ_exponent} $U \in \cC_{\SQ}$ if and only if there exists $T_i \in L^2_0 (\mu_x)$ for all $i\in U$ such that $\| \xi_{U} \|_{L^2 (\mu_y)} >0 $ where
        \[
        \xi_U (y) = \E_{\mu_{y,\bz}}\Big[ \prod_{i \in U} T_i(z_i)  \Big\vert y\Big].
        \]

\end{proposition}

\begin{proof}[Proof of Proposition \ref{prop:properties-SQ_exponent}]
  In the forward direction, consider any $U \in \cC_{\SQ}$, then by Definition \ref{def:leap_cover_exponent}, there exists $T_i \in L^2_0(\mu_x)$ and $T \in L^2(\mu_y)$ such that
   $$0 \neq \E_{\mu_{y,\bz}} \left[T(y) \prod_{i \in U} T_i(z_i)\right] =\E_{\mu_y}\left[ T(y) \xi_U(y)\right],$$
  where we defined $\xi_U(y):=\E_{\mu_{y,\bz}} \left[ \prod_{i \in U} T_i(z_i) \mid y \right]$. We then have
   \begin{align*}
       0 < | \E_{\mu_y}\left[ T(y) \xi_U(y)\right] | \leq \norm{T}_{L^2(\mu_y)} \norm{\xi_U}_{L^2(\mu_y)}, \text{ by Cauchy-Schwarz inequality.}
   \end{align*}
   Therefore, we conclude that $\norm{\xi_U}_{L^2(\mu_y)} >0$ as desired. In the opposite direction, let us consider $U\subseteq [P]$ such that there exists $T_i\in L^2_0(\mu_x)$ for every $i \in U$ such that $\norm{\xi_U}_{L^2(\mu_y)} > 0$, where 
    $$\xi_{U}(y)= \E_{\mu_{y,\bz}} \left[ \prod_{i \in U} T_i(z_i) \mid y\right].$$
    Then let $T(y):=\xi_U(y)$. It is straightforward to verify that
    $$\norm{T}^2_{\mu_y}=\E_{\mu_y} \left[ T(y)^2\right]=\E_{\mu_y} \left[ \E_{\mu_{y,\bz}} \left[ \prod_{i \in U} T_i(z_i) \mid y\right]^2\right] \leq \E_{\mu_{y,\bz}} \left[ \prod_{i \in U} T_i(z_i)^2 \right] < \infty.$$
    We verified $T(y) \in L^2(\mu_y)$. Moreover,
    \[
    \begin{aligned}
    \E_{\mu_y} \left[ T(y) \prod_{i \in U} T_i(z_i)\right]=&~\E_{\mu_y} \left[ T(y) \E_{\mu_{y,\bz}} \left[ \prod_{i \in U} T_i(z_i) \mid y\right]\right] \\
    =&~ \E_{\mu_y} \left[\xi_U(y) \xi_U(y)\right] =\norm{\xi_U}^2_{L^2(\mu_y)} \neq 0.
    \end{aligned}
    \]
    Therefore, by Definition \ref{def:leap_cover_exponent}, we obtain $U \in \cC_{\SQ}$.
\end{proof}

\section{Proof of Theorem \ref{thm:meta_complexity_leap}}
\label{app:proof_meta_complexity}

\subsection{Preliminaries}

\paragraph*{Tensor basis of $L^2(\cX^k,\mu_{x}^k)$:} Since we assumed that $(\cX,\mu_x)$ is a Polish probability space, $L^2 (\cX,\mu_{x})$ is separable \citep{dudley2018real}. Consider $\{\psi_i\}_{i \in \cI}$, $\cI \subseteq \naturals$, an orthonormal basis of $L^2 (\cX ,\mu_{x})$ such that $\psi_0 = 1$ without loss of generality. In particular, the space $L^2(\cX^k,\mu_{x}^k)$ admits the following tensor basis
\begin{equation}\label{eq:tensor_basis}
\{ \psi_{\bi} : = \psi_{i_1} \otimes \psi_{i_2} \otimes \ldots \otimes \psi_{i_k} \; : \; \bi = (i_1, \ldots , i_k ) \in \cI^k\}.
\end{equation}
We will denote $\supp (\bi) = \{ j \in [d]: i_j >0\}$.

\paragraph*{Distributions $\nu_S^\sigma$:} For any non-repeating sequence $\sigma \in \perm{[d]}{P}$ of length $P$, a subset $U \subseteq [P]$, and a system of subsets $\cC = \{ U_1, \ldots , U_m\} \subseteq 2^{[P]}$, we denote $\sigma(U):= \{\sigma(i):i \in U\}$ and $\sigma (\cC):= \{ \sigma(U_1), \ldots, \sigma(U_m)\}$. The following analysis holds for fix $\mu$ and hence we will often omit the subscript $\mu$ by denoting $\cD^{d}_{\mu,\sigma} =\cD^{d}_{\sigma}$ for any $\sigma \in \perm{[d]}{\P}$. For $\sigma=\Id$, i.e. $\sigma(1)=1,\dots, \sigma(P)=P$, we will further omit writing $\Id$ and denote $\cD^d=\cD^d_{\Id}$. For $U \subseteq [P]$, $\mu_{y,\bz,U}^0 \in \cP (\cY \times \cX^P)$ is the distribution of $(y_U,\bz)$ corresponding to decoupling $y$ from $(z_i)_{i\not\in U}$ in $\mu_{y,\bz}$ following Eq.~\eqref{eq:def_cD_S}, i.e., $(y_U, (z_i)_{i \in U}) \sim \mu_{y,\bz,U}$ and $(z_i)_{i \not\in U} \sim \mu_{x}^{P-|U|}$ independently. 

For clarity, for every $U \subseteq [P]$ and $\sigma \in \perm{[d]}{P}$, 
\begin{equation}
\nu^\sigma_U := [\Id \otimes \sigma]_{\#} [\mu^0_{y,\bz,U} \otimes \mu_{x}^{d-P}],
\end{equation}
where we abuse the notation and view $\sigma$ as a permutation in $\Pi_{d}$, by extending the sequence $\sigma \in \perm{[d]}{P}$ to an ordering of irrelevant coordinates, as notice that it yields the same $\nu_U^{\sigma}$. In words, $\nu_U^\sigma$ is the distribution of $(y,\bx)$ where we decoupled $y$ from $(x_{\sigma(i)})_{i \not\in U}$ in $\cD^d_{\sigma}$. For $\sigma = \Id$, we denote $\nu_U = \nu_U^\Id$. For $U = [P]$, we simply denote $\nu^\sigma := \cD^d_{\sigma}$. Note that $\nu_0 := \nu_{\emptyset} = \mu_y \otimes \mu_x^d$ and $\nu_U^\sigma = \nu_U$ if $\sigma(i) = i$ for all $i \in U$.

\paragraph{Basic properties under Assumption \ref{ass:well-behaved}.} Recall that our lower bounds in Theorem \ref{thm:meta_complexity_leap} are under Assumption \ref{ass:well-behaved}. It states that the junta problem $\mu=(\mu_x,\mu_{y\mid \bz})$ is such that Eq.~\eqref{eq:def_ocP_P} holds. 
An immediate consequence of this assumption, using Jensen's inequality is that
\begin{equation}\label{eq:L2_Radon_proof}
\frac{\de \mu_{y,\bz,U}^0}{ \de \mu_y \otimes \mu_x^P } \in L^2 ( \mu_y \otimes \mu_x^P).
\end{equation} 
\vspace{-6mm}
\begin{align*}
\text{This follows from } \hspace{3mm}  \left\lVert \frac{\de \mu_{y,\bz,U}^0}{ \de \mu_y \otimes \mu_x^P } \right\rVert^2_{L^2(\mu_y \otimes \mu_{x}^P)} \hspace{-5mm}\leq\hspace{3mm}  \left \lVert \frac{\de \mu_{y,\bz}}{ \de \mu_{y} \otimes \mu_{x}^P } \right \rVert_{L^2(\mu_y \otimes \mu_{x}^P)}^2 < \infty,
\end{align*}
where the first inequality follows from Jensen's inequality and the second by \eqref{eq:def_ocP_P}.

Another consequence of this assumption is that, for any $U \subseteq [P]$ and $\sigma \in \perm{[d]}{P}$, we have $L^2(\nu_0) \subseteq L^1(\nu_{U}^{\sigma})$. This is because for any $\phi \in L^2(\nu_0)$
\begin{equation}
\norm{\phi}_{L^1(\nu_{U}^{\sigma})}=\E_{(y,\bx)\sim \nu_U^\sigma} \left[ | \phi (y, \bx) |\right] = \E_{(y,\bx)\sim \nu_0} \left[  \frac{\de \nu_{U}^\sigma}{\de \nu_0} (y,\bx) \cdot | \phi (y, \bx) |\right] \leq \left\| \frac{\de \nu_{U}^\sigma}{\de \nu_0}\right\|_{L^2 ( \nu_0)} \hspace{-2mm}\norm{\phi}_{L^2(\nu_0)} < \infty,
\end{equation}
where we used Cauchy-Schwarz inequality and Eq.~\eqref{eq:L2_Radon_proof}.

\paragraph*{Algorithms in $\sA$:} Throughout the proof, we consider $\sA \in \{ \SQ,\CSQ,\DLQ_\ell\}$ and $\cC_\sA := \cC_{\sA} (\mu)$ defined in Definition \ref{def:leap_cover_exponent}. First, observe that one can alternatively define $\cQ_{\SQ}=L^2(\nu_0)$ because for any measurable $\phi \notin L^2(\nu_0)$, the received query response can infinite by \eqref{eq:SQ_def} and the learner does not gain any information. Therefore, it suffices to prove lower bound against the query set $\cQ_{\SQ}=L^2(\nu_0)$. For the same reason, consider the allowed query sets, $\cQ_{\CSQ} \subseteq L^2(\nu_0)$ containing queries of the form $\phi (y,\bx) = y \tilde \phi (\bx)$, and $\cQ_{\DLQ_{\ell}} \subseteq L^2(\nu_0)$ with queries of the form
\begin{equation}
\phi(y,\bx) = \left[\frac{\de}{\de \omega} f(\bx,\omega) \right]_{\omega=0}^\sT \nabla \ell ( f (\bx, 0),y ).
\end{equation}

We claim that, for $\bi \in \cI^d$ and $\psi_{\bi}$ defined in Eq.~\eqref{eq:tensor_basis} and any $\phi \in \cQ_\sA$, Definition \ref{def:leap_cover_exponent} implies that if $\supp(\bi) \not\in \sigma (\cC_\sA)$, then almost surely over $\tbx \sim \mu_x^d$
\begin{equation}\label{eq:partial_exp_zero}
\E_{(y,\bx) \sim \nu^\sigma} \left[\phi(y,\tbx) \psi_{\bi} (\bx) \right] = 0.
\end{equation}
This follows from Definition \ref{def:leap_cover_exponent} of $\Psi$-detectibility with Eq.~\eqref{eq:Psi}, and using the fact that $\psi_\bi$ is the product of $\psi_{i_j} (x_j)$ with $\psi_{i_j} \in L^2_0 (\mu_{x})$ for $j \in \supp(\bi)$. More specifically, since $\phi \in L^2(\nu_0)$, the function $\phi(\cdot, \tbx) \in L^2 (\mu_y)$ almost surely over $\tbx \sim \mu_x^d$, and  In particular, almost surely over $\tbx \sim \mu_{x}^d$:
\begin{align*}
    & \text{For any $\phi \in \cQ_{\SQ}$}, && \text{we have } \phi(\cdot, \tbx) \in L^2(\mu_y)=\Psi_{\SQ},\\
    & \text{For any $\phi \in \cQ_{\CSQ}$}, && \text{we have } \phi(\cdot, \tbx)=y \Tilde{\phi}(\Tilde{\bx}) \text{ (i.e. a scaled identity),}\\
    & \text{For any $\phi \in \cQ_{\DLQ_{\ell}}$}, && \text{we have } \phi(\cdot, \tbx) \in \Psi_{\DLQ_{\ell}} \text{ (i.e. $\phi(\cdot,\tbx )$ of the form $\bv^\sT \nabla \ell(\bu,y)$).}\\
\end{align*}
\subsection{Theorem \ref{thm:meta_complexity_leap}.(a): lower bound for adaptive queries}

\paragraph*{Definition of set $\cS_*$:} Let $k_{*}= \Leap (\cC_{\sA})$. Define $\cC_{*} \subset \cC_{\sA}$ to be the maximal subset of $\cC_{\sA}$ such that $\Leap (\cC_{*})=k_*-1$, and $\cS_{*}= \cup_{U \in \cC_{*}} U$. In words, $\cS_*$ is the collection of all coordinates in the support that can be reached by adding at most $k_*-1$ coordinates at a time. In particular, by definition of $\cC_*$, we have $|U \setminus \cS_{*}| \geq k_*$ for every $U \in \cC_{\sA} \setminus \cC_{*}$. Denote $r = | \cS_*|$ and without loss of generality, assume that $\cS_* = \{1, \ldots , r\}$. We decompose the covariate into $\bx= (\bx_{\cS_*}, \bx_{\cS_*^c})$ where $\cS_*^c = [d] \setminus \cS_*$. Using these notations, $(y_{\cS_*},\bx) \sim \nu_{\cS_*}^\sigma$ has 
$(y_{\cS_*},\bx_{\sigma(\cS_*)}) \sim \cD_{y,\bz,\cS_*}$ and $\bx_{\sigma(\cS_*^c)} \sim \mu_{x}^{d-r}$ independently.

\paragraph*{Definition of $\cH_{\mu,\cS_*}^d$:} We introduce $\perm{[d]}{P,\cS_*} \subseteq \perm{[d]}{P}$ the subset of non-repeating sequence of length $[P]$ from elements $[d]$ such that $\sigma (i) = i$ for all $i \in \cS_*$, and $\cH_{\mu,\cS_*}^d \subseteq \cH_{\mu}^d$ the subset of hypotheses $\cD^d_{\sigma}$ with $\sigma \in \perm{[d]}{P,\cS_*}$. Recall that for all $\sigma \in \perm{[d]}{P,\cS_*}$, we have $\nu_{\cS_*}^\sigma = \nu_{\cS_*}$. To prove our lower bound, we will lower bound the complexity of learning $\cH_{\mu,\cS_*}^d$ which implies a lower bound on $\cH_{\mu}^d$. Specifically, we will show that with $q/\tau^2 \leq c d^{k_*}$ for some constant $c>0$ that only depends on $P, \mu$ (and the loss $\ell$ for $\DLQ_{\ell}$), statistical query algorithms in $\sA$ cannot distinguish for all $\sigma \in \perm{[d]}{P,\cS_*}$ between $\nu^\sigma=\cD_{\sigma}^d$ and $\nu_{\cS_*}$.

\paragraph*{Proof outline:} For any $\sigma \in \perm{[d]}{P,\cS_*}$ and query $\phi \in \cQ_{\sA} \subseteq L^2(\nu_0)$, define
\begin{equation}
\Delta_\sigma (\phi) := \E_{\nu^\sigma} [ \phi] - \E_{\nu_{\cS_*}} [\phi].
\end{equation}
We will show below that there exists a constant $C>0$ that only depends on $\mu$ such that
\begin{equation}\label{eq:second_moment_Delta_bound}
    \sup_{\phi \in \cQ_\sA} \frac{\E_{\sigma} [\Delta_\sigma (\phi)^2 ]}{\norm{\phi}^2_{\scriptscriptstyle L^2(\nu_0)}} \leq C d^{-k_*},
\end{equation}
where $\E_\sigma$ denote the expectation with respect to $\sigma \sim \Unif (\perm{[d]}{P,\cS_*})$.

Let $\phi_1, \ldots, \phi_q$ be the sequence of adaptive queries with responses $\E_{\nu_{\cS_*}} [ \phi_t ]$. Then, by Markov's inequality,
\begin{equation}\label{eq:bound_proba_queries}
\begin{aligned}
    \PP_{\sigma} \left( \exists t \in [q], \; | \Delta_\sigma (\phi_t)| > \tau \norm{\phi}_{L^2(\nu_0)} \right) \leq&~ \frac{q }{\tau^2} \sup_{\phi \in \cQ_\sA} \frac{\E_{\sigma} [\Delta_\sigma (\phi)^2 ]}{\norm{\phi}^2_{\scriptscriptstyle L^2(\nu_0)}}
    \leq C \frac{q d^{-k_*}}{\tau^2}.
\end{aligned}
\end{equation}
Hence, for $q/\tau^2 < d^{k_*}/C$, there exists $\sigma \in \perm{[d]}{P,\cS_*}$ such that for every $t \in [q]$, we have $|\E_{\nu^\sigma} [ \phi_t] - \E_{\nu_{\cS_*}} [\phi_t] | \leq \tau \norm{\phi_t}_{\scriptscriptstyle L^2(\nu_0)}$. Therefore $\E_{\nu_{\cS_*}} [\phi_t]$ responses are compatible to our queries: $\nu^\sigma$ is indistinguishable from $\nu_{\cS_*}$ and the algorithm fails to learn $\cH^d_{\mu}$.

\paragraph*{Decomposing $\Delta_\sigma (\phi)$:} We rewrite $\Delta_\sigma (\phi)$ in terms of $\E_{\nu_{\cS_*}}$ using the Radon-Nikodym derivative:
\begin{equation}
    \Delta_\sigma (\phi) = \E_{\nu_{\cS_*}} \left[\frac{\de \nu^{\sigma}}{\de \nu_{\cS_*}} \phi \right] - \E_{\nu_{\cS_*}} [\phi] = \E_{ \nu_{\cS_*}} \left[  \left(\frac{\de \nu^{\sigma}}{\de \nu_{\cS_*}}(y_{\cS_*},\bx)-1\right) \phi(y_{\cS_*},\bx)  \right].
\end{equation}
Recall that $(y_{\cS_*},\bx_{\cS_*})$ is independent of $\bx_{\sigma (\cS_*^c)} \sim \mu_{x}^{d-r}$ under $\nu_{\cS_*}$, and that $\de \nu^\sigma / \de \nu_{\cS_*} \in L^2 ( \nu_{\cS_*})$ by Assumption \ref{ass:well-behaved}. Therefore, we have the following orthogonal decomposition in $L^2 (\nu_{\cS_*})$ using the tensor basis \eqref{eq:tensor_basis}:
\begin{equation}\label{eq:ortho_decompo_xi_bi}
    \frac{\de \nu^{\sigma}}{\de \nu_{\cS_*}}(y_{\cS_*},\bx)-1 = \sum_{\bi \in \cI^{d-r} \setminus \{\bzero\}} \xi_{\bi}^\sigma ( y_{\cS_*}, \bx_{\cS_*}) \psi_\bi (\bx_{\cS_*^c}), 
\end{equation}
where
\begin{equation}\label{eq:def_xi_bi}
    \xi_{\bi}^\sigma ( y_{\cS_*}, \bx_{\cS_*}) := \E_{\nu_{\cS_*}}\left[ \frac{\de \nu^{\sigma}}{\de \nu_{\cS_*}}(y_{\cS_*},\bx) \psi_\bi (\bx_{\cS_*^c}) \Big|  y_{\cS_*}, \bx_{\cS_*} \right] = \E_{\nu^\sigma} \left[ \psi_\bi (\bx_{\cS_*^c}) \big|  y_{\cS_*}, \bx_{\cS_*} \right].
\end{equation}
Similarly, we decompose $\phi$ orthogonally in $L^2 (\nu_{\cS_*})$:
\begin{equation}\label{eq:ortho_decompo_alpha_bi}
    \phi(y_{\cS_*},\bx) = \sum_{\bi \in \cI^{d-r}} \alpha_{\bi} ( y_{\cS_*}, \bx_{\cS_*}) \psi_\bi (\bx_{\cS_*^c}),
\end{equation}
where
\begin{equation}\label{eq:def_alpha_bi}
    \alpha_{\bi} ( y_{\cS_*}, \bx_{\cS_*}) := \E_{\nu_{\cS_*}} \left[ \phi(y_{\cS_*} , \bx) \psi_\bi ( \bx_{\cS_*^c}) \big|  y_{\cS_*}, \bx_{\cS_*}\right].
\end{equation}
We deduce that we can decompose $\Delta_\sigma (\phi)$ as
\begin{equation}\label{eq:decompo_delta_mi}
\Delta_\sigma (\phi) = \sum_{\bi \in \cI^{d-r} \setminus \{\bzero\}} \E_{\nu_{\cS_*}} \left[ \xi_{\bi}^\sigma ( y_{\cS_*}, \bx_{\cS_*}) \alpha_{\bi} ( y_{\cS_*}, \bx_{\cS_*})\right].
\end{equation}
For convenience, denote $m_\bi^\sigma$ the summand. From the expressions \eqref{eq:def_xi_bi} and \eqref{eq:def_alpha_bi} and by Fubini's theorem, we can rewrite $m_{\bi}^\sigma$ as
\begin{equation}\label{eq:first_Fubini}
\begin{aligned}
    m_\bi^\sigma =&~ \E_{y_{\cS_*},\bx_{\cS_*}} \left[\E_{\nu^\sigma} \left[ \psi_\bi (\bx_{\cS_*^c}) \big|  y_{\cS_*}, \bx_{\cS_*} \right] \cdot \E_{\tbx_{\cS_*^c} \sim \mu_{x}^{P-r}} \left[ \phi(y_{\cS_*} , \bx_{\cS_*},\tbx_{\cS_*^c}) \psi_\bi ( \tbx_{\cS_*^c})\right] \right] \\
    =&~ \E_{(y,\bx,\tbx_{\cS_*^c}) \sim \nu^\sigma \otimes \mu_{x}^{P-r}} \left[ \psi_{\bi} (\bx_{\cS_*^c}) \phi(y , \bx_{\cS_*},\tbx_{\cS_*^c}) \psi_\bi ( \tbx_{\cS_*^c}) \right]\\
    =&~ \E_{\tbx_{\cS_*^c} \sim \mu_{x}^{P-r}} \left[ \psi_\bi ( \tbx_{\cS_*^c}) \E_{(y,\bx) \sim \nu^\sigma} \left[ \phi(y , \bx_{\cS_*},\tbx_{\cS_*^c})  \psi_{\bi} (\bx_{\cS_*^c}) \right] \right].
\end{aligned}
\end{equation}
Let us decouple $y$ from $\bx_{\cS_*}$ inside the query $\phi$ using the Radon-Nikodym derivative: conditional on $\tbx_{\cS_*^c}$, we have
\begin{equation}\label{eq:partial_exp}
\begin{aligned}
    &~\E_{(y,\bx) \sim \nu^\sigma} \left[ \phi(y , \bx_{\cS_*},\bv)  \psi_{\bi} (\bx_{\cS_*^c}) \right] \\
    &~ \qquad \qquad= \E_{(y_{\cS_*^c},\bx)\sim \nu_{\cS_*^c}^\sigma} \left[ \frac{\de \nu^\sigma}{\de \nu_{\cS_*^c}^\sigma} (y_{\cS_*^c},\bx) \cdot  \phi(y_{\cS_*^c} , \bx_{\cS_*},\tbx_{\cS_*^c})  \psi_{\bi} (\bx_{\cS_*^c}) \right].
\end{aligned}
\end{equation}
Similarly to Eq.~\eqref{eq:ortho_decompo_xi_bi}, we have the following orthogonal decomposition in $L^2 (\nu_{\cS_*^c}^\sigma)$:
\begin{equation}\label{eq:second_ortho_xi_j}
    \frac{\de \nu^\sigma}{\de \nu_{\cS_*^c}^\sigma} (y_{\cS_*^c},\bx) = \sum_{\bj \in \cI^r} \xi_\bj^\sigma (y_{\cS_*^c},\bx_{\cS_*^c}) \psi_{\bj} (\bx_{\cS_*}),
\end{equation}
where
\begin{equation}
    \xi_\bj^\sigma (y_{\cS_*^c},\bx_{\cS_*^c}) := \E_{\nu^\sigma} \left[ \psi_{\bj} (\bx_{\cS_*}) \big| y_{\cS_*^c},\bx_{\cS_*^c} \right].
\end{equation}
By Fubini's theorem,
\begin{equation}\label{eq:second_Fubini}
    \begin{aligned}
        &~\E_{(y_{\cS_*^c},\tbx_{\cS_*},\bx_{\cS_*^c})\sim \nu_{\cS_*^c}^\sigma} \left[ \xi_\bj^\sigma (y_{\cS_*^c},\bx_{\cS_*^c}) \psi_{\bj} (\tbx_{\cS_*}) \cdot  \phi(y_{\cS_*^c} , \tbx_{\cS_*},\tbx_{\cS_*^c})  \psi_{\bi} (\bx_{\cS_*^c}) \right]\\
        =&~ \E_{\tbx_{\cS_*}\sim \mu_{x}^r} \left[\psi_{\bj} (\tbx_{\cS_*}) \E_{(y,\bx) \sim \nu^\sigma} \left[  \phi(y,\tbx_{\cS_*},\tbx_{\cS_*^c}) \psi_{\bj} (\bx_{\cS_*})  \psi_{\bi} (\bx_{\cS_*^c}) \right] \right].
    \end{aligned}
\end{equation}
Combining Eqs.~\eqref{eq:partial_exp}, \eqref{eq:second_ortho_xi_j}, and \eqref{eq:second_Fubini} into Eq.~\eqref{eq:first_Fubini} yields the expression
\begin{equation}
    m_\bi^\sigma = \sum_{\bj \in \cI^r} \E_{\tbx \sim \mu_{x}^d}\left[ \psi_{\bj} (\tbx_{\cS_*})  \psi_{\bi} (\tbx_{\cS_*^c}) \E_{(y,\bx) \sim \nu^\sigma} \left[ \phi (y,\tbx) \psi_{\bj} (\bx_{\cS_*})  \psi_{\bi} (\bx_{\cS_*^c}) \right]\right].
\end{equation}

From Eq.~\eqref{eq:decompo_delta_mi}, we deduce the following simple decomposition
\begin{equation}\label{eq:decompo_delta_final}
    \Delta_\sigma (\phi)  = \sum_{ \bi \in \cI^{d}, \;  \supp(\bi) \not\subseteq \cS_* } \E_{\tbx \sim \mu_{x}^d}\left[ \psi_{\bi} (\tbx) \E_{(y,\bx) \sim \nu^\sigma} \left[ \phi (y,\tbx) \psi_{\bi} (\bx) \right]\right].
\end{equation}

\paragraph*{Bounding $\E_{\sigma} [\Delta_\sigma (\phi)^2 ]$:} If $\supp (\bi) \not\in \sigma( \cC_\sA \setminus \cC_*)$, we have by Eq.~\eqref{eq:partial_exp_zero} that 
\begin{equation}
    \E_{\tbx \sim \mu_{x}^d}\left[ \psi_{\bi} (\tbx) \E_{(y,\bx) \sim \nu^\sigma} \left[ \phi (y,\tbx) \psi_{\bi} (\bx) \right]\right] = 0.
\end{equation}
Therefore we can rewrite the sum in Eq.~\eqref{eq:decompo_delta_final} with
\begin{equation}\label{eq:Delta_pi_ind}
\begin{aligned}
    \Delta_\sigma (\phi) =&~  \sum_{\bi \in \cI^d} \; \sum_{S \in \cC_\sA \setminus \cC_*} \E_{(y,\bx,\tbx) \sim \nu^\sigma \otimes \mu_{x}^d}\left[ \psi_{\bi} (\tbx) \phi (y,\tbx) \psi_{\bi} (\bx) \right] \cdot \ind [ \supp(\bi) = \sigma(S)] \\
    =&~  \sum_{\bi \in \cI^d} \; \sum_{S \in \cC_\sA \setminus \cC_*} \E_{(y,\bx,\tbx) \sim \nu \otimes \mu_{x}^d}\left[ \psi_{\bi} (\tbx) \phi (y,\tbx) \psi_{\bi} (\bx) \right] \cdot \ind [ \sigma(\supp(\bi)) = S].
    \end{aligned}
\end{equation}
Recall that for every $S \in \cC_\sA \setminus \cC_*$ by construction. Hence,
\begin{equation}
    \E_\sigma \left[\sum_{S_1,S_2 \in \cC_\sA \setminus \cC_*} \ind [ \sigma(\supp(\bi)) = S_1, \sigma(\supp(\bj)) = S_2]\right] \leq C d^{-k_*}.
\end{equation}
By squaring Eq.~\eqref{eq:Delta_pi_ind} and the previous display, we deduce that
\begin{equation}
\begin{aligned}
&~\E_{\sigma} [\Delta_\sigma (\phi)^2 ]  \\
\leq&~ C d^{-k_*} \left( \sum_{\bi \in \cI^d} |\E_{(y,\bx,\tbx) \sim \nu \otimes \mu_{x}^d}\left[ \psi_{\bi} (\tbx) \phi (y,\tbx) \psi_{\bi} (\bx) \right]| \right)^2  \\
\leq&~C d^{-k_*} \left( \sum_{\bi \in \cI^d} \left|\E_{y \sim \mu_y }\left[ \E_{\bx \sim \mu_{x}^d} \left[ \frac{\de \nu}{\de \nu_0} (y,\bx) \psi_{\bi} (\bx) \right]  \E_{\tbx \sim \mu_{x}^d} \left[\phi (y,\tbx) \psi_{\bi} (\tbx) \right] \right] \right| \right)^2 \\
\leq&~ Cd^{-k_*} \left( \sum_{\bi \in \cI^d} \E_{\mu_y }\left[ \E_{\mu_{x}^d} \left[ \frac{\de \nu}{\de \nu_0} (y,\bx) \psi_{\bi} (\bx) \right]^2 \right]^{1/2}  \E_{\mu_y }\left[\E_{\tbx \sim \mu_{x}^d} \left[\phi (y,\tbx) \psi_{\bi} (\tbx) \right]^2 \right]^{1/2} \right)^2\\
\leq&~ C d^{-k_*} \left(\sum_{\bi \in \cI^d} \E_{\mu_y }\left[ \E_{\mu_{x}^d} \left[ \frac{\de \nu}{\de \nu_0} (y,\bx) \psi_{\bi} (\bx) \right]^2 \right] \right)  \left( \sum_{\bi \in \cI^d} \E_{\mu_y }\left[\E_{\tbx \sim \mu_{x}^d} \left[\phi (y,\tbx) \psi_{\bi} (\tbx) \right]^2 \right]\right) \\
\leq&~ C d^{-k_*} \E_{y\sim \mu_y} \left[ \left\| \frac{\de \nu}{\de \nu_0} (y,\cdot )  \right\|_{L^2 ( \mu_{x}^d)}^2 \right] \cdot \E_{y\sim \mu_y} \left[ \left\| \phi (y,\cdot )  \right\|_{L^2 ( \mu_{x}^d)}^2 \right] \\
=&~ C d^{-k_*} \left\| \frac{\de \nu}{\de \nu_0} \right\|^2_{L^2 (\nu_0)} \| \phi \|_{L^2(\nu_0)}^2 \leq C' d^{-k_*} \| \phi \|_{L^2(\nu_0)}^2,
\end{aligned}
\end{equation}
where we used in the third and fourth line Cauchy-Schwarz inequality, and in the last line
\begin{equation}
    \frac{\de \nu}{\de \nu_0} = \frac{\de \mu_{y,\bz}}{\de \mu_y \otimes \mu_x^P} \in L^2 (\nu_0),
\end{equation}
by Assumption \ref{ass:well-behaved}. Note that the above bound is uniform over all $\phi \in L^2(\nu_0)$. This concludes the proof.
\begin{remark}
  We remark that indeed these proofs, after alternatively defining $k_*=\relLeap(\cC_{\sA})$ and $\cC_*$ to be the maximal subset of $\cC_{\sA}$ such that $\relLeap(\cC_*)=k_*-1$, shows that if $q/\tau^2 \leq c d^{k_*}$, then mthods in $\sA$ fail to recover coordinates from $\cC_{\sA} \setminus \cC_*$. 
\end{remark}

\subsection{Theorem \ref{thm:meta_complexity_leap}.(b): lower bound for non-adaptive queries}

The proof follows from a similar argument to the adaptive case.

\paragraph*{Proof outline:} Let $k_* = \Cover (\cC_\sA)$ and $l_*$ that maximizes $\min_{S \in \cC_\sA, l \in S} |S|$. By definition of the cover complexity, $|S| \geq k_*$ for all $S \in \cC_\sA$ with $l_* \in S$. Let $\cS_* := [P] \setminus \{l_*\}$. We define in the non-adaptive case for all $\phi \in \cQ_\sA$ and $\sigma \in \perm{[d]}{P}$,
\begin{equation}
    \Delta_\sigma (\phi) = \E_{\nu^\sigma} [\phi] - \E_{\nu_{\cS_*}^\sigma} [\phi],
\end{equation}
meaning that we compare the query expectation between $\nu^\sigma$ and $\nu^\sigma_{\cS_*}$ where we decoupled the label $y$ from coordinate $\sigma(l_*)$. We show again that the bound \eqref{eq:second_moment_Delta_bound} holds on the second moment of $\Delta_\sigma (\phi)/\norm{\phi}_{\scriptscriptstyle L^2(\nu_0)}$.
Therefore, for any set of $q$ queries $\{\phi_t\}_{t \in [q]} \subset \cQ_\sA$, we get a similar bound \eqref{eq:bound_proba_queries} on the probability of all queries being less or equal to $\tau$ in absolute value. We deduce that for $q/\tau^2 \leq c d^{k_*}$, the algorithm will not be able to distinguish for all $\sigma \in \perm{[d]}{P}$ between $\nu^\sigma$ and $\nu^\sigma_{\cS_*}$, and therefore recover the coordinate $\sigma (l_*)$ in the support.
\paragraph*{Bounding $\E_\sigma [ \Delta_\sigma (\phi)^2]$:} Following the same decomposition of $\Delta_\sigma (\phi)$ as above, we first obtain
\begin{equation}
    \Delta_\sigma (\phi) = \sum_{i \in \cI \setminus \{0\}} \E_{\nu_{l_*}^\sigma} \left[ \xi_{i}^\sigma ( y_{l_*}, \bx_{\sigma(l_*)}) \alpha_{i} ( y_{l_*}, \bx_{\sigma(l_*)})\right],
\end{equation}
and then
\begin{equation}
        \Delta_\sigma (\phi)  = \sum_{ \bi \in \cI^{d}, \;  \sigma(l_*) \in \supp(\bi) } \E_{\tbx \sim \mu_{x}^d}\left[ \psi_{\bi} (\tbx) \E_{(y,\bx) \sim \nu^\sigma} \left[ \phi (y,\tbx) \psi_{\bi} (\bx) \right]\right].
\end{equation}
Therefore, by Eq.~\eqref{eq:partial_exp_zero}, the summand is non zero only if $\supp (\bi) \subseteq \sigma (\cC_\sA)$ with $\sigma (l_*) \in \supp (\bi)$. Recall that in this case $| \supp (\bi)| \geq k_*$ by the choice of $l_*$. The rest of the proof follows using the same computation as in the previous section.

\subsection{Upper Bounds for Adaptive and Non-Adaptive Queries}
Define
\begin{equation}
    \beta = \min_{U \in \cC_\sA} \left| \E_{\mu_{y,\bz}}\Big[ T^U (y) \prod_{i \in U} T^U_i(z_i) \Big]  \right|,
\end{equation}
with $T^U \in \Psi_\sA$ and $T^U_i \in L^2_0(\mu_x)$ chosen to make the expectation non-zero, where without loss of generality we normalize them so that $\| T^U\|_{L^2 (\mu_y)}^2 \prod_{i \in S} \|T_i^U \|_{L^2 (\mu_x)} = 1$. We can then emulate the support recovery algorithm described below Definition \ref{def:leap-and-cover} in Section \ref{sec:Leap-and-Cover} using for every $U_* \in \cC_{\sA}$ and $|S| = |U_*|$, $\phi_S (y,\bx) = T^{U_*}(y) \prod_{i \in S} T^{U_*}_i (x_i)$ (note that here $S$ is an ordered subset) and precision $\tau < \beta/2$, so that if the response $|v_t| > \beta/2$, then $S \in \sigma_* (\cC_\sA)$. Note that $\phi_S (y,\bx)$ are indeed in $\cQ_\sA$: for $\SQ,\CSQ$ this is direct, for $\DLQ_\ell$, we can take $f_t (\bx;\omega) = \bu + \omega \cdot \bv \prod_{i \in S} T^{U_*}_i (x_i)$.

If we take $\tau < \beta /2 $, the number of queries only needs to scale as $O( d^{\Cover_\sA})$ or $O( d^{\Leap_\sA})$. We can further trade-off the precision and accuracy and get the result of several of the above queries with one query with small enough $\tau$. For example, consider testing $d$ subset $\{i\}$ to find $s_*(1)$. Then, one can group each of this query in $\Theta(\log (d))$ groups $G_k$ following the binary representation of $i$. For each of these groups, consider the query
\[
\frac{1}{\sqrt{d}} \sum_{s \in G_k} T^{\{1\}} (y) T^{\{1\}}_1 (z_i). 
\]
Note that this is indeed a query in $\Psi_\sA$: it has $L^2(\nu_0)$ norm $O(1)$ (indeed $\E[T^{(i)}(z_i) T^{(j)}(z_j)] = 0$ for any $i\neq j$ with at least one not in the support).
With precision $O(1/\sqrt{d})$, one can test if one of those expectation is non-zero. Repeating for each of the $\Theta(\log (d))$ groups allows to recover the binary representation of $\{s_*(1)\}$. This idea allows to trade-off the query and precision for any $q/\tau^2 \geq C d^{k_*} \log (d)$.

\section{Proofs of Technical Results}\label{app:proof_technical}
\paragraph*{Proposition \ref{prop:SQvsCSQ}:} Denote $\cY = \{a,b\}$. For any $U \in \cC_{\SQ} (\mu)$, by Definition \ref{def:leap_cover_exponent}, there exists $T \in L^2(\mu_y)$ and $\{T_i \in L^2_0(\mu_x)\}_{i\in S}$ such that
\begin{equation}
\begin{aligned}
    0 \neq&~ \E \Big[ T(y) \prod_{i \in U} T_i (z_i)\Big] \\
    =&~ T(a) \mu_{y,\bz} (y = a) \E \Big[ \prod_{i \in U} T_i (z_i)\Big\vert y = a\Big] + T(b) \mu_{y,\bz} (y = b) \E \Big[ \prod_{i \in U} T_i (z_i)\Big\vert y = b\Big] \\
    =&~ (T(b) - T(a)) \mu_{y,\bz} (y = b) \E \Big[ \prod_{i \in U} T_i (z_i)\Big\vert y = b\Big],
\end{aligned}
\end{equation}
where we used that $\E [ T_i (z_i) ] = 0$. Hence, this expectation is non-zero for any mapping $T(b) \neq T(a)$, in particular $T = \Id$. We deduce that $U \in \cC_{\CSQ} ( \mu)$ too.

Consider $\cX = \R$ and $\mu_x = \gamma$ the standard Gaussian measure. We assume that the label $y = z_1 z_2 \cdots z_P$. Then $\cC_{\CSQ}$ only contains the entire support $[P]$ and $\Cover_{\CSQ}(\mu) = \Leap_{\CSQ} (\mu) = P$. On the other hand, consider the mappings $T (y) = \ind [ |y| \geq \tau ]$ for some $\tau >0$, and $T_i (z_i) = z_i^2 - 1$. Then
\begin{equation}\label{eq:sq_gaussian}
\E [ T(y) T(z_i)] = \E_{\bz_{-i}} \Big[ \E_{z_i} \Big[ (z_i^2 - 1) \ind \big[ |z_i| \cdot \prod_{j \neq i} |z_j| \geq \tau \big]\Big] \Big] > 0,
\end{equation}
and $\cC_{\SQ} ( \mu )$ contains all the singletons $\{i\}$ for $i \in [P]$. We deduce $\Cover_{\SQ} (\mu) = \Leap_{\SQ} (\mu) = 1$.

\paragraph*{Proposition \ref{prop:DLQvsSQ_CSQ}.(a):} Note that $\ell'(u,y)=(u-y)$ for the squared loss. For any $U \subseteq [P]$, and any $\{T_i \in L_2^0(\mu_x): i \in [P]\}$, we have
$$ \E_{\mu_{y,\bz}}\left[(u-y)\prod_{i \in U} T_i(z_i)\right]= \E_{\mu_{y,\bz}}\left[(u\prod_{i \in U} T_i(z_i)\right]-\E_{\mu_{y,\bz}}\left[y\prod_{i \in U} T_i(z_i)\right]=-\E_{\mu_{y,\bz}}\left[y\prod_{i \in U} T_i(z_i)\right].$$
Thus,
$$\E_{\mu_{y,\bz}}\left[(u-y)\prod_{i \in U} T_i(z_i)\right] \neq 0 \text{ if and only if } \E_{\mu_{y,\bz}}\left[(y\prod_{i \in U} T_i(z_i)\right].$$
This immediately implies $\cC_{\DLQ_{\ell}}=\cC_{\SQ}$ by Definition \ref{def:leap_cover_exponent}, and as a consequence $\Leap_{\CSQ}=\Leap_{\DLQ_{\ell}}$ and $\Cover_{\CSQ}=\Cover_{\DLQ_{\ell}}$.
\paragraph*{Proposition \ref{prop:DLQvsSQ_CSQ}.(b):} Consider $U \in \cC_{\SQ}$ with
\begin{equation}
     \E_{\mu_{y,\bz}}\Big[ T (y) \prod_{i \in U} T_i(z_i) \Big] \neq 0.
\end{equation}
Note that without loss of generality we can assume $T \in L^2_0 (\mu_y)$.
If ${\rm span} (\Psi_{\DLQ_\ell})$ is dense in $L^2_0 (\mu_y)$, then for every $\eps >0$, there exists $M_\eps \in \naturals$ and $(\bv_j,\bu_j)_{j \in [M_\eps]} \subseteq \cV \times \cF$ such that
\begin{equation}
    \Tilde{T}_{M_\eps}(y) = \sum_{j \in [M_\eps]}  \bv_j^\sT \nabla \ell (\bu_j,y) 
\end{equation}
has $\| T  - \Tilde{T}_{M_\eps}\|_{L^2(\mu_y)} \leq \eps$. In particular,
\begin{equation}
    \begin{aligned}
        \hspace{-8pt}\left| \E_{\mu_{y,\bz}}\Big[ \Tilde{T}_{M_\eps} (y) \prod_{i \in U} T_i(z_i) \Big]  - \E_{\mu_{y,\bz}}\Big[ T (y) \prod_{i \in U} T_i(z_i) \Big] \right| \leq \| T  - \Tilde{T}_{M_\eps}\|_{L^2(\mu_y)} \prod_{i \in U} \| T_i \|_{L^2 (\mu_x)}.
    \end{aligned}
\end{equation}
Therefore taking $\eps$ sufficiently small, $\Tilde{T}_{M_\eps}$ has non zero correlation with $\prod_{i\in U} T_i$, and in particular, one of the $\bv_j^\sT \nabla \ell (\bu_j,y)$ must have non-zero correlation too. We deduce that $\cC_{\SQ} \subseteq \cC_{\DLQ_\ell}$ and we conclude using $\cC_{\DLQ_\ell} \subseteq \cC_{\SQ}$.

Let us assume that there exists nonzero bounded $T \in L^2_0(\mu_y)$, which we take without loss of generality $T(y) \subseteq [-1,1]$, such that $\E[ \bv^\sT \nabla \ell (\bu,y) T(y)] = 0$ for all $\bu \in \cF, \bv \in V$. The goal is to define $\mu_{y,\bz}$ such that $\cC_{\DLQ_\ell} \subsetneq \cC_{\SQ}$. Specifically, we will construct a joint distribution on $(y,z_i)$ with $\{i\} \in \cC_{\SQ}$ but $\{i \} \not\in \cC_{\DLQ_\ell}$. Let $A \subset \cX$ with $\mu_x(A) \in (0,1)$ and consider $y$ that only depends on $z_i$ through $\ind [z_i \in \cA]$. Denote $P(A|y) = \PP(z_i \in A|y)$. For any $T_i \in L^2_0 (\mu_x)$,
\begin{equation}
    \E [ T_i(z_i) | y] = P(A|y) \E[T_i(z_i) | z_i \in A] + P(A^c|y) \E[T_i(z_i) | z_i \not\in A].
\end{equation}
Denote $\kappa(A) = \E[T_i(z_i) | z_i \in A]$. By definition $\E[T_i(z_i)] = 0 = \mu_x (A) \kappa (A) + \mu_x (A^c) \kappa(A^c) $, and therefore $\kappa(A^c) = - \mu_x(A) \kappa (A)/(1-\mu_x(A))$. Hence,
\begin{equation}
    \E [ T_i(z_i) | y] = \frac{\kappa(A)}{1-\mu_x (A)} \left[ P(A|y) - \mu_x (A) \right] .
\end{equation}
Taking $\lambda> (1-\mu_x(A))/\mu_x(A)$, we can set
\begin{equation}
    P(A|y) := \lambda^{-1} (1 - \mu_x (A)) T(y) + \mu_x(A) \in (0,1).
\end{equation}
We then define the joint distribution of $(y,z_i)$ as 
\begin{equation}
    \mu_{y,z_i} := [\mu_x (z_i|z_i \in A) P(A|y) + \mu_x (z_i|z_i \not\in A) P(A^c|y) ] \mu_y (y).
\end{equation}
Note that $\E [ P(A|y)] = \mu_x (A)$ using that $\E[T(y)] =0$, and therefore the marginals of $\mu_{y,z_i}$ are indeed $\mu_y$ and $\mu_x$. 
For any $T_i \in L^2_0 (\mu_x)$ and $(\bu,\bv)\in\cF\times V$, we have
\begin{equation}
    \E_{\mu_{y,z_i}} [\bv^\sT\nabla \ell (\bu,y) T_i(z_i)] = \frac{\kappa(A)}{\lambda} \E_{\mu_y} [ \bv^\sT\nabla \ell (\bu,y) T(y)] = 0,
\end{equation}
and therefore $\{i\} \not\in \cC_{\DLQ_\ell}$. On the other hand $T \in L^2 (\mu_y)$ and therefore $\{i\} \in \cC_{\SQ}$. We conclude that the joint distribution $(y,z_i)$ has $\Leap_{\DLQ_\ell} = \Cover_{\DLQ_\ell} = \infty$ while $\Leap_{\SQ} = \Cover_{\SQ} =1$.

\paragraph*{Proposition \ref{prop:DLQvsSQ_CSQ}.(c):} We construct a simple example with $\cX = \{+1,-1\}$ and $ \mu_x = \Unif(\cX)$, i.e., uniform on the discrete hypercube. Consider $P=2k$ and label
\begin{equation}
    y=z_1 z_2 \ldots z_{2k} \left(\sum_{i \in [2k]} z_i \right).
\end{equation}
The set $\cC_{\CSQ} (\mu)$ contains all subsets $[2k] \setminus \{i\}$ for all $i \in [2k]$, and therefore 
\begin{equation}
\Cover_\CSQ  (\mu) = \Leap_\CSQ (\mu) = 2k-1.
\end{equation}
Consider the loss function $\ell (u,y) = \frac{1}{2}(y - u)^2 + \frac{1}{4} (y-u)^4$. The derivative is given by
\begin{equation}
    \ell' (u,y) = u - y + u^3 - 3u^2y + 3u y^2 - y^3. 
\end{equation}
We have
\begin{equation}
\begin{aligned}
    y^2 =&~ 2k + 2\sum_{i<j} z_i z_j,\qquad y^3 =  z_1 z_2 \ldots z_{2k} \left(\sum_{i \in [2k]} z_i \right)^3.
\end{aligned}
\end{equation}
For $k\geq 3$, $y^3$ only contains monomials over $\geq 2k -3 \geq 3$ coordinates. Therefore $\cC_{\DLQ_\ell} (\mu)$ contains all pairs $\{i,j\}$ but no singleton. Hence,
\begin{equation}
\Cover_{\DLQ_\ell}  (\mu) = \Leap_{\DLQ_\ell} (\mu) = 2.
\end{equation}
Finally, for $\cC_\SQ$, taking $T(y)= y^{2k-1}$, it contains all singleton $\{i\}$ and therefore
\begin{equation}
  \Cover_{\SQ}  (\mu) = \Leap_{\SQ} (\mu) = 1.  
\end{equation}

\subsection{Proof of Theorem \ref{thm:universal_loss}}

Note that $\cC_{\sA}$ does not change if we add or remove constant functions to $\Psi_{\sA}$.

\paragraph*{$\ell_1$-loss:} First consider $\ell (u,y) = |y - u |$. We have $\nabla \ell (u,y) := \ell'(u,y)= \sign(u-y)$, where we can set without loss of generality $\sign (0) = 0$. From  \citet[Theorem 5]{hornik1991approximation} and \citet[Theorem 1]{hornik1991approximation}, we directly conclude that for any probability measure $\mu_y$ on $\R$, we have ${\rm span} (\Psi_\sA)$ dense in $L^2(\mu_y)$ and we conclude using Proposition \ref{prop:DLQvsSQ_CSQ}.(b). 

\paragraph*{Hinge loss:} Consider the Hinge loss $\ell (u,y) = \max (1 - yu,0)$. We get $ \ell'(u,y) = - y \ind_{uy \leq 1}$ where without loss of generality, we set $\ell' (1/y,y) = - y $. For $a> 0$ and $b<0$, we have $\ell'(a,y) - \ell'(b,y) = y \ind [y > a]  - y \ind[y<b]$. Taking $b \to - \infty$, we have $\ell'(a,y) - \ell'(b,y)$ that converges to $y \ind [ y >a]$ in $L^2 (\mu_y)$ (recall that we assume that the second moment of $\mu_y$ is bounded). Similarly, we can construct $y \ind[y >a]$ for $a \leq 0$. Following the proof of \citet[Theorem 1]{hornik1991approximation}, if ${\rm span} (\Psi_\sA)$ is not dense in $L^2 (\mu_y)$, there exists $g \in L^2 (\mu_y)$ such that 
\begin{equation}\label{eq:cond_orth_hinge}
    \int y \ind [ y > u] g (y) \de \mu_y = 0 , \qquad \forall u \in \R.
\end{equation}
Define the signed measure $\sigma (B) = \int_B y g(y) \de \mu_y$. It is finite since $\int |y g(y)|  \de \mu_y \leq \| y \|_{L^2(\mu_y)} \| g \|_{L^2(\mu_y)} < \infty$ by assumption. We deduce that
\begin{equation}
    \int \ind [ ay - u >0] \de \sigma (y) = 0 \qquad \forall (a,u) \in \R^2.
\end{equation}
(The case $a = 0$ is obtained by taking $u \to -\infty$ in Eq.~\eqref{eq:cond_orth_hinge}.)
We then use that $\ind[ay-u>0]$ is discriminatory and therefore no such finite signed measure exists. 

\paragraph*{Exponential loss:} Consider $\cY \subseteq [-M,M]$ and $\ell (u,y) = e^{-uy}$, so that $\ell'(u,y) = - y e^{-yu}$. Any Borel probability measure $\mu_y$ on $[-M,M]$ is regular, and in particular continuous functions and therefore polynomials are dense in $L^2 (\mu_y)$. To show that ${\rm span} \{ y e^{-yu} : u \in \R\}$ is dense in $L^2 (\mu_y)$, it is sufficient to show that we can approximate any monomial $y^k$ in $L^2 (\mu_y)$. This is readily proven recursively, using that
\begin{equation}
    y e^{-yu} = \sum_{l=0}^{k-1} \frac{(-1)^l}{l!} u^l y^{l+1} + O(u^l)\cdot e^M.
\end{equation}
This concludes the proof of this theorem.

\section{SGD, mean-field and limiting dimension-free dynamics}\label{app:DF_PDE}
Recall our setup of a two-layer neural network \eqref{eq:NN_original} with $M$ hidden units
\begin{equation}\label{eq:app-NN_original}
f_{\NN} ( \bx ; \bTheta) = \frac{1}{M} \sum_{j \in [M]} a_j \sigma (\< \bw_j,\bx\> + b_j) + c, \qquad c, \{ b_j\}_{j \in [M]} \in \R , \quad \{ \bw_j \}_{j \in [M]} \in \R^d. \tag{NN1}
\end{equation}
Also, recall the reparameterized form \eqref{eq:NN_param}: for $\bTheta = (\btheta_j)_{j \in [M]}\in \R^{M(d+3)}$ 
\begin{equation}\label{eq:app-NN_param}
f_{\NN} ( \bx ; \bTheta) = \frac{1}{M} \sum_{j \in [M]}  \sigma_* (\bx;\btheta_j), \qquad \sigma_* (\bx;\btheta_j) = a_j \sigma ( \<\bw_j,\bx\> + b_j ) + c_j. \tag{NN2}
\end{equation}
As argued, the two neural networks \eqref{eq:NN_original} and \eqref{eq:NN_param} are equivalent throughout under the initialization $c_1 = \ldots = c_M = c$. Also, recall the definitions of the risk and the excess risk, respectively, for a data distribution $(y,\bx ) \sim \cD$, with respect to a loss function $\ell : \R \times \cY \to \R_{\geq 0}$. 
\begin{equation}\label{eq:app-risk_functional}
\cR (f) = \E_{\cD} \left[  \ell (f(\bx), y)\right], \quad \text{and}\quad \ocR (f) = \cR(f) \,- \hspace{-2mm}\inf_{\scriptscriptstyle \bar{f}:\{\pm 1\}^d \to \R} \cR(\bar{f}).
\end{equation}
\paragraph{Batch Stochastic Gradient Descent.} We train the parameters $\bTheta$ using batch-SGD with loss $\ell$ and the batch size $b$. Even in more generality than stated in \eqref{eq:batch_SGD}, we allow for time-varying step size $( \bfeta_k)_{k\geq 0}$, where $\bfeta_k \in \R^{d+3}$ can be different for different parameters (e.g., different layers), and $\ell_2$-regularization $\blambda \in \R^{d+3}$. We initialize the weights $(\btheta_j)_{j \in [M]} \iid \rho_0$ from \eqref{eq:init-main}, and at each step, given samples $(\{(\bx_{ki},y_{ki}): i \in [b]\})_{k \geq 0}$, the weights are updated using
\begin{equation}\label{eq:app-batch_SGD}
\btheta_j^{k+1} = \btheta_j^k - \frac{1}{b} \sum_{i \in [b]}  \ell' (f_\NN (\bx_{ki};\bTheta^t),y_{ki}) \cdot \bH_k \nabla_{\btheta} \sigma_* ( \bx_{ki};\btheta_j^k ) - \bH_k \bLambda \btheta_j^k, \tag{$\ell$-bSGD-g}
\end{equation}
where we introduced $\bH_k:= \diag(\bfeta_{k})$ and $\bLambda:= \diag ( \blambda)$. 
\paragraph*{Mean-Field Dynamics.} A rich line work \citep{chizat2018global, mei2018mean,rotskoff2018neural,sirignano2020mean} has established a crisp approximation between the dynamics of one-pass-batch-SGD \eqref{eq:app-batch_SGD} and a continuous dynamics in the space of probability distributions in $\R^{d+3}$. To any distribution $\rho \in \cP(\R^{d+3})$, we associate the infinite-width neural network
\begin{equation}
f_{\NN} (\bx;\rho) = \int \sigma_* (\bx;\btheta) \rho (\de \btheta).
\end{equation}
In particular, we recover the finite-width neural network \eqref{eq:NN_param} by taking $\hat \rho^{(M)} := \frac{1}{M} \sum_{j \in [M]} \delta_{\btheta_j}$.

Denote $\hat \rho^{(M)}_k$ the empirical distribution of the weights $\{\btheta^k_j\}_{j \in [M]}$ after $k$-steps of \eqref{eq:app-batch_SGD}. In our \eqref{eq:app-batch_SGD} dynamics, we allow for a generic step-size schedule that is captured by 
\begin{equation}\label{eq:step-size-schedule-discrete}
    \bfeta_k = \eta \bxi (k \eta) \quad \text{ for some function } \quad \bxi : \R \to \R^{d+3}_{\geq 0}\, ,
\end{equation}
and that the data at each step are sampled i.i.d. from $\cD$; the case of $\cD$ empirical distribution on training data corresponds to multi-pass batch-SGD, while $\cD$ population distribution corresponds to online batch-SGD. Under some reasonable assumptions (that we will list below on the activation, step-size schedule $\bxi$, initialization $\rho_0$, and the loss $\ell$) on taking $M$ sufficiently large and $\eta$ sufficiently small, setting $k = t/\eta$, $\hat \rho^{(M)}$ is well approximated by a distribution $\rho_t \in \cP(\R^{d+3})$ that evolves according to the PDE:
\begin{equation}\label{eq:MF_dynamics}
    \begin{aligned}
        \partial_t \rho_t  =&~ \nabla_{\btheta} \cdot ( \rho_t \bH(t) \nabla_\btheta \psi (\btheta ;\rho_t)), \\
        \psi(\btheta;\rho_t) = &~ \E_{\cD} [ \ell'( f_{\NN} (\bx; \rho_t),y)  \sigma_* (\bx;\btheta )] + \frac{1}{2}\btheta^\sT \bLambda \btheta, 
    \end{aligned}\tag{MF-PDE}
\end{equation}
where the initial distribution is $\rho_0$ and we defined $\bH(t) = \diag (\bxi (t))$. We will refer to the distribution dynamics \eqref{eq:MF_dynamics} as the mean-field (MF) dynamics. We now specify the assumptions on hyperparameters under which a non-asymptotic equivalence can be derived.
  \begin{assump}[Activation]\label{assump:lip-activation} Our $\sigma : \R \to \R$ is three times differentiable with $\| \sigma^{(k)} \|_\infty \leq K$, $k = 0, \ldots ,3$.
  \end{assump} 
\begin{assump}[Bounded, Lipschitz step-size schedule]\label{assump:lip-step-size-function}
    $t \mapsto \bxi (t) = (\xi_i (t))_{i \in [d+3]}$ has bounded Lipschitz entries: $\| \xi_i \|_\infty \leq K$ and $ \| \xi_i \|_{\Lip} \leq K$.
\end{assump} 
  \begin{assump}[Initialization]\label{assump:init} The initial distribution $\rho_0 \in \cP (\R^{d+3})$ is of the form:
\[
(a^0,b^0,\sqrt{d} \cdot \bw^0, c^0) \sim \mu_a \otimes \mu_b \otimes \mu_w^{\otimes d} \otimes \delta_{c = \overline{c}},
\]
where $\mu_a,\mu_w, \mu_b$ are independent of $d$. We further assume that $\mu_a$ supported on $|a | \leq K$, $|\overline{c}| \leq K$, and $\mu_w$ is symmetric and $K^2$-sub-Gaussian.
  \end{assump} 
  \begin{assump}[Loss]\label{assump:nice-loss} The loss $\ell : \R \times \cY \to \R_{\geq0}$ is twice-differentiable in its first argument for all $y \in \cY$ and satisfies 
  $$\text{$|\ell'(u,y)| \leq K(1+|u|)$ and $|\ell''(u,y) | \leq K$ for all $y\in \cY$.}$$
  \end{assump}
Assumptions \ref{assump:lip-activation}-\ref{assump:init} are similar to \cite{abbe2022merged}. Informally, Assumption \ref{assump:nice-loss} states that loss has a quadratic upper bound (which is analogues to the smoothness assumption in the convex optimization literature). Later, to show the learnability of leap one junta settings, we will further require $\ell(\cdot,y)$ to be convex and analytic. We note that together these assumptions holds for $(i)$ the standard logistic loss for the classification setting when $\cY \in \{\pm 1\}$, and $(ii)$ for the squared loss and any of its ``analytic perturbations'' for the regression setting when $\cY \subseteq [-B,B]$. In particular, we would like to mention $\ell(u,y)=(u-y) \text{archsinh}(u-y)$, for which we have $\Leap_{\DLQ_{\ell}}=\Leap_{\SQ}$ and these assumptions also hold, and hence our results apply. For classification, by Proposition \ref{prop:SQvsCSQ}, anyway $\Leap{\SQ}=\Leap{\CSQ}=\Leap_{\DLQ_{\ell}}$ for any non-degenerate loss (including the standard logistic loss for which our results hold). 
\paragraph{Limiting Dimension-Free Dynamics.} Consider the isotropic parameters $\bH_k = \diag (\eta^a_k,\eta_k^w \id_d, \eta_k^b, \eta^c_k)$ and $\bLambda = \diag ( \lambda^a,\lambda^w \id_d, \lambda^b, \lambda^c)$. Using technical ideas from \citet[Secion 3]{abbe2022merged}, the \eqref{eq:app-batch_SGD} concentrates on an effective \textit{dimension-free} dynamics in the limit $M,d \to \infty$ and $\eta \to 0$. This limiting dynamics corresponds to the gradient flow on $\cR(f)$ of the following effective \textit{infinite-width} neural network 
\begin{equation}
    f_{\NN} (\bz ; \barrho_t) = \int \osigma_* (\bz;\barbtheta^t) \barrho_t (\de \barbtheta^t), \qquad  \osigma_* (\bz;\barbtheta^t) = c^t + a^t \E_G [\sigma ( \< \bz , \bu^t \> +s^tG + b^t)],
\end{equation}
where $G \sim \normal (0,1)$ and $\barrho_t \in \cP(\R^{P+4})$ is the distribution over $\barbtheta^t = (a^t,b^t,\bu^t,c^t,s^t)$ with $\bu^t \in \R^P$.
\begin{align}\label{eq:app-DF-dynamics}
       \partial_t \barrho_t =&~ \nabla_{\barbtheta} \cdot\left(\barrho_t \obH(t) \cdot \nabla_{\barbtheta} \psi (\barbtheta , \barrho_t) \right), \nonumber\\
    \psi (\barbtheta , \barrho_t) =&~  \E_{\mu_{y,\bz}} \left[ \ell'(f_{\NN}(\bz ; \barrho_t), y) \osigma_* ( \bz ; \barbtheta )\right] + \frac{1}{2} \barbtheta^\sT \obLambda \barbtheta, \tag{DF-PDE}
\end{align}
where $\obH(t)=\diag(\xi^a, \xi^{b},\xi^{w}\bI_{P}, \xi^{c},\xi^{w})(t)$ and $\obLambda=\diag(\lambda^a,\lambda^b, \lambda^w \bI_{P}, \lambda^w)$ from initialization 
\[
\barrho_0 :=  \mu_a \otimes \mu_b \otimes \delta_{\bu^0 = \bzero} \otimes \delta_{c^0 = \overline{c}} \otimes \delta_{s^0 = \sqrt{m_2^w}},
\]
where $m_2^w=\E_{W\sim \mu_{w}}[W^2]^{1/2}$ is the second moment of $\mu_w$. The non-asymptotic equivalence is characterized by the following theorem that can be seen as a generalization of \cite[Theorem 5]{abbe2022merged} to non-squared losses.
\begin{theorem}[Formal statement of Eq.~\eqref{eq:DF-SGD-approx}]\label{thm:SGD_to_DF}
     Fix activation, step-size schedule, initialization, and the loss such that Assumptions \ref{assump:lip-activation} to \ref{assump:nice-loss} hold and consider any $T \geq 1$ independent of $d$. Let $(\barrho_t)_{t \geq 0}$ be the solution of \eqref{eq:app-DF-dynamics}, and $\{\bTheta_k \}_{k \geq 0}$ the trajectory of SGD \eqref{eq:app-batch_SGD} with initialization $\{ \btheta_j^0\}_{j \in [M]} \iid \rho_0$. Then there exist a constant $C_{K,T} >0$ that only depends on $K,T$, such that for any $d,M,\eta>$0 with $M \leq e^d$ and $0<\eta \leq 1/[C_{K,T} (d + \log(M))]$, and
    \begin{equation}
      \hspace{-7pt}  \left| \cR (f_{\NN} (\cdot ; \bTheta^k )) - \cR (f_{\NN} (\cdot ; \barrho_{\eta k} ))\right| \leq C_{K,T} \left\{ \sqrt{\frac{P}{d}} + \sqrt{\frac{\log(M)}{M}} + \sqrt{\eta}\sqrt{\frac{d + \log(M)}{b} \vee 1}\right\}
    \end{equation}
    for all $k \leq T/\eta$, with probability at least $1 - 1/M$.
\end{theorem}
This theorem follows by a straightforward modification of the proof of \cite[Theorem 16, Propoition 15]{abbe2022merged} using propagation of chaos argument. We note that the only property about the loss that is needed is Assumption \ref{assump:nice-loss}. See the intuition in Section \ref{sec:gd-on-NN} that \eqref{eq:app-batch_SGD} dynamics are well approximated by \eqref{eq:app-DF-dynamics} in the linear scaling.
\subsection{DF-PDE alignment with the support.}
In this subsection, we provide a formal statement of Theorem \ref{thm:DF_PDE_Leap_informal} and its proof. The following is an extension of \cite[Theorem 7]{abbe2022merged} from square loss to general loss. 
\begin{theorem}[Formal statement of Theorem \ref{thm:DF_PDE_Leap_informal}]\label{thm:DF_PDE_Leap}
    Assume that $\Leap_{\DLQ_\ell} (\mu)> 1$. Then there exists $U \subset [P]$, $|U|\geq 2$, such that for all $t \geq0$, the DF dynamics solution remains independent of coordinates $(z_i)_{ i \in U}$, i.e.,  $f_{\NN} (\bz;\rho_t) = f_{\NN} ((z_i)_{i \not\in U} ; \rho_t)$, and therefore fails at recovering the support. 
    
    Conversely, if $\Leap_{\DLQ_\ell} (\mu) = 1$ and $\ell$ is analytic with respect to its first argument, then almost surely over $\overline{c}$, there exists $t>0$ such that $f_{\NN} (\bz;\rho_t)$ depends on all $[P]$.
\end{theorem}




\begin{proof}[Proof of Theorem \ref{thm:DF_PDE_Leap}]
The proof follows similarly to the proof of \cite[Theorem 7]{abbe2022merged}. Assume $\Leap_{\DLQ_\ell} (\mu) >1$. Consider $\cC_* \subset \cC_{\DLQ_\ell}$ the maximal subset  such that $\Leap ( \cC_*) = 1$ and denote $U_* = \bigcup_{U\in \cC_*} U$. In words, $U_*$ contains the subset of coordinates that are reachable by doing leaps of at most 1. By the assumption that $\Leap_{\DLQ_\ell} (\mu) >1$, we must have $|[P] \setminus U_*| \geq 2$. Let us show that the weights $u_i$ for $i \in [P] \setminus U_*$ remain equal to $0$ throughout the dynamics. 

For simplicity, we forget about the other parameters which we set to $b=c=s =0$. For convenience, denote $\Omega = [P] \setminus U_*$. First, we can bound by Gronwall's inequality, for $t \leq C$, $|a^t| \leq KC'$ for all $a^t$ on the support of $\rho_t$. From the proof of Theorem \ref{thm:meta_complexity_leap}, conditioning on already explored coordinates does not change the leap-complexity. In particular,
\[
\E [ \ell' (g(\bz_{U_*}),y) z_i | \bz_{U_*} ] = 0, \qquad \forall i \in \Omega.
\]
Consider the time derivative of $u_i^t$ for $i \in \Omega$. We have
\[
\begin{aligned}
   &~|\ell' (f_\NN (\bz_{U_*},\bz_{\Omega};\rho_t),y) \sigma ' (\<\bu_{U_*}^t,\bz_{U_*}\> + \<\bu_{\Omega}^t,\bz_{\Omega}\>) -  \ell'(f_\NN (\bz_{U_*},\bzero;\rho_t),y) \sigma ' (\<\bu_{U_*}^t,\bz_{U_*}\> ) |\\
   \leq&~K \| \sigma \|_\infty  \int |\tilde a^t| | \sigma (\<\tilde\bu_{U_*}^t,\bz_{U_*}\> + \<\tilde\bu_{\Omega}^t,\bz_{\Omega}\>) - \sigma (\<\tilde\bu_{U_*}^t,\bz_{U_*}\> ) | \rho_t (\de \Tilde \btheta) \\
   &~ + K \| f_\NN (\cdot;\rho_t) \|_\infty  | \sigma (\<\bu_{U_*}^t,\bz_{U_*}\> + \<\bu_{\Omega}^t,\bz_{\Omega}\>) - \sigma (\<\bu_{U_*}^t,\bz_{U_*}\> ) | \\
   \leq&~ K \sup_{j \in \Omega, u_j^t \in \supp (\rho_t)} |u_j^t|,
\end{aligned}
\]
where we used Assumptions \ref{assump:lip-activation} and \ref{assump:nice-loss}. Hence,
\[
\begin{aligned}
    \left| \frac{\de }{\de t} u_i^t \right| =&~ \left| a^t \E [ \ell'( f_\NN (\bz, \rho_t),y) \sigma' ( \< \bu^t, \bz\>) z_i]\right| \\
    \leq &~ \left| a^t \E [\ell' (f_\NN (\bz_{U_*},\bzero;\rho_t),y) \sigma ' (\<\bu_{U_*}^t,\bz_{U_*}\> ) z_i]\right| + K \sup_{j \in \Omega, u_j^t \in \supp (\rho_t)} |u_j^t|
\end{aligned}
\]
The first expectation is equal to $0$. Note that $m_\Omega^t : = \sup_{j \in \Omega, u_j^t \in \supp (\rho_t)} |u_j^t|$ has $m_\Omega^0 = 0$ and therefore $m_\Omega^t = 0$ for all $t \geq 0$.
\end{proof}

\newpage


\section{Learning Leap One Juntas with SGD: Formalizing Theorem \ref{thm:learning_leap_1_informal}}
In this section, we provide our precise layer-wise training algorithm and the proof that it succeeds in learning $\Leap_{\DLQ_{\ell}}=1$ junta problems. This section follows the similar training procedure from \cite{abbe2022merged} for the layer-wise training but adapted to training with general loss function $\ell: \R \times \cY \rightarrow \R_{\geq 0}$ to show Theorem \ref{thm:learning_leap_1_informal}. The key difference in the training algorithm from \cite{abbe2022merged} is the use of coordinate-wise perturbed step-sizes for the first layer weights to break the symmetry between the coordinates. This allows us to learn even some degenerate cases having coordinate symmetries with leap one, in contrast to \cite{abbe2022merged}. 
\paragraph{Discrete Dimension-free Dynamics.} We first consider the dimension-free dynamics similar to \eqref{eq:DF-dynamics} but for discrete step-size regime (see \cite[Appendix C]{abbe2022merged}). We initialize to 
$$(a^0,b^0,\bu^0,c^0,s^0) \sim \barrho_0 \in \cP(\R^{P+4}),$$ and $(\barrho_k)_{k\in \naturals}$ are induced distribution on $(a^0,b^0,\bu^0,c^0,s^0)$ recursively defined as
\begin{align}\label{eq:discrerte-DF-dynamics}
   a^{k+1}&=a^k-\eta_k^a \left( \E_{\bz,G} [ \ell'(\fNN(\bz;\barrho_k),y) \sigma(\<\bu^k,\bz\>+s^k G+b^k)]+\lambda^a a^k \right) \nonumber\\
    \bu^{k+1}&=\bu^k-\bfeta_k^{\bu} \circ \left( \E_{\bz,G} [ \ell'(\fNN(\bz;\barrho_k),y) a^k\sigma'(\<\bu^k,\bz\>+s^k G+b^k) \bz]+\lambda^w \bu^k \right) \nonumber\\
    s^{k+1}&=s^k-\eta^w_k\left( \E_{\bz,G} [ \ell'(\fNN(\bz;\barrho_k),y) a^k\sigma'(\<\bu^k,\bz\>+s^k G+b^k) G ]+\lambda^w s^k \right) \tag{d-DF-PDE}\\
    b^{k+1}&=b^k-\eta^b_k\left( \E_{\bz,G} [ \ell'(\fNN(\bz;\barrho_k),y) a^k\sigma'(\<\bu^k,\bz\>+s^k G+b^k)]+\lambda^b b^k \right) \nonumber\\
    c^{k+1}&=c^k-\eta^c_k\left( \E_{\bz,G} [ \ell'(\fNN(\bz;\barrho_k),y)]+\lambda^c c^k \right). \nonumber
\end{align}
\subsection{Training Algorithm}
Choose a loss $\ell: \R \times \cY \to 
\R_{\geq 0}$ and the activation $\sigma(x)=(1+x)^{L}$ for $L \geq 2^{8P}$. We train for $\Theta(1)$ total steps in two phases and the batch size of $\Theta(d)$ with the following choice of hyperparameters.
\begin{itemize}
    \item \textbf{No regularization:} set the regularization parameters $\lambda^w=\lambda^{a}=\lambda^{b}=\lambda^{c}=0$.
    \item \textbf{Initialization:} Initialize the first layer weights and biases to deterministically 0. 
    $$(b^0_j,\sqrt{d} \bw_j^0) \iid \mu_b \otimes \mu_{w}^{\otimes d} \equiv \delta_{b=0} \otimes \delta_{w=0}^{\otimes d}.$$
    The second layer weights are sampled uniformly from $[-1,1]$, e.g. $a_j \iid \mu_a \equiv \Unif([-1,1]).$ Finally, choose $c^0 \sim \delta_{c=\bar{c}}$ for the given global bias choice $\bar{c} \in \R$. 
    \item For the dimension-free dynamics, this corresponds to taking 
    $$(a^0, b^0,\bu^0, c^0,s^0) \sim \bar{\rho}^0 \equiv \Unif([-1,1]) \otimes \delta_{b=0} \otimes \delta_{w=0}^{\otimes P} \otimes \delta_{c=\bar{c}} \otimes \delta_{s=0}.$$  
   \item We deploy a two-phase training procedure. Given parameters $\eta$ and $\bkappa \in [1/2,3/2]^{d}$:
   \begin{enumerate}
       \item \textbf{Phase 1:} For all $k \in \{0,\dots,k_1-1\}$, set $\eta_{k}^{a}=\eta_k^b=\eta^c_k=0$ and $\bfeta_k^{\bw} \in \R^d$ such that $ \eta_{k}^{w_i} =\eta \kappa_{i}$ for all $i \in [d]$. For the dimension-free dynamics, train the first layer weights $\bu^k$ for $k_1$ steps, while keeping other parameters fixed, i.e $a^k=a^0, b^k=0, c^k=\bar{c}$.
        \item \textbf{Phase 2:} Set $\eta_k^{a}=\eta$ and $\bfeta_{k}^{\bw}=\bzero, \eta^b_k=\eta^c_k=0$ for $k \in \{k_1, \dots,k_1+k_2-1\}$. In words, train the second layer weights $a^k$ for $k_2$ steps, while keeping the first layer weights fixed at $\bu^k=\bu^{k_1}$.
   \end{enumerate}       
\end{itemize}
We will take $\eta>0$ to be a small constant to be specified layer and $ \kappa_{i}\in \Unif([1/2,3/2])$ be the random perturbation. We also let $b=\Omega(d)$, hiding constants in $\eta,\varepsilon,\P,K,\mu$. For the first phase, it suffices to train for $k_1=P$ time steps. For the second phase, we train for $k_2=k_2(\eta,\varepsilon,\P,K,\mu)=\Theta(1)$ time steps to be specified later as we analyze. Note that only unspecified hyperparameter is the global bias initialization $\bar{c}$ and the results holds almost surely over this choice.

\begin{theorem}[Formal statement of Theorem \ref{thm:learning_leap_1_informal}]\label{thm:leap-one-ub}
Assume $\ell$ is analytic, convex and that Assumption \ref{assump:nice-loss} holds. Then for any $\Leap_{\DLQ_{\ell}}(\mu)=1$ setting, for any $\cD_{\mu,s_*}^d \in \cH^d_{\mu}$, almost surely over the initialization $\Bar{c} \in \R$, and $\bkappa \in [1/2,3/2]^d$, the following holds. For any $\eps>0$, with 
$$b \geq \Omega(d), M=\Omega(1), \text{ and } k_2=\Omega(1),$$ 
using $\Omega(\cdot)$ to hide constants in $\eps, K,P,\mu, \eta,\bkappa$ and $\bar{c}$, the above specified layer-wise training dynamics reaches $\ocR  (f_{\NN} (\cdot;\bTheta^{k_1+k_2}))\leq \eps$ with probability at least $9/10$.
\end{theorem}
\subsection{Proof of Theorem \ref{thm:leap-one-ub}}
The proof of the theorem proceeds by first showing a non-asymptotic approximation guarantee between \eqref{eq:discrerte-DF-dynamics} and \eqref{eq:app-batch_SGD}. To this end, we first start by noting a generalization of \cite[Theorem 23]{abbe2022merged} to any loss under satisfying Assumption \ref{assump:nice-loss}.
\begin{proposition}\label{prop:discrete-DF-SGD}
    Assume that Assumption \ref{assump:nice-loss} on the loss holds. Then for any $0<\eta \leq K$ and for any $\Upsilon \in \naturals$, there exists a constant $C(K,\Upsilon)$ such that for \eqref{eq:app-batch_SGD} and \eqref{eq:discrerte-DF-dynamics} of the specified layer-wise training dynamics, with probability $1-1/M$, we have
    $$\inf_{k=0,\dots,\Upsilon} \left|\cR(\fNN(\cdot;\bTheta^k)) - \cR(\fNN(\cdot;\barrho_k))\right| \leq C(K,\Upsilon) \left( \sqrt{\frac{P+\log d}{d}}+\sqrt{\frac{\log M}{M}} + \sqrt{\frac{d+\log M}{b}} \right)$$
\end{proposition}
The proof of the proposition follows the similar arguments like propagation of chaos used in Theorem \ref{thm:SGD_to_DF}. A direct corollary of this proposition is that it suffices to focus only on the \eqref{eq:discrerte-DF-dynamics} to show Theorem \ref{thm:leap-one-ub}.  
\begin{corollary}\label{cor:last-step}
    Assume that Assumption \ref{assump:nice-loss} holds. Then for total $\Upsilon$ number of \eqref{eq:app-batch_SGD} iterations on a neural network with $M \geq C_0(K,\Upsilon, \varepsilon) $ and the batch-size $b \geq C_1(K,\Upsilon,\eps) \cdot d$, the aforementioned layer-wise training dynamics, with probability at least $9/10$
    $$ \inf_{k=0,\dots,\Upsilon} \left|\cR(\fNN(\cdot;\bTheta^k)) - \cR(\fNN(\cdot;\barrho_k))\right| \leq \frac{\eps}{2}.$$
\end{corollary}
\begin{proof}
    It is easy to see that it is possible to choose $M \geq C_0(K,\Upsilon,\eps)$ and $b \geq C_1(K,\Upsilon,\varepsilon)d$ for sufficiently large constants such that the right hand side of equality in Proposition \ref{prop:discrete-DF-SGD} is bounded by $\eps/2$ and $1/M \leq 1/10$. The corollary then follows from Proposition \ref{prop:discrete-DF-SGD}. 
\end{proof}
In light of Corollary \ref{cor:last-step}, it suffices to simply focus on \eqref{eq:discrerte-DF-dynamics} and show that it achieves the excess error of at most $\eps/2$ within total steps $\Upsilon=O(1)$ hiding constants in $K,\mu,P,\eps$. This will be done by analyzing each phase separately---$(i)$ showing that the training of the top layer weights for steps $k\in \{k_1, \dots,k_1+k_2-1\}$ in Phase 2 is a linear model trained with a convex loss so it succeeds in reaching a vanishing risk as long as the corresponding kernel matrix of the feature map is non-degenerate; $(ii)$ indeed showing that the corresponding kernel matrix of the feature map after $k_1=P$ steps of Phase 1 training on the first layer weights is non-degenerate.

\paragraph{Simplified dynamics.} We first make a simple observation that $b^k=0, s^k=0$ and $c^k=\bar{c}$ throughout; this allows us to analyze a simpler dynamics. To see this, for $(b^k,c^k)$, we use $\eta^c=\eta^b=0$ throughout, and thus, they remain the same from the initialization. We start with $s^0=0$ and using \eqref{eq:discrerte-DF-dynamics} (assuming $s^k=0)$, gives us
$$s^{k+1}=0-\eta^w_k\left( \E_{\bz,G} [ \ell'(\fNN(\bz;\barrho_k),y) a^k\sigma'(\<\bu^k,\bz\>+b^k) G ]+\lambda^w 0 \right)=0.$$

\subsection{Phase 2 (linear training)}
We first analyze a slightly simpler Phase 2 of training. For a convex loss $\ell$, the second layer training just corresponds to linear dynamics with kernel $\bK^{k_1} : \{+1, -1\}^P \times \{+1, -1\}^P \to \R$ given by
\begin{equation}\label{eq:kernel}
    \bK^{k_1}(\bz,\bz')=\E_{a\sim \mu_a}[ \sigma(\<\bu^{k_1}(a),\bz\>)\sigma(\<\bu^{k_1}(a),\bz'\>) ].\tag{kernel}
\end{equation}

We first show some helper lemmas, and using them, show the main lemma. We will then show the proof of the helper lemmas. We start by proving the existence of a sparse function with a small excess error.
\begin{lemma}\label{lem:f*-eps-close}
   For any $\eps>0$, there exists a function $f_*:\{\pm 1\}^d \to \R$ such that
   \begin{equation}\label{eq:f*-eps-close}
       \ocR(f_*)=\left[\cR(f_*)-\inf_{\scriptscriptstyle \bar{f}:\{\pm 1\}^d \to \R} \cR(\bar{f}) \right]\leq \frac{\varepsilon}{4}.
   \end{equation}
   Moreover, $f_*(\bx)=h_*(\bz)$ for some $h_*:\{\pm 1\}^P \to \R$ indpendent of $d$, where $\bz$ is the support of $\cD_{s_*}^d$.
\end{lemma} 
\paragraph{Approximation.} The neural network output after training $\bu$ for $k_1$ steps is given by:
$$\fNN(\bz; \barrho_{k_1})=\bar{c}+\int a \sigma(\<\bu^{k_1}(a),\bz\>) \de \mu_a=\bar{c}+\E_{a\sim \mu_a}\left[a \sigma(\<\bu^{k_1}(a),\bz\>)\right].$$
Observing that for the next $k_2$ steps, only $a$ is trained while keeping $\bu^{k_1}$ fixed, the neural network output is of the following form
\begin{equation}\label{eq:resultant-nn}
    \fNN(\bz; \barrho)=\bar{c}+\int a \sigma(\<\bu,\bz\>) \de \barrho(a,\bu)=\bar{c}+\int a \sigma(\<\bu,\bz\>) \de \barrho(a(\bu)) \de \barrho_{k_1}(\bu) .
\end{equation}
now start by showing that our neural network, there exists $\barrho_*$ such that we can exactly represent $h_*(\bz)$ as long as the resultant \eqref{eq:kernel} after $1^{\mathrm{st}}$ layer training is non-degenerate.
\begin{lemma} \label{lem:approximation}
If $\lambda_{\min}(\bK^{k_1}) \geq c$ for some constant $c$, then there exists $\barrho_* \in \cP(\R^{P+1})$ such that $\de \barrho_*(a,\bu)=\de \barrho(a_*(\bu)) \de \barrho_{k_1}(\bu)$ and for all $\bz \in \{\pm 1\}^P$
$$ h_*(\bz)=\fNN(\bz; \barrho_{*})=\bar{c}+\int a_*\sigma(\<\bu,\bz\>) \de \barrho_{*}(a_*(\bu)) \de \barrho_{k_1}(\bu).$$
Moreover, $\norm{a^*}_{\barrho_*}^2=\int a_*^2 (\bu) \de \barrho_*(a_*(\bu)) \leq c_1$ for some constant $c_1:=c_1(\lambda_{\min}(\bK^{k_1}), h_*, \bar{c}).$
\end{lemma}
\paragraph{Convexity and Smoothness.} Finally, we will show that the risk objective in terms of $\barrho$ (of the form \eqref{eq:resultant-nn}) is convex and smooth in terms of $\barrho$. The convexity follows from the convexity of $\ell(\cdot,y)$ and observing that the neural network output in \eqref{eq:resultant-nn} is linear in $\barrho(a(\bu))$. The smoothness follows from Assumption \ref{assump:nice-loss}.

\begin{lemma}\label{lem:convex-smoothe-properties}Consider $\ell: \R \times \cY \to \R_{\geq 0}$ such that $\ell(\cdot, y)$ is convex for every $y \in \cY$ and that Assumption \ref{assump:nice-loss} holds. Then
 $$\cR(\barrho)=\E[\ell(\fNN(\bz;\barrho),y)], \quad\text{where } \de \barrho(a,\bu)= \de \barrho(a(\bu)) \de \barrho_{k_1}(\bu),$$
 is convex and $H$-smooth with $H=K^3$.
\end{lemma}
While we denote $\cR(\barrho)$ as an objective in terms of the joint measure $\barrho(a,\bu)$, it is actually an objective in $\barrho(a(\bu)))$ since it is of the form where $\de(a,\bu)=\de \barrho(a(\bu)) \de \barrho_{k_1}(\bu)$. Also, the convexity and smoothness is in terms of the argument $\barrho(a(\bu))$.
Having established the convexity and smoothness of the $2^\mathrm{nd}$ layer weights training, we will directly apply the convergence rate guarantees for convex and smooth objectives.
\begin{lemma}\label{lem:convex-smooth-convergence}
    Consider any convex, differentiable objective $f(x)$ that is $H$-smooth. Then for any step-size $\eta \leq \frac{1}{H}$, the gradient descent from initialization $x^0$ reaches an iterate $x^k$ after $k$ steps such that for any $\Tilde{x}$ (not necessarily a minimizer), we have
    $$f(x^T)-f(\tilde{x}) \leq \frac{\norm{x^0-\tilde{x}}^2}{2\eta k}.$$
\end{lemma}
This is a classical convergence rate guarantee for gradient descent on convex smooth objective \citep[Theorem 3.3]{bubeck2015convex}, but noting that a proof generalizes even when $\Tilde{x}$ is not necessarily a minimizer. We will apply this guarantees with $\Tilde{x}$ as a near minimizer of $f$, especially when the actual minimizer may not exist (for example with the logistic loss, the infimum is never achieved). We provide the proof with other lemmas for completeness.
\begin{lemma}\label{eq:dDF-close-for-linear-training}
    Assume that $\lambda_{\min}(\bK^{k_1}) >c>0$ for some constant $c$, where $\bK^{k_1}$ is the \eqref{eq:kernel} matrix. Then discrete dimension-free dynamics \eqref{eq:discrerte-DF-dynamics} in Phase 2 of layer-wise training with step-size $\eta \leq \frac{1}{K^3}$ as specified before after $k_2:=k_2(\eta, \eps,P,K,\mu,\bar{c})$ steps achieves 
$$\ocR(\fNN(\bz;\barrho^{k_1+k_2}))= \left[\cR(\fNN(\bz;\barrho^{k_1+k_2}))-\inf_{\bar{f}:\{\pm 1\}^d \to \R} \cR(\bar{f}) \right] \leq \frac{\eps}{2}.$$
\end{lemma}
\begin{proof}
    First of all, using Lemma \ref{lem:f*-eps-close}, there exists $\barrho_*$ such that
    \begin{align}\label{eq:best-close-to-inf}
    \ocR(\fNN(\bz;\barrho_*))=\cR(\fNN(\bz;\barrho_*))-\inf_{\scriptscriptstyle \bar{f}:\{\pm 1\}^d \to \R}\cR(\bar{f}) \leq \frac{\eps}{4}.
    \end{align}
Moreover, by Lemma \ref{lem:convex-smoothe-properties}, the objective $\cR(\barrho)$ is convex and $K^3$-smooth, and observe that \eqref{eq:discrerte-DF-dynamics} are performing gradient descent with step-size $\eta_k^a=\eta \leq \frac{1}{K^3}$ on this objective. Thus, using Lemma \ref{lem:convex-smooth-convergence}
\begin{equation}
  \left[\cR(\fNN(\bz;\barrho^{k_1+k}))-\cR(\fNN(\bz;\barrho_*)) \right]:= \left[\cR(\barrho^{k_1+k})-\cR(\barrho_*) \right] \leq \frac{\lVert a^{k_1}-a^* \rVert_{\barrho_{k_1},\barrho_*}^2}{2 \eta k}.
\end{equation}
Recall that $a^{k_1}\sim \mu_a \equiv \Unif([-1,1])$.
\begin{align*}
  \lVert a^{k_1}-a^* \rVert_{\barrho_{k_1},\barrho_*}^2 =\int (a-a^*(\bu))^2 \de \mu_a \de \barrho_*(a^*(\bu)) \de \barrho_{k_1}(\bu) \leq 1+2\norm{a^*}_{\barrho_*}+ \norm{a^*}^2_{\barrho_*} \leq c_2,
\end{align*}
for some constant $c_2:=c_2(\lambda_{\min}(\bK^{k_1}) ,P,\mu, \bar{c})$. Therefore, choosing $k_2 \geq \frac{2c_2}{\eta \eps}$ 
\begin{equation}\label{eq:dynamics-close-to-best}
  \left[\cR(\fNN(\bz;\barrho^{k_1+k_2}))-\cR(\fNN(\bz;\barrho_*)) \right] \leq \frac{\eps}{4}.
\end{equation}
Combining \eqref{eq:dynamics-close-to-best} and \eqref{eq:best-close-to-inf}
\begin{align*}
  \ocR(\fNN(\bz;\barrho^{k_1+k_2})) =\left[\cR(\fNN(\bz;\barrho^{k_1+k_2}))- \cR(\fNN(\bz;\barrho_*)) \right]+ \left[\cR(\fNN(\bz;\barrho_*))- \inf_{\bar{f}:\{\pm 1\}^d \to \R} \cR(\bar{f}) \right]  \leq \frac{\eps}{2}.
\end{align*}
\end{proof}
Thus, if we can show that $\lambda_{\min}(\bK^{k_1}) \geq c>0$ for some constant $c:=c(\eta, P,K,\mu)$ for a sufficiently small constant $\eta >0$ that only depends on $K,P,\mu$, then it immediately implies the desired result by Corollary \ref{cor:last-step} with $\Upsilon=k_1+k_2$. The goal of Appendix \ref{app:leap_one_ub} is to show that this holds after $k_1=P$ steps of the first layer weight training. We now return to the deferred proofs of helper lemmas.
\subsubsection{Proof of helper lemmas}
\begin{proof}[Proof of Lemma \ref{lem:f*-eps-close}] Note that Eq.~\eqref{eq:f*-eps-close} simply follows from the definition of infimum and the excess risk functional $\ocR(\cdot)$. To show that $f_*(\bx)=h_*(\bz)$, we make an observation that for any function $f:\{\pm 1\}^d \to \R$, one can define $h_*:\{\pm 1\}^P \to \R$ that only depend on the support and achieve risk no worse than $f_*$. Define $h_*(\bz)=\E_{\bz^c} [f_*(\bx)]$, where $\bz^c=\bx\setminus \bz$ (the coordinates outside the support). In other words, $h_*$ is the boolean function after ignoring the monominals that do not depend on $\bz$. Then
\begin{align*}
    \cR(f_*)&=\E_{(y,\bx)\sim \cD_{s_*}^d}\left[\ell(f_*(\bx),y)\right]= \E_{\mu_{y,\bz}}[\E_{\mu_{\bz^c \mid y,\bz}}[\ell(f_*(\bx),y)]] \\
    &\geq \E_{\mu_{y,\bz}}[\ell(\E_{\mu_{\bz^c \mid y,\bz}}[f_*(\bx)],y)]=\E[\ell(h_*(\bz),y)]=\cR(h),
\end{align*}
where the only inequality follows from Jensen's inequality and convexity of $\ell(\cdot,y)$.
\end{proof}

\begin{proof}[Proof of Lemma \ref{lem:approximation}] Define $g_*(\bz)=h_*(\bz)-\bar{c}.$ Our goal is to show the existence of $\barrho_*$ such that
$$g_*(\bz)=\E_{(a_*,\bu) \sim \barrho_*}[a_* \sigma(\<\bu,\bz\>)].$$
Consider the following objective least square objective for the domain 
$$\cL(a(\bu); \lambda)=\sum_{\bz \in \{\pm 1\}^P} \left( \E_{\bu\sim \barrho_{k_1}(\bu)}[a(\bu)\sigma(\<\bu,\bz\>)] - g_*(\bz) \right)^2+ \lambda \int a^2(\bu) \de \barrho_{k_1}(\bu)$$
While this is an infinite dimensional problem, the Representer's theorem holds and the interpolating solution exists. We refer the reader to \citep{celentano2021minimum} for a detailed analysis of interpolation with the random feature model. Moreover, $$\norm{a^*}_{\barrho_*} \leq \lambda_{\min}(\bK^{k_1})^{-1} \sqrt{\sum_{\bz \in \{\pm 1\}^P} g_*(\bz)^2}\leq \; c_1,$$ for some constant $c_1$ that only depends on $\lambda_{\min}(\bK^{k_1}), h_*, P$ and $ \bar{c}$.
\end{proof}

\begin{proof}[Proof of Lemma \ref{lem:convex-smoothe-properties}] For $t \in [0,1]$ and any $\barrho^{(1)}$ and $\barrho^{(2)}$, consider the density $\barrho=t \barrho^{(1)}+(1-t)\barrho^{(2)}.$ Then
\begin{align*}
 \cR(\barrho)&= \cR(t \barrho^{(1)}+(1-t)\barrho^{(2)}) =\E[\ell(\fNN(\bz;t \barrho^{(1)}+(1-t)\barrho^{(2)}),y)]\\
 &= \E[\ell(t\fNN(\bz;\barrho^{(1)})+(1-t)\fNN(\bz;\barrho^{(2)}),y)] \\
 & \leq \E[t\ell(\fNN(\bz;\barrho^{(1)}),y)+(1-t)\ell(\fNN(\bz;\barrho^{(2)}),y)]  \tag{using convexity}\\
 &= t \E\left[\ell(\fNN(\bz;\barrho^{(1)}),y)\right]+(1-t)\E \left[\ell(\fNN(\bz;\barrho^{(2)}),y) \right]=t\, \cR(\barrho^{(1)})+(1-t)\cR(\barrho^{(2)}).
\end{align*}
To show smoothness, first observe that a direct consequence of Assumption \ref{assump:nice-loss} (the condition $|\ell''(u,y)| \leq K$) is that for any $u_1,u_2 \in \R$, we have
\begin{equation}\label{eq:smothness-of-ell}
   \ell(u_1,y) \leq \ell(u_2,y)+\ell'(u_2,y) (u_1-u_2)+\frac{K}{2}(u_2-u_1)^2. 
\end{equation}
Using this, for any $\barrho^{(1)}$ and $\barrho^{(2)}$
\begin{align}
\cR(\barrho^{(1)})&=\E[\ell(\fNN(\bz;\barrho^{(1)}),y)] \nonumber\\
&  \leq \E[\ell(\fNN(\bz;\barrho^{(2)}),y)]+ \E\left[\ell'(\fNN(\bz;\barrho^{(2)}),y)\left(\fNN(\bz;\barrho^{(1)})-\fNN(\bz;\barrho^{(2)})\right) \right] \tag{using \eqref{eq:smothness-of-ell}} \nonumber\\
& \hspace{3mm}+ \frac{K}{2} \E\left[\left(\fNN(\bz;\barrho^{(1)})-\fNN(\bz;\barrho^{(2)}) \right)^2\right] \label{eq:first-dumb}
\end{align}
Since $\fNN(\bz;\barrho^{(1)})-\fNN(\bz;\barrho^{(2)})= \E_{a^{\scriptscriptstyle(1)},a^{\scriptscriptstyle(2)},\bu} \left[(a^{(1)}-a^{(2)})\sigma(\<\bu,\bz\>) \right]$, further simplifying the above
\begin{align}
&\E\left[\ell'(\fNN(\bz;\barrho^{(2)}),y)\left(\fNN(\bz;\barrho^{(1)})-\fNN(\bz;\barrho^{(2)})\right) \right] \nonumber\\
&= \E_{(y,\bz)\sim \mu_{y,\bz}}\left[\ell'(\fNN(\bz;\barrho^{(2)}),y) \E_{a^{\scriptscriptstyle(1)},a^{\scriptscriptstyle(2)},\bu} \left[(a^{(1)}-a^{(2)})\sigma(\<\bu,\bz\>) \right]\right]\nonumber\\
&= \E_{a^{\scriptscriptstyle(1)},a^{\scriptscriptstyle(2)},\bu} \left[\E_{(y,\bz)\sim \mu_{y,\bz}}\left[\ell'(\fNN(\bz;\barrho^{(2)}),y)\sigma(\<\bu,\bz\>)\right] (a^{(1)}-a^{(2)}) \right] \nonumber\\
&=\left\<\nabla_a\cR (\barrho^{(2)}),a^{(1)}-a^{(2)}\right\>_{\barrho^{(1)},\barrho^{(2)}}. \label{eq:2nd-dumb}
\end{align}
Also, simplifying the third term of \eqref{eq:first-dumb}
\begin{align}
\frac{K}{2}& \E\left[\left(\fNN(\bz;\barrho^{(1)})-\fNN(\bz;\barrho^{(2)}) \right)^2\right]= \frac{K}{2}\E_{(y,\bz)\sim \mu_{y,\bz}}\left[ \E_{a^{\scriptscriptstyle(1)},a^{\scriptscriptstyle(2)},\bu}\left[(a^{(1)}-a^{(2)}) \sigma(\<\bu,\bz\>)\right]^2\right] \nonumber\\
&\leq  \frac{K}{2}\E_{(y,\bz)\sim \mu_{y,\bz}} \E_{a^{\scriptscriptstyle(1)},a^{\scriptscriptstyle(2)},\bu}\left[(a^{(1)}-a^{(2)})^2 \sigma^2(\<\bu,\bz\>)\right]\nonumber \tag{by Jensen's ineqaulity}\\
&\leq \frac{K^3}{2}\lVert a^{(1)}-a^{(2)} \rVert^2_{\barrho^{(1)},\barrho^{(2)}}. \label{eq:3rd-dumb}
\end{align}
Finally, combining \eqref{eq:2nd-dumb} and \eqref{eq:3rd-dumb} with \eqref{eq:first-dumb}, we obtain
\begin{align*}
    \cR(\barrho^{(1)})&=\cR(\barrho^{(2)})+\left\<\nabla_a\cR (\barrho^{(2)}),a^{(1)}-a^{(2)}\right\>_{\barrho^{(1)},\barrho^{(2)}}+\frac{K^3}{2}\lVert a^{(1)}-a^{(2)} \rVert^2_{\barrho^{(1)},\barrho^{(2)}},
\end{align*}
which, by definition, gives us $H$-smoothness with $H=K^3$.
\end{proof}
\begin{proof}[Proof of Lemma \ref{lem:convex-smooth-convergence}]
First, the classical descent lemma holds due to $H$-smoothness for step-size $\eta \leq \frac{1}{H}$.
\begin{align}
    f(x^{j+1}) & \leq f(x^{j}) + \<\nabla f(x^j), x^{j+1}-x^{j} \> + \frac{H}{2} \norm{x^{j+1}-x^j}^2_2 \nonumber\\
    & =f(x^{j}) -\eta\left(1-\frac{\eta H}{2} \right)\norm{\nabla f(x^j)}^2_2 \leq f(x^{j}) -\frac{\eta }{2}\norm{\nabla f(x^j)}^2_2 \leq f(x^j). \label{eq:descent} 
\end{align}
By convexity, for any $\tilde{x}$ (not necessarily a minimizer), we also have 
$$f(x^j) \leq f(\Tilde{x})+\<\nabla f(x^j), x^j-\tilde{x}\>.$$
Substituting this in \eqref{eq:descent}
\begin{align*}
     f(x^{j+1}) & \leq f(\Tilde{x})+\<\nabla f(x^j), x^j-\tilde{x}\>-\frac{\eta }{2}\norm{\nabla f(x^j)}^2_2 \\
     &=f(\Tilde{x})+ \frac{1}{2\eta} \left(2\eta \<\nabla f(x^j), x^j-\tilde{x}\>-\eta^2\norm{\nabla f(x^j)}^2_2 \right)\\
     &=f(\Tilde{x})+ \frac{1}{2\eta} \left(2\eta \<\nabla f(x^j), x^j-\tilde{x}\>-\eta^2\norm{\nabla f(x^j)}^2_2 - \norm{x^j-\Tilde{x}}_2^2 + \norm{x^j-\Tilde{x}}_2^2\right)\\
     &=f(\Tilde{x})+ \frac{1}{2\eta} \left( \norm{x^j-\Tilde{x}}_2^2 - \norm{x^{j+1}-\Tilde{x}^2}_2^2\right).
\end{align*}
Therefore, summing over $j \in \{0,\dots,k-1\}$, since the right hand side is a telescopic sum
\begin{align*}
    \sum_{j=0}^{k-1} \left(f(x^{j+1})-f(\Tilde{x}) \right) \leq \frac{1}{2\eta}\left( \norm{x^0-\Tilde{x}}_2^2-\norm{x^{k}-\tilde{x}}_2^2 \right) \leq \frac{\norm{x^0-\Tilde{x}}_2^2}{2\eta}.
\end{align*}
Note that the lemma statements trivially holds if $f(x^k)-f(\Tilde{x}) \leq 0$ already. Therefore, in the case when $f(x^k) > f(\Tilde{x})$ and using the descent lemma, we finally conclude the proof.
$$ f(x^k)-f(\Tilde{x}) \leq \frac{1}{k}  \sum_{j=0}^{k-1} \left(f(x^{j})-f(\Tilde{x}) \right) \leq \frac{\norm{x^0-\Tilde{x}}^2}{2\eta k}.$$
\end{proof}
 \subsection{Phase 1 (non-linear training)}\label{app:leap_one_ub}
 We now analyze Phase 1 of the algorithm and show that $\lambda_{\min}(\bK^{k_1}) \geq c$, for a constant  $c:=c(\eta, \mu, \P,K)$ where $\eta$ is also a small constant (in terms of $\mu,P,K$).
\paragraph{Writing the weight evolution with a polynomial.}
We will by considering a general polynomial activation $\sigma(x)=\sum_{l=0}^{L} m_l x^l$ of degree $L$, whose coefficients are given by $\bm=(m_0,\dots,m_L)$.
\begin{lemma}(Training dynamics given by a polynomial)\label{lem:pki-baru} Let $\bxi=(\xi_{S,k})_{S \subseteq [\P], 0 \leq k \leq k_1-1} \in \R^{2^\P k_1}$, $\zeta \in \R$, $\brho \in \R^{L+1}, \bgamma \in \R^{\P}$ be variables. 
For each $i \in [\P]$, define $p_{0,i}(\zeta, \bxi, \brho,\bgamma) \equiv 0$. For each, $0 \leq k \leq k_1-1$, 
\begin{align*}
    p_{k+1,i}(\zeta, \bxi, \brho, \bgamma)&=p_{k,i}(\zeta, \bxi, \brho, \gamma)+\zeta \gamma_{i} \rho_1 \bxi_{\{i\},k} \\
    & \hspace{2mm}+ \zeta \gamma_i \sum_{r=1}^{L-1} \frac{\rho_{r+1}}{r!} \sum_{(i_1,\dots, i_r) \in [P]^r} \xi_{i \oplus i_1 \oplus \dots \oplus i_r,k} \prod_{l=1}^{r} p_{k, i_l}(\zeta, \bxi, \brho, \bgamma).
\end{align*}
Then there is a constant $c:=c(k_1, P, K)$ such that for any $0<\eta< c$
$$ u_{i}^k(a)=p_{k,i}(\eta a, \bbeta, \bm, \bkappa),$$
and $\bbeta = (\beta_{S,k})_{S \subseteq [P], 0 \leq k \leq k_1 - 1}$ is given by  $\beta_{S,k}=-\E[\ell'(\fNN(\bz;\bar\rho_k),y)\chi_S(\bz)].$
\end{lemma}
The term $\fNN(\cdot;\bar\rho_k)$ evolves non-linearly, and difficult to analyze directly. However, for sufficiently small step size $\eta$, the interaction term $\fNN(\cdot; \bar\rho_k)$ is small, and we show that it can be ignored. Formally, we define the simplified dynamics $\hatbu^k(a)$ for each $0 \leq k \leq k_1$ by letting $\hatbu^0(a) = \bzero$ and inductively setting for each $k \in \{0,\ldots,k_1-1\}$,
\begin{align*}
\hatbu^{k+1}(a) = \hatbu^k(a) - \eta \bkappa \circ \E[\ell'(\fNN(\bz;\bar\rho_0),y) \cdot a \cdot \sigma'(\<\hatbu^k(a),\bz\>)\bz].
\end{align*}
Using a similar argument, we may show:
\begin{lemma}[Simplified training dynamics are given by a polynomial]\label{lem:pki-hatu-discrete}
There is a constant $c > 0$ depending only on $k_1,P,K$, such that for any $0 < \eta < c$, any $i \in [P]$ and any $0 \leq k \leq k_1$, we have
$$\hat u_{i}^k(a) = p_{k,i}(\eta a, \balpha,\bm,\bkappa),$$
where we abuse notation (since $\balpha = (\alpha_S)_{S \subseteq [P]}$ otherwise) and let $\balpha = (\alpha_{S,k})_{S \subseteq [P], 0 \leq k \leq k_1 - 1}$ be given by
\begin{align*}
\alpha_{S,k} &= \alpha_S = \E[-\ell'(\fNN(\bz;\bar\rho_0),y)\chi_S(\bz)]
\end{align*}
\end{lemma}

\paragraph{Reducing to analyzing simplified dynamics} We lower-bound $\lambda_{\mathrm{min}}(\bK^{k_1})$ in terms of the determinant of a certain random matrix. Let $\bzeta = [\zeta_1,\ldots,\zeta_{2^P}]$ be a vector of $2^P$ variables. Define $\bM = \bM(\bzeta,\bxi,\brho,\bgamma) \in \R^{2^P \times 2^P}$ to be the matrix indexed by $\bz \in \{+1,-1\}^P$ and $j \in [2^P]$ with entries
\begin{align*}
M_{\bz,j}(\bzeta,\bxi,\brho,\bgamma) = \sum_{r=0}^{L} \frac{\rho_r}{r!} \left(\sum_{i=1}^P z_i p_{k,i}(\zeta_j,\bxi,\brho,\bgamma)\right)^r.
\end{align*}

This matrix is motivated by the following fact:
\begin{lemma}\label{lem:plug-into-M}
There is a constant $c > 0$ depending only on $k_1,P,K$, such that for any $0 < \eta < c$, and any $\ba = [a_1,\ldots,a_{2^P}] \in [-1,1]^{2^P}$, we have
\begin{align*}
M_{\bz,j}(\eta \ba, \bbeta, \bm,\bkappa) &= \sigma(\<\bu^{k_1}(a_j), \bz\>) \\
M_{\bz,j}(\eta \ba, \balpha, \bm, \bkappa) &= \sigma(\<\hatbu^{k_1}(a_j), \bz\>)
\end{align*}
\end{lemma}
Under this notation, using the exactly same analysis as in \cite[Lemma E.9]{abbe2022merged}, we can also show 
\begin{lemma}\label{lem:k-min-from-det-m}
There is a constant $c > 0$ depending on $K,P$ such that for any $0 < \eta < c$,
$$\lambda_{\mathrm{min}}(\bK^{k_1}) \geq c \E_{\ba \sim \mu_a^{\otimes 2^P}}[\det(\bM(\eta \ba, \bbeta, \bm, \bkappa))^2].$$
\end{lemma}
On the other hand, we can prove a lower-bound on $\E[\det(\bM(\eta \ba, \bbeta, \bm,\bkappa))^2]$ simply by lower-bounding the sum of magnitudes of coefficients of $\det(\bM(\bzeta, \balpha, \bm,\bkappa))$ when viewed as a polynomial in $\bzeta$ almost surely over the choice of $\bar{c}$. This is because of (a) the fact that $\det(\bM(\bzeta, \balpha, \bm,\bkappa))$ and $\det(\bM(\bzeta, \bbeta, \bm,\bkappa))$ have coefficients in $\bzeta$ that are $O(\eta)$-close for $\eta$ small, and (b) the fact that polynomials anti-concentrate over random inputs.
\paragraph{Analysis of simplified dynamics.} The proof is now reduced to showing that $\det(\bM(\bzeta,\balpha,\bm,\bkappa)) \not\equiv 0$, as a polynomial in $\bzeta$. In other words, by Lemma~\ref{lem:plug-into-M}, we are now focusing on only the simplified dynamics $\hat\bu^k$.
If we can show that $\det(\bM(\bzeta,\balpha,\bm,\bkappa)) \not\equiv 0$ almost surely over the choice of $\bkappa \in [1/2,3/2]^\P$, then Theorem \ref{thm:leap-one-ub} follows. Therefore, introducing the variables $\bgamma \in \R^{\P}$, it suffices to show that
\begin{equation}\label{eq:suffices-poly-nonzero}
    \det(\bM(\bzeta,\balpha,\bm,\bgamma)) \not\equiv 0, \text{ as a polynomial in } \bzeta, \bgamma.
\end{equation} 
We start by following simple observation.
\begin{claim}\label{clm:Leap-1-of-alpha-sets}
   Almost surely over the choice of initialization $\bar{c}$, the set of subsets $\cK:=\{S: \alpha_S \neq 0\}$ has $\Leap(\cK)=1$. 
\end{claim}
\begin{proof}
This basically follows from the fact that $\ell$ is piecewise analytic. The argument is as follows: we have that $\Leap_{\DLQ_{\ell}}=\Leap(\cC_{\DLQ_{\ell}})=1$, where $\cC_{\DLQ_{\ell}}=\{S: \exists u \in \R \text{ such that } \E[\ell'(u,y)\chi_S(\bz)] \neq 0\}.$ Since $\ell$ is piece-wise analytic, so is $\ell'(u,y)$ in the first argument. Thus, for any $S \in \cC_{\DLQ_{\ell}}$, the set $\{u: \E[\ell'(u,y)\chi_S(\bz)]=0\}$ has measure 0. Thus, almost surely over the choice of $\bar{c}=\fNN(\bz;\bar\rho_0)$, we have that for any $S \in \cC_{\DLQ_{\ell}}$ 
$$\E[\ell'(\bar{c},y) \chi_S(\bz)]=\E[\ell'(\fNN(\bz;\bar\rho_0),y)\chi_S(\bz)]=-\alpha_S \neq 0.$$
This implies that $S \in \cK$ as well. We finally conclude that $\Leap_{\DLQ_{\ell}}=\Leap(\cK)=1$.
\end{proof}
Using ideas similar to \cite{abbe2022merged}, we now prove \eqref{eq:suffices-poly-nonzero} by analyzing the recurrence relations for $p_{k,i}$ to show that to first-order the polynomials $p_{k_1,i}$ are distinct for all $i \in [P]$, but now doing smooth analysis over $\bgamma$ instead. Formally we show the following lemma.
\begin{lemma}\label{lem:nonzero-poly-discrete-msp}
Suppose that $L \geq 2^{8P}$ and let $m_i = i! \binom{L}{i}$ for all $0 \leq i \leq L$, which correspond to the activation function $\sigma(x) = (1+x)^L$. Let $k_1 = P$. Then $\det(\bM(\bzeta,\balpha,\bm,\bgamma)) \not\equiv 0$, i.e. \eqref{eq:suffices-poly-nonzero} holds.
\end{lemma}

\begin{claim}\label{clm:ubarwtbound}
For any time step $0 \leq k \leq k_1$ any $j \in [N]$, a learning rate $\eta < 1/ (4 K^2(1+K)P k )$, and any $a \in [-1,1]$ we have $$\|\bu^k(a)\|_1, \|\hat{\bu}^k(a)\|_1 \leq 2\eta K^2 (1+K)P k \leq 1/2.$$
\end{claim}
\begin{proof}
We prove this by induction on $k$. For $k=0$, we have $\bu^0 = \hatbu^0 = \bzero$ and the claim holds trivially. For the inductive step, $\fNN(\bz;\bar\rho_k) \leq \E_{a \sim \mu_a}[|a| |\sigma(\<\bu^k,\bz\>)|] \leq \|\sigma\|_{\infty} \leq K$, since $a \sim \Unif([-1,1])$.
Therefore
\begin{align*}
& \|\bu^{k+1}(a)\|_1 \leq \|\bu^k(a)\|_1 + \eta \| \bkappa \circ \E_{y,\bz}[-\ell'(\fNN(\bz;\bar\rho_k),y)a \sigma'(\<\bu^k, \bz\>) \bz]\|_1 \\
& \leq \|\bu^k(a)\|_1 + \eta \infnorm{\bkappa} \E_{y,\bz}[\| -\ell'(\fNN(\bz;\bar\rho_k),y)a \sigma'(\<\bu^k, \bz\>) \bz\|_1]\\
&\leq \|\bu^k(a)\|_1 + 2 \eta K(1+K) \cdot K \cdot \E_{\bz}[\|\bz\|_1] \leq \|\bu^k(a)\|_1 + 2 \eta K^2 (1+K) P \leq 2 \eta K^2(1+K) P k. 
\end{align*}
Note that here we used $\infnorm{\bkappa} \leq 2$, $|\ell'(\fNN(\bz;\bar\rho_k),y)| \leq K (1+|\fNN(\bz;\bar\rho_k)|) \leq K(1+K)$ using $\sA2$, and $|a \sigma'(\<\bu^k, \bz\>)| \leq |a|\cdot |\sigma'(\<\bu^k, \bz\>)| \leq K$. The bound for $\|\hat\bu^k(a)\|_1$ follows exactly the same argument but now we use $|\ell'(\fNN(\bz; \bar\rho_0),y)| \leq K$. 
\end{proof}
\begin{proof}[Proof of Lemmas \ref{lem:pki-baru} and \ref{lem:pki-hatu-discrete}]
We have $\zeta \in \R$, $\bxi = (\xi_{S,k})_{S \subseteq [P], k \in \{0,\ldots,k_1-1\}}$, $\brho \in \R^{L+1}$, and $\bgamma \in \R^P$ as variables. Let $s_0,\ldots,s_{k_1-1} : \{+1,-1\}^P \to \R$ be $s_k(\bz) = \sum_{S \subseteq [P]} \xi_{S,k} \chi_S(\bz)$. Consider the recurrence relation $\bnu^k \in \R^P$, where we initialize $\bnu^0 = \bzero$ and, for $0 \leq k \leq k_1-1$,
 \begin{align}\label{eq:genericrecrelation}
 \bnu^{k+1}& = \bnu^k + \zeta \bgamma \circ \E_{\bz}\Big[s_k(\bz) \sum_{r=0}^{L-1} \frac{\rho_{r+1}}{r!}\<\bnu^k,\bz\>^r \bz \Big].
 \end{align}
 This recurrence is satisfied by $\zeta=\eta a$, $\brho=\bm$, $ \bgamma=\bkappa$, and $s_k(\bz)=\sum_{S} \beta_{S,k} \chi_S(\bz)$. This is because
 \begin{align*}
     \bu^{k+1}&=\bu^{k}-\eta \cdot \bkappa \circ \E [\ell'(\fNN(\bz; \bar\rho_k),y)a \sigma'(\<\bu^{k}, \bz\>) \bz]\\
    &=\bu^{k}+\eta a \cdot \bkappa \circ \E [-\ell'(\fNN(\bz; \bar\rho_k),y) \sum_{r=0}^{L-1} \frac{\bm_{r+1}}{r!}\<\bu^{k}, \bz\>^{r} \bz] \\
    &= \bu^{k}-\eta \cdot \bkappa \circ \E [\ell'(\fNN(\bz; \bar\rho_k),y)a \sigma'(\<\bu^{k}, \bz\>) \bz]\\
    &=\bu^{k}+\eta a \cdot \bkappa \circ \E [-\ell'(\fNN(\bz; \bar\rho_k),y) \sum_{r=0}^{L-1} \frac{\bm_{r+1}}{r!}\<\bu^{k}, \bz\>^{r} \bz] \\
    &=\bu^{k}+\eta a \cdot \bkappa \circ \E_{\bz} [ \E_{y\mid \bz}[-\ell'(\fNN(\bz; \bar\rho_k),y)] \sum_{r=0}^{L-1} \frac{\bm_{r+1}}{r!}\<\bu^{k}, \bz\>^{r} \bz] 
 \end{align*}
Finally, noting that $ \E_{y\mid \bz}[-\ell'(\fNN(\bz; \bar\rho_k),y)]=\sum_{S} \beta_{S,k} \chi_S(\bz)$, we showed that recurrence \eqref{eq:genericrecrelation} holds with specified value of variables. Similarly, $\hat\bu^k(a)$ with $s_k(\bz) = \sum_{S} \alpha_S \chi_S(\bz)$ for $\alpha_S=-\E[\ell'(\fNN(\bz; \bar\rho_0,y)\chi_S(\bz)]$. This is because $|\<\bu^k,\bz\>|, |\<\hatbu^k,\bz\>| \leq 1/2 < 1$ by Claim~\ref{clm:ubarwtbound} and in the interval $(-1,1)$, we have $\sigma(x) = \sum_{r=0}^L \frac{m_r}{r!} x^r$. It finally remains to show that
$$\nu_i^k = p_{k,i}(\zeta,\bxi,\brho, \bgamma).$$  
The proof is by induction on $k$. For $k = 0$, it is true that $p_{0,i}(\zeta, \bxi,\brho,\bgamma) = 0 = \nu_i^0$. For the inductive step, for any $r \geq 1$ and $i \in [d]$, we can write
\begin{align*}
\gamma_i \cdot \E_{\bz}[s_k(\bz)\<\bnu^k,\bz\>^r z_i] &= \gamma_i \E_{\bz} \Big[s_k(\bz) z_i \sum_{(i_1,\ldots,i_r) \in [P]^r} \prod_{l=1}^r \nu_{i_{l}}^k z_{i_{l}} \Big] \\
&= \gamma_i \sum_{(i_1,\ldots,i_r) \in [P]^r}
\E_{\bz} \Big[s_k(\bz)\chi_i(\bz)\prod_{l=1}^r \chi_{i_l}(\bz)\Big] \prod_{l=1}^r p_{k,i_l}(\zeta,\bxi,\brho,\bgamma) \\
&= \gamma_i \sum_{(i_1,\ldots,i_r) \in [P]^r} \xi_{\{i\} \oplus \{i_1\} \oplus \dots \oplus \{i_r\},k} \prod_{l=1}^r p_{k,i_l}(\zeta,\bxi,\brho,\bgamma),
\end{align*}
and $\gamma_i \E_{\bz}[s_k(\bz) \<\bnu^k,\bz\>^0 z_i] = \gamma_i \E_{\bz}[s_k(\bz) z_i] = \gamma_i \xi_{\{i\},k}$. The inductive step follows by linearity of expectation.
\end{proof}

\subsection{Proof of Lemma~\ref{lem:nonzero-poly-discrete-msp}}

In Lemma \ref{lem:nonzero-poly-discrete-msp}, we have already fixed $\bm \in \R^{L+1}$ to be $m_i = i! \binom{L}{i}$ for all $i \in \{0,\ldots,L\}$. This corresponds to the activation function $\sigma(x) = (1+x)^L$.

\subsubsection{Reducing to minimal Leap 1 structure}
To show that $\det(\bM(\bzeta,\balpha,\bm,\bgamma)) \not\equiv 0$ as a polynomial in $\bzeta,\bgamma$, we first show that it suffices to consider ``minimal'' Leap 1 set structure, without loss of generality.
\begin{claim}
Let $\cK' \subseteq \cK$ such that $\Leap(\cK')=1$. Then if $$\det(\bM(\bzeta,\balpha,\bm,\bgamma))\mid_{\alpha_S = 0 \text{ for all } \alpha \in \cK \setminus \cK'} \not\equiv 0, \text{ as a polynomial of } \bzeta,\bgamma$$ we have
$$\det(\bM(\bzeta,\balpha,\bm,\bgamma)) \not\equiv 0 \text{ as a polynomial of } \bzeta,\bgamma.$$
\end{claim}
\begin{proof}
Directly substituting $\alpha_S=0$ for all $S \in \cK \setminus \cK'$.
\end{proof}
Therefore, without loss of generality (up to relabelling the indices of the variables), we assume that 
$$\cK = \{S_1,\ldots,S_P\},\text{ where, for all $i \in [P]$ }, i \in S_i \text{ and } S_i \subseteq [i].$$ Otherwise, we could remove a set from $\cK$ while still having $\Leap(\cK)=1$. 

\subsubsection{Weights to leading order}
Let us define the polynomials $q_{k,i}$ in variables $\bzeta,\bphi,\brho, \bgamma$ where $\bphi=(\phi_S)_{S\in \cK}$. For all $k \in \{0,\ldots,k_1-1\}$ and $i \in [P]$,
\begin{align*}
q_{k,i}(\zeta,\bphi,\brho,\bgamma) = p_{k,i}(\zeta,\bxi,\brho,\bgamma) \mid_{\xi_{S,k} = 0 \mbox{ for all } S \not\in \cK \text{ and } \xi_{S,k} = \phi_S \mbox{ for all } S \in \cK}.
\end{align*}
Therefore $\bM(\bzeta,\bphi,\brho,\bgamma)$ has entries $M_{\bz,j}(\bzeta,\bphi,\brho,\bgamma) = \sum_{r=0}^L \frac{\rho_r}{r!} \left(\sum_{i=1}^P q_{k,i}(\zeta_j,\bphi,\brho,\bgamma)\right)^r$.
We will explicitly compute the lowest degree term in $\zeta$. First, we show that many terms are zero. To this end, consider the following notation:
Recursively define 
$$o_i = 1 + \sum_{i' \in S_i \setminus \{i\}} o_{i'}\text{ for all } i \in [P],$$ where the sum over an empty set is $0$ by convention. Additionally, let $\tilde{q}_{k,i}(\phi,\brho,\bgamma)$ to be the coefficient of the term in $q_{k,i}(\zeta, \bphi,\brho,\bgamma)$ whose degree in $\zeta$ is $o_i$. 
Furthermore, recursively define 
$$g_i(\bgamma)=\gamma_i \text{ if } \{i\}\in \cK,\text{ and otherwise } g_i(\bgamma)=\gamma_i \prod_{i'\in S_i \setminus \{i'\}}g_{i'}(\bgamma).$$
\begin{lemma}\label{lem:oi-def}
Under the above notation, we have that $q_{k,i}(\zeta,\bphi,\bm, \bgamma)$ has no nonzero terms of degree less than $o_i$ in $\zeta$. Furthermore, $\tilde{q}_{k,i}(\bphi,\brho,\bgamma)$ can be decomposed as $\tilde{q}_{k,i}(\bphi,\brho, \bgamma)=g_i(\bgamma) \cdot \hat{q}_{k,i}(\bphi,\brho)$ for some $\hat{q}_{k,i}(\bphi,\brho)$.
\end{lemma}
\begin{proof}
We will prove this by induction on $k$. The base case for $k = 0$ trivially holds because $q_{0,i} \equiv 0$. For the inductive step, we assume the statement holds for all $k' \in \{0,\ldots,k\}$ and we prove the statement for $k+1$. By the recurrence relation of the polynomials $q_{k,i}$, we have
\begin{align*}
&q_{k+1,i}(\zeta,\bphi,\brho,\bgamma) = q_{k,i}(\zeta,\bphi,\brho,\bgamma) +  \gamma_i \zeta \rho_1 \phi_{\{i\}} \ind(\{i\} \in \cK) \\
&\quad + \gamma_i \zeta \sum_{r=1}^{L-1} \frac{\rho_{r+1}}{r!} \sum_{(i_1,\ldots,i_r) \in [P]^r} \phi_{\{i\} \oplus \{i_1\} \oplus \dots \oplus \{i_r\}} \ind(\{i\} \oplus \{i_1\} \oplus \dots \oplus \{i_r\} \in \cK) \prod_{l=1}^r q_{k,i_l}(\zeta,\bphi,\brho,\bgamma).
\end{align*}
It suffices to show that the claim holds for each one of the three terms above.
\begin{enumerate}
\item The first term, $q_{k,i}(\zeta,\bphi,\brho,\bgamma)$, is handled by the inductive hypothesis. We can conclude that all the terms of degree less than $o_i$ in $\zeta$ are zero. Moreover, $\tilde{q}_{k,i}=g_i(\bgamma)\hat{q}_{k,i}(\brho,\bgamma)$.
\item The second term is nonzero only in the case that $\{i\} \in \cK$, in which case $S_i = \{i\} \in \K$ and $o_i = 1$, so we do not have a contradiction since the $\zeta^{o_i}=\zeta$ only and we do not have terms with degree less than $o_i$. Moreover, the coefficient of $\zeta$ is $\gamma_i \rho_1 \phi_{\{i\}}=g_i(\bgamma) \rho_1 \phi_{\{i\}}$.
\item We break into cases. \textit{Case a}. If $\{i\}\oplus \{i_1\} \oplus \dots \oplus \{i_r\} = S_i$, then $S_i \setminus \{i\} \subset \{i_1,\ldots,i_r\}$, so $1 + \sum_{l=1}^r o_{i_l} \geq o_i$, and so no new terms of degree less than $o_i$ are added. Moreover, the only way the third term has degree $o_i$ in $\zeta$ is when $i \neq i_1 \neq \dots \neq i_r$ while still having $\{i, i_1, \dots, i_r\}=S_i$; otherwise the power of $\zeta$ is $1+\sum_{l=1}^{r}o_{i_l}>o_i$. In that case, letting $r'=|S_i|-1$, the coefficient of $\zeta^{o_i}$ in the third term is 
\begin{align*}
   & \gamma_i\left( \frac{\rho_{r'+1}}{r'!} \phi_{S_i} \prod_{i'\in S_i \setminus \{i\}}^r \tilde{q}_{k,i'}(\bphi,\brho,\bgamma)  \sum_{(i_1,\ldots,i_{r'}) \in [P]^{r'}}  \ind(\{i, i_1,\dots,i_{r'}\}=S_i ) \right) \\
   & =\gamma_i \left( \frac{\rho_{r'+1}}{r'!} \phi_{S_i} \prod_{i' \in S_i \setminus\{i\}} g_{i'}(\bgamma)\hat{q}_{k,i'}(\brho,\bgamma) \sum_{(i_1,\ldots,i_{r'}) \in [P]^{r'}} \ind(\{i, i_1,\dots,i_{r'}\}=S_i)  \right) \\
   & =\gamma_i \prod_{i'\in S_i \setminus\{i'\}}g_{i'}(\bgamma) \left(  \frac{\rho_{r'+1}}{r'!} \phi_{S_i} \prod_{i' \in S_i \setminus\{i\}} \hat{q}_{k,i'}(\brho,\bgamma) \sum_{(i_1,\ldots,i_{r'}) \in [P]^{r'}} \ind(\{i, i_1,\dots,i_{r'}\}=S_i) \right) \\
   &=g_i(\bgamma) \cdot \hat{q}(\brho,\bgamma) \text{ for a certain $\hat{q}$}.
\end{align*}
\textit{Case b}. If $\{i\}\oplus \{i_1\} \oplus \dots \oplus \{i_r\} = S_{i'}$ for some $i' \neq i$, then either $i \in \{i_1,\ldots,i_r\}$, in which case $1 + \sum_{l=1}^r o_{i_{l}} > o_i$. Otherwise, we must have $i' > i$. But in this case $o_{i'} > o_i$ since $i \in S_{i'}$, so we also have $\sum_{l=1}^r o_{i_{l}} > o_i$ and again no new terms of degree less than $o_i$ are added. In fact, only terms of degree strictly more than $o_i$ are added. Thus, in either case the coefficient of the term with degree $o_i$ in $\zeta$ is simply 0.
\end{enumerate} 
Finally, we showed that each of the three terms have at least the degree $o_i$ in $\zeta$. Moreover, each term can be decomposed as $g_i(\bgamma) \hat{q}(\brho,\bgamma)$ for a certain $\hat{q}$. Thus, overall we can decompose
$$\tilde{q}_{k,i}(\bphi,\brho,\bgamma)=g_i(\bgamma) \cdot \hat{q}_{k+1,i}(\bphi,\brho), \text{concluding the proof of the lemma.}$$
\end{proof}

\begin{claim}\label{claim:rzdistinct-discrete} Let $$r_{\bz}(\zeta,\bphi,\brho,\bgamma) = \sum_i z_i q_{k_1,i}(\zeta,\bphi,\brho,\bgamma).$$
Then, for each distinct pair $\bz,\bz' \in \{+1,-1\}^P$, we have
$r_{\bz}(\zeta,\bphi,\bm) - r_{\bz'}(\zeta,\bphi,\bm) \not\equiv 0$ as a polynomial in $\zeta$ and $\bgamma$.
\end{claim}
\begin{proof}
Recall the definition of $o_i$ from Lemma~\ref{lem:oi-def}. Fix $i \in [P]$ be such that $z_i \neq z'_i$ and $o_i$ is minimized. It is always possible to choose such an $i$ since $\bz \neq \bz'$. Again $\tilde{r}_{\bz}(\bphi,\brho,\bgamma)$ is the term with degree $o_i$ in $\zeta$. Using Lemma~\ref{lem:oi-def} then
\begin{align*}
\tilde{r}_{\bz}(\bphi,\brho,\bgamma) - \tilde{r}_{\bz'}(\bphi,\brho,\bgamma) = \sum_{i' \mbox{ s.t. } o_{i'} = o_i, z_{i'} \neq z'_{i'}} (z_{i'} - z'_{i'}) \tilde{q}_{k_1,i'}(\bphi,\brho,\bgamma),
\end{align*}
but $\tilde{q}_{k_1,i'}$ are non-zero polynomials in $\bgamma$ with different dependence that is captured by $g_{i'}(\bgamma)$ according to Lemma~\ref{lem:oi-def}. Thus, $r_{\bz}(\zeta,\bphi,\bm) - r_{\bz'}(\zeta,\bphi,\bm) \not\equiv 0$.
\end{proof}

\subsubsection{Linear Independence of Powers of Polynomials}
Similar to \cite{abbe2022merged}, we finish our proof using the classical result of \cite{newman1979waring} about the linear independence of large powers of polynomials. 
\begin{proposition}[Remark 5.2 in \cite{newman1979waring}]\label{prop:newmanslater}
Let $R_1,\ldots,R_m \in \C[\zeta]$ be non-constant polynomials such that for all $i \neq i' \in [m]$ we have $R_i(\zeta)$ is not a constant multiple of $R_{i'}(\zeta)$. Then for $L \geq 8m^2$ we have that $(R_1)^L,\ldots,(R_m)^L \in \C[\zeta]$ are $\C$-linearly independent.
\end{proposition}
We will finally show that $\det(\bM(\bzeta,\balpha,\bm,\bgamma)) \not\equiv 0$.

\begin{proof}[Proof of Lemma~\ref{lem:nonzero-poly-discrete-msp}]
Let us fix $\bkappa \in \R^P$ such that for all $\bz \neq \bz'$, we have $r_{\bz}(\zeta,\balpha,\bm, \bkappa) - r_{\bz'}(\zeta,\balpha,\bm,\bkappa) \not\equiv 0$ as polynomials in $\zeta$. This can be ensured by drawing $\bkappa \sim \Unif([0.5,1.5]^P)$, since for all $\bz \neq \bz'$ we have $r_{\bz}(\zeta,\balpha,\bm, \bgamma) - r_{\bz'}(\zeta,\balpha,\bm, \bgamma) \not\equiv 0$ as polynomials in $\zeta,\bgamma$ by Claim~\ref{claim:rzdistinct-discrete}. We now rewrite $\tilde{r}_{\bz}(\zeta) = r_{\bz}(\zeta,\balpha,\bm, \bkappa)$ to further indicate that the variables $\bphi = \balpha, \brho = \bm$ and $\bgamma=\bkappa$ are instantiated, and that we are looking at a polynomial over $\zeta$. We have also chosen $m_i = i! \binom{L}{i}$ for all $i \in \{0,\ldots,L\}$. We then have $$M_{\bz,j}(\bzeta,\balpha,\bm, \bkappa) = (1 + \tilde{r}_{\bz}(\zeta_j))^L.$$ 
Also, observe from the recurrence relations $\zeta$ divides $q_{k_1,i}(\zeta,\balpha,\bm,\bkappa)$ for each $i \in [P]$, so $\zeta_j$ divides $\tilde{r}_{\bz}(\zeta_j) = \sum_{i=1}^P z_i q_{k_1,i}(\zeta_j,\balpha,\bm,\bkappa)$. Therefore, polynomials $(1 + \tilde{r}_{\bz}(\zeta_j))$ and $ ,(1 + \tilde{r}_{\bz'}(\zeta_j))$ are not constant multiples of each other for any $\bz \neq \bz'$.

Now see $\zeta$ as variables, and construct the Wronskian matrix over the $L$-th power polynomials $\{(1 + \tilde{r}_{\bz}(\zeta))^L\}_{\bz \in \{+1,-1\}^P}$. This is a $2^P \times 2^P$ matrix $\bH(\zeta)$ whose entries are indexed by $\bz$ and $l \in [2^P]$ and defined by:
\begin{align*}
H_{\bz,l}(\zeta) = \frac{\partial^{l-1}}{\partial \zeta^{l-1}} (1 + \tilde{r}_{\bz}(\zeta))^L.
\end{align*}
Applying Proposition~\ref{prop:newmanslater} implies the polynomials $\{(1 +\tilde{r}_{\bz}(\zeta))^L\}_{\bz \in \{+1,-1\}^P}$ are linearly-independent, and so the Wronskian determinant is nonzero as a polynomial in $\zeta$:
\begin{align*}
\det(\bH(\zeta)) \not\equiv 0.
\end{align*}
Also, observe that we can write $\det(\bH(\zeta)) = \frac{\partial}{\partial \zeta_2} \frac{\partial^2}{\partial \zeta_3^2} \dots \frac{\partial^{2^P-1} }{ \partial \zeta_{2^P}^{2^P-1}} \det(\bM(\bzeta,\balpha,\bm,\bkappa)) \mid_{\zeta = \zeta_1 = \dots = \zeta_{2^P}}$. 
This finally  gives us  $\det(\bM(\bzeta,\balpha,\bm, \bkappa)) \not\equiv 0$ as a polynomial in $\bzeta$, and thus, $\det(\bM(\bzeta,\balpha,\bm, \bgamma)) \not\equiv 0$ as a polynomial in $\bzeta$ and $\bgamma$.
\end{proof}


\end{document}